\documentclass[10pt,letterpaper,twoside,twocolumn,journal,final]{IEEEtran}

\usepackage{amsmath,amsfonts}
\usepackage{algorithm}
\usepackage{array}
\usepackage[caption=false,justification=raggedright]{subfig}
\usepackage{textcomp}
\usepackage{stfloats}
\usepackage{url}
\usepackage{verbatim}
\usepackage{graphicx}
\usepackage{cite}
\usepackage{bm}
\usepackage{mathtools}
\usepackage{url}
\usepackage{multirow}
\usepackage{colortab}
\usepackage{colortbl}
\usepackage{arydshln}
\usepackage{algpseudocode}
\usepackage{tablefootnote}
\usepackage{amssymb}

\usepackage{newtxmath}
\definecolor{g}{rgb}{0.925, 0.957, 0.831} 
\definecolor{p}{rgb}{0.980,0.910,0.922}
\newcommand{\bhline}[1]{\noalign{\hrule height #1}}   
\newcommand{\ours}{$\beta_4$-UCS}
\newcommand{\ourrep}{FBR}

\usepackage{hyperref}
\usepackage{xcolor}
\hypersetup{
colorlinks=true,
linkcolor=[HTML]{B2001D},
citecolor=[HTML]{27408B},
urlcolor=[HTML]{27408B}
}
\usepackage{orcidlink}

\newlength{\basis} \newlength{\barlength}
  \newcommand*{\barchart}[2]{%
  \setlength{\barlength}{0.01\basis}\setlength{\barlength}{#1\barlength}%
  \textcolor{blue}{\rule{\barlength}{1.6ex}}%
	\setlength{\barlength}{0.01\basis}\setlength{\barlength}{#2\barlength}%
  \textcolor{red}{\rule{\barlength}{1.6ex}}&#1\%&#2\%}

\begin{document}
\title{Adapting Rule Representation With Four-Parameter Beta Distribution for Learning Classifier Systems}

\author{Hiroki Shiraishi$^{\orcidlink{0000-0001-8730-1276}}$,
Yohei Hayamizu$^{\orcidlink{0000-0003-1642-4919}}$,~\IEEEmembership{Graduate Student Member,~IEEE}, 
Tomonori Hashiyama$^{\orcidlink{0000-0001-7218-2999}}$,~\IEEEmembership{Member,~IEEE},\\
Keiki Takadama$^{\orcidlink{0009-0007-0916-5505}}$,~\IEEEmembership{Member,~IEEE},
Hisao Ishibuchi$^{\orcidlink{0000-0001-9186-6472}}$,~\IEEEmembership{Fellow,~IEEE}, 
and Masaya Nakata$^{\orcidlink{0000-0003-3428-7890}}$,~\IEEEmembership{Member,~IEEE}
\thanks{
Manuscript received 24 June 2024; revised 27 November 2024 and
1 February 2025; accepted 4 March 2025. 
This work was supported by JSPS KAKENHI (Grant Nos. JP23KJ0993, JP23K20388), National Natural Science Foundation of China (Grant No. 62376115), and Guangdong Provincial Key Laboratory (Grant No. 2020B121201001). (\textit{Corresponding authors: Hisao Ishibuchi and Masaya Nakata.})}
\thanks{Hiroki Shiraishi and Masaya Nakata are with the Faculty of Engineering, Yokohama National University, Yokohama 240-8501, Japan (e-mail: shiraishi-hiroki-yw@ynu.jp; nakata-masaya-tb@ynu.ac.jp).}
\thanks{Yohei Hayamizu is with the Department of Computer Science, the State University of New York at Binghamton, Binghamton, NY 13902, USA (e-mail: yhayami1@binghamton.edu).}
\thanks{Tomonori Hashiyama is with the Department of Informatics, the University of Electro-Communications, Tokyo 182-8585, Japan (e-mail: hashiyama.tomonori@uec.ac.jp).}
\thanks{Keiki Takadama is with the Information Technology Center, the University of Tokyo, Chiba 277-0882, Japan (e-mail: takadama@g.ecc.u-tokyo.ac.jp).}
\thanks{Hisao Ishibuchi is with the Department of Computer Science and Engineering, Southern University of Science and Technology, Shenzhen 518055, China (e-mail: hisao@sustech.edu.cn).}
\thanks{Digital Object Identifier 10.1109/TEVC.2025.3550915}
}


\markboth{IEEE TRANSACTIONS ON EVOLUTIONARY COMPUTATION, DOI: 10.1109/TEVC.2025.3550915, MARCH 2025}{SHIRAISHI \MakeLowercase{{\em et al.}}:~ADAPTING RULE REPRESENTATION FOR LEARNING CLASSIFIER SYSTEMS}
 
\maketitle
\begin{abstract}
Rule representations significantly influence the search capabilities and decision boundaries within the search space of Learning Classifier Systems (LCSs), a family of rule-based machine learning systems that evolve interpretable models through evolutionary processes. However, it is very difficult to choose an appropriate rule representation for each problem. Additionally, some problems {benefit from using different representations for different subspaces within the input space.} Thus, an adaptive mechanism is needed to choose an appropriate rule representation for each rule in LCSs. This {article} introduces a flexible rule representation using a four-parameter beta distribution and integrates it into a fuzzy-style LCS. {The four-parameter beta distribution can form various function shapes, and this flexibility enables our LCS to automatically select appropriate representations for different subspaces.} Our rule representation can represent crisp/fuzzy decision boundaries in various boundary shapes, such as rectangles and bells, by controlling four parameters{, compared to the standard representations such as trapezoidal ones.} {Leveraging this flexibility,} our LCS is designed to adapt the appropriate rule representation for each subspace. 
 Moreover, our LCS incorporates a generalization bias favoring crisp rules where feasible, enhancing model interpretability without compromising accuracy. 
 Experimental results on real-world classification tasks show that our LCS achieves significantly superior test accuracy and produces more compact rule sets. 
 Our implementation is available at \url{https://github.com/YNU-NakataLab/Beta4-UCS}. An extended abstract related to this work is available at \url{https://doi.org/10.36227/techrxiv.174900805.59801248/v1} \cite{shiraishi2025evolutionary}.
\end{abstract}

\newpage
\section{Introduction}
\label{sec: introduction}

\IEEEPARstart{D}{ividing} a complex input space into a number of ``easy-to-solve'' subspaces is a pervasive problem-solving principle in the machine learning field, such as data mining and image processing. This allows machine learning algorithms to solve a problem with simple logic, improving their performance and interpretability. Learning Classifier Systems (LCSs) \cite{urbanowicz2017introduction} are one of the most successful machine learning algorithms with such a problem-solving principle. The basic idea of LCSs is to solve a problem with simple condition-action rules by optimizing a set of rules using an evolutionary algorithm \cite{butz2006rule}. Each rule is responsible for an input subspace represented by its rule condition. LCSs attempt to produce an optimal set of rules that together cover the entire input space with a minimal number of accurate rules \cite{kovacs1998xcs}. Taking this feature, LCSs have been utilized as data mining tools and automated classifier design tools \cite{preen2021autoencoding}. 
{For instance, the sUpervised Classifier System (UCS) \cite{bernado2003accuracy}, a prominent classification-focused LCS, has recently been extended to survival analysis \cite{woodward2024survival}.}

For most LCSs including their applications, the design of appropriate rule-condition representations is essential to efficiently solve a problem. This is because the rule representation determines the shape and fuzziness of the subspace boundary{. The fuzziness of a subspace boundary indicates whether the matching degree transitions abruptly (crisp) or gradually (fuzzy) between matched and unmatched regions. The adequacy of this approach} depends on problem features, e.g., linear and noisy class boundaries \cite{shiraishi2022can}. Classical LCS studies often used {ternary} representations in the context of solving binary input problems. However, many application tasks involve real-valued inputs. To fill this gap, numerous efforts have been dedicated to developing various rule representations for real-valued inputs. The developed rule representations can be roughly summarized in terms of the shape and fuzziness of the subspace boundary as follows. 

\begin{itemize}
    \item \emph{Crisp-geometric-form.} In this form, the input space is divided using a pre-specified geometric shape with a crisp boundary. Note that the crisp boundary means that the matching degree of an input to each rule is zero or one. Frequently-used geometric shapes include hyperrectangle \cite{wilson2000mining} and hyperellipsoid \cite{butz2005kernel}. In general, the choice of a geometric shape depends on the shape of the class boundaries in each problem. This form can boost the LCS performance {(i.e., the classification accuracy on test data)} with a concise rule set if the input space is well divided using the selected geometric shape; otherwise, it tends to suffer from over-specification (i.e., creating a large number of rules covering small subspaces), which degrades the LCS performance \cite{shiraishi2022can}. 
     
    \item \emph{Crisp-flexible-form.} To reduce the dependence of the rule representation on problems, {several works have been proposed, e.g., gene expression programming \cite{wilson2008classifier} and neural network \cite{bull2002accuracy} based representations.} These approaches can flexibly adjust the subspace shape of each rule in an online manner, enhancing the performances of LCSs. However, the search space of rule structures becomes complicated, requiring many iterations to derive the optimum performance \cite{wilson2008classifier}. In addition, this form tends to suffer from overfitting especially when a problem includes noisy inputs and/or the number of available training patterns is small. 
    
    \item \emph{Fuzzy form.} 
    Unlike the crisp form, the fuzzy form divides the input space with fuzzy boundaries. Specifically, it uses fuzzy logic to express a matching degree of a rule to an input (i.e., the matching degree is not binary $\{0, 1\}$ but a real number in $[0, 1]$). An advantage of this form is that it can improve the robustness of noisy problems and prevent severe overfitting due to the nature of fuzzy boundaries. In addition, the shape and fuzziness of the subspace boundary can be flexibly customized by changing the type of membership functions {(MFs)} such as the triangular-shaped \cite{casillas2007fuzzy} and trapezoidal-shaped \cite{shoeleh2011towards} functions. However, as the crisp-flexible-form, the fuzzy form tends to require many iterations to perform well due to the complexity of the search space of rule structures \cite{shiraishi2023fuzzy}. Moreover, the use of appropriate {MFs} is crucial to boost the LCS performance.  
\end{itemize}
Note that a detailed survey for the existing rule representations is provided in Section \ref{sec: related work}. 

In general, relying solely on a particular rule representation may degrade the performance of LCSs \cite{orriols2008fuzzy}. However, an appropriate representation typically cannot be known in advance since we do not know the optimal class boundary for each problem. A possible solution for this limitation is to select a rule representation adaptively for each rule during the execution of an LCS algorithm. However, such adaptation techniques have not been considered sufficiently, probably due to the following challenges. First, using multiple representations exceedingly increases the search space of rule structures, degrading the efficiency of the evolutionary search. Second, a huge number of rules may be generated. This can cause the computational time of LCS to become prohibitively large. An efficient rule-deletion mechanism, such as subsumption deletion \cite{wilson1998generalization}, is essential. However, realizing this mechanism consistently for different representations is difficult. Third, the appropriate representation may be different for each subspace of the input space. {This challenge is particularly evident in real-world applications such as medical diagnosis, where some disease indicators require crisp thresholds (e.g., age), while others involve fuzzy boundaries (e.g., inherently noisy vital measurements such as heart rate) \cite{takadama2015extracting}.}

Based on these considerations, this {article} presents a novel rule representation that can flexibly represent crisp/fuzzy subspace boundaries and their various shapes while addressing the above challenges. Our main idea is to develop a fuzzy representation whose {MFs} can be changed adaptively. Specifically, we develop a four-parameter beta distribution-based representation and incorporate it into a supervised fuzzy-style LCS (Fuzzy-UCS) \cite{orriols2008fuzzy}. The four-parameter beta distribution is a popular probabilistic distribution used in Bayesian inference, and it can form various distribution shapes, such as bell, rectangular, and triangular shapes \cite{crackel2017bayesian}, by tuning its four parameters. Taking this advantage, our rule representation uses the four-parameter beta distribution as a {MF}, enabling it to represent various subspace boundaries. Since our representation is based on the same distribution, it can prevent the increase of the search space of rule structures, thus preventing the deterioration of the search efficiency of our LCS. In addition, we can design a subsumption deletion mechanism that can be used consistently for various subspace boundaries represented by the proposed representation. This enables our LCS to produce a compact set of rules and reduce the computational time while adapting the subspace boundary of each rule.

The main contributions of this {article} are as follows. 
\begin{itemize}
    \item To the best of our knowledge, this {article} is the first attempt that adapts rule representations in terms of the complexity of shapes and fuzziness simultaneously for LCSs. 
    \item A novel, flexible rule representation using the four-parameter beta distribution (termed \ourrep) is proposed. Furthermore, a Fuzzy-UCS which incorporates a genetic operator and a subsumption operator adapted for the \ourrep\ (termed \ours) is also proposed. 
    \item Comprehensive comparison results of various rule representations on classification tasks with real-world datasets are reported for the first time.
\end{itemize}

Note that the two recent related works \cite{shiraishi2022can,shiraishi2023fuzzy} attempted to adjust the shape and fuzziness independently. Specifically, the self-adaptive bivariate beta distribution-based representation \cite{shiraishi2022can} can adjust the shape of the subspaces, but this approach is restricted to the crisp form, degrading the LCS performance for classification tasks including uncertainty. The other work \cite{shiraishi2023fuzzy} has considered an adaptation mechanism of the crisp and fuzzy forms under the same shape of the subspaces.  
Different from these works, this {article} proposes a novel rule representation that adjusts the shape and fuzziness simultaneously, providing a significant performance improvement as demonstrated through computational experiments in Section \ref{sec: experiment}.

The rest of this {article} is organized as follows. Section \ref{sec: related work} provides a literature survey for rule representations. In Section \ref{sec: preliminaries}, the Fuzzy-UCS framework and definitions of the four-parameter beta distribution are described as preliminary knowledge. Section \ref{sec: proposed} explains \ours, which is the proposed algorithm with a novel rule representation based on the four-parameter beta distribution. Section \ref{sec: experiment} presents comparative experiments using 25 datasets. Section \ref{sec: analysis} provides analytical results to investigate the characteristics of \ours. Finally, Section \ref{sec: concluding remarks} {concludes this article.}\label{others-1}

\section{Related Work}
\label{sec: related work}
 {This section provides a comprehensive review of existing rule representations in LCSs. We first present an overview of various rule representations, and then categorize their key characteristics.}
Table \ref{tb: related_work} summarizes popular rule representations used in LCSs for real-valued inputs. In the table, each representation is roughly categorized in terms of the fuzziness and shape of subspace boundaries.  

\subsection{Crisp Representations}
\label{ss: crisp representations}

\begin{table*}[t]
    \caption{Categorization of Existing Rule Representations for Real-Valued Inputs. }
    \label{tb: related_work}
    \centering
    \resizebox{\textwidth}{!}{
    \begin{tabular}{r|c l }
    \bhline{1pt}
        Name & Fuzziness & Subspace shape (for Crisp) / Membership function (for Fuzzy) \\
        \bhline{1pt}
        Center-spread representation \cite{wilson1999xcsr}                                                &	 Crisp 	      & Hyperrectangle                          \\
        Ordered-bound representation \cite{wilson2000mining} \hspace{0.5mm}	                                          &	 Crisp 	      & Hyperrectangle                          \\
        Unordered-bound representation \cite{stone2003real} 	                                          &	 Crisp 	      & Hyperrectangle                          \\
        Min-percentage representation \cite{dam2005real} 	                                              &	 Crisp 	      & Hyperrectangle                          \\
        Neural network-based representation \cite{bull2002accuracy}                                  &	 Crisp 	      & \textit{Not limited to a specific shape}                                       \\
        Kernel-based representation \cite{butz2005kernel}	                                              &	 Crisp 	      & Hyperellipsoid                          \\
        Convex hulls-based representation \cite{lanzi2006using} 	                                      &	 Crisp 	      & Convex hull                             \\
        Gene expression programming-based representation \cite{wilson2008classifier} 	                  &	 Crisp        & \textit{Not limited to a specific shape}                                               \\
        Code fragment-based representation 	\cite{arif2017solving} 	                                      &	 Crisp        & \textit{Not limited to a specific shape}                                  \\
        Bivariate beta distribution-based representation \cite{shiraishi2022beta} 	                              &	 Crisp        & Curved polytope                \\
        Self-adaptive bivariate beta distribution-based representation	\cite{shiraishi2022can} \hspace{0.45mm}   &  Crisp        & Hyperrectangle, Curved polytope \\
        {Coevolved different knowledge representation \cite{llora2002coevolving}} & {Crisp} & {Hyperrectangle, Hyperpolygon, Complex nonlinear shape} \\
        {Adaptive discretization intervals representation \cite{bacardit2003evolving}} & {Crisp} & {Combination of hyperrectangles}\\
        Radial basis function network-based representation \cite{bull2002accuracy}                      &	 Fuzzy        & Symmetric bell                          \\
        Disjunctive normal form triangular-shaped membership functions \cite{casillas2007fuzzy} 	      &	 Fuzzy        & Triangle                                \\
        Non-grid-oriented triangular-shaped membership functions \cite{orriols2008approximate}            &	 Fuzzy        & Triangle                                \\
        Trapezoidal-shaped membership functions \cite{shoeleh2011towards}                                 &	 Fuzzy        & Trapezoid                               \\
        Normal distribution-shaped membership functions \cite{tadokoro2021xcs} 	                          &	 Fuzzy        & Symmetric bell                          \\
        Rectangular- and triangular-shaped membership functions \cite{shiraishi2023fuzzy}                 &	 Crisp, Fuzzy & Rectangle, Triangle                     \\
        Four-parameter beta distribution-based representation (Ours) \hspace{4.8mm}	                      &	 Crisp, Fuzzy & Rectangle, Triangle, Symmetric/Asymmetric bell, Monotonic \\
        \bhline{1pt}
    \end{tabular}
    }
    \vspace{-2mm}
\end{table*}

\begin{figure*}[!t]
    \centering
    \subfloat[\\Hyperrectangle]{\includegraphics[width=0.095\textwidth]{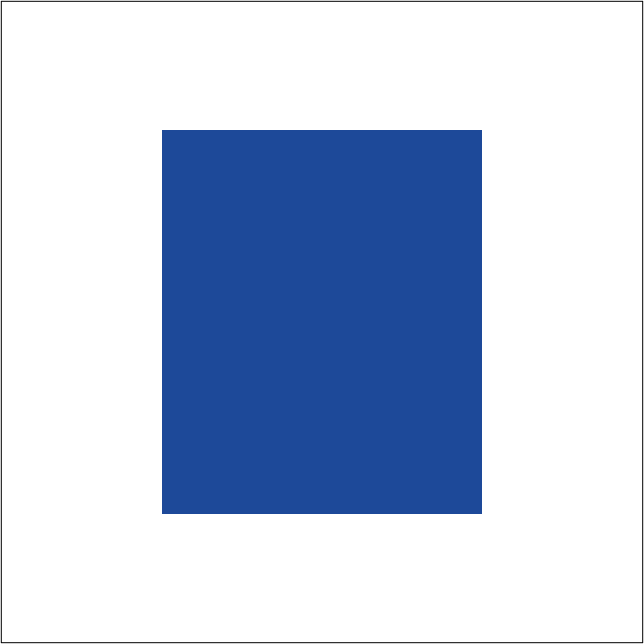}%
    \label{fig: rec}}\hfill
    \subfloat[\\Hyperellipsoid]{\includegraphics[width=0.095\textwidth]{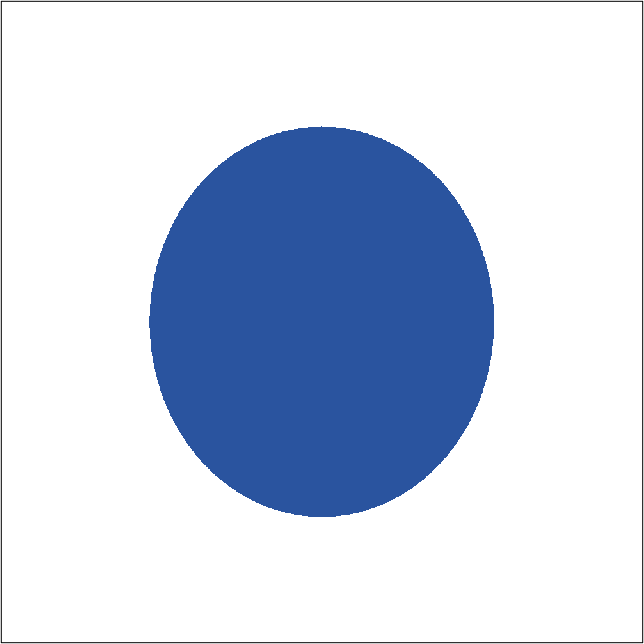}%
    \label{fig: elp}}\hfill
    \subfloat[Curved polytope]{\includegraphics[width=0.095\textwidth]{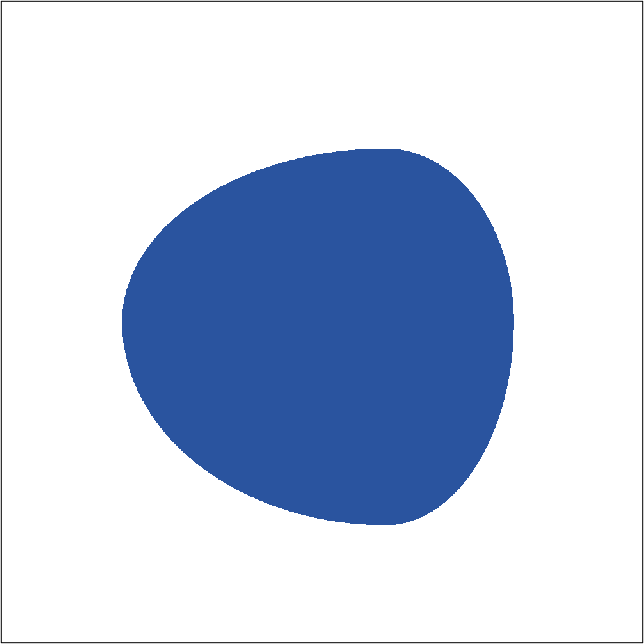}%
    \label{fig: eclipselike}}\hfill
    \subfloat[Convex hull]{\includegraphics[width=0.095\textwidth]{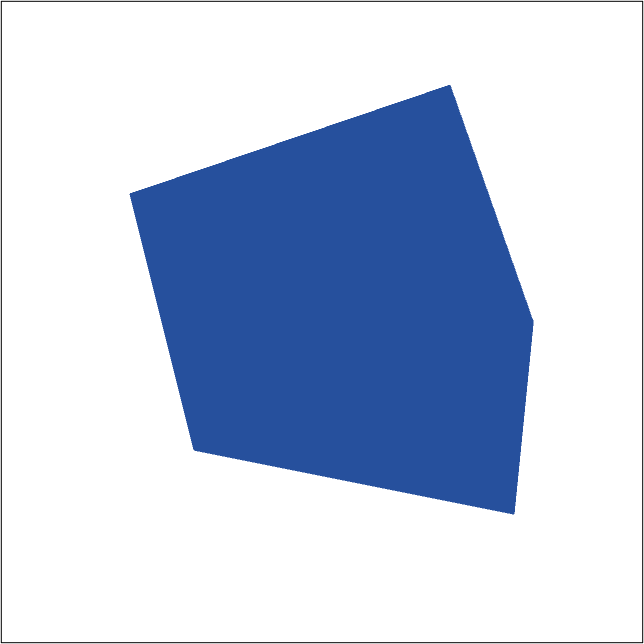}%
    \label{fig: con}}\hfill
    \subfloat[Triangular MF]{\includegraphics[width=0.11\textwidth]{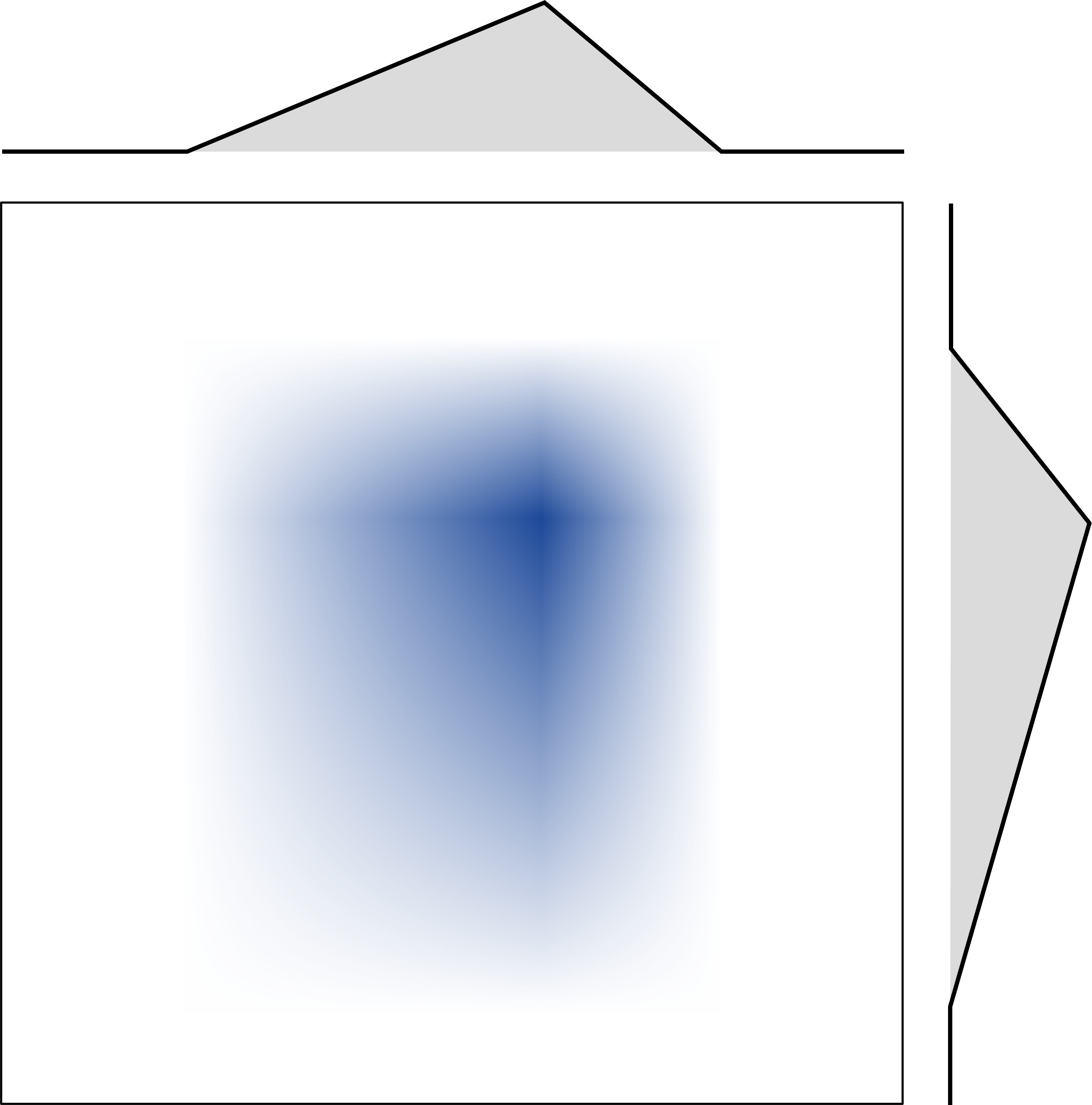}%
    \label{fig: tri}}\hfill
    \subfloat[Symmetric bell-shaped MF]{\includegraphics[width=0.11\textwidth]{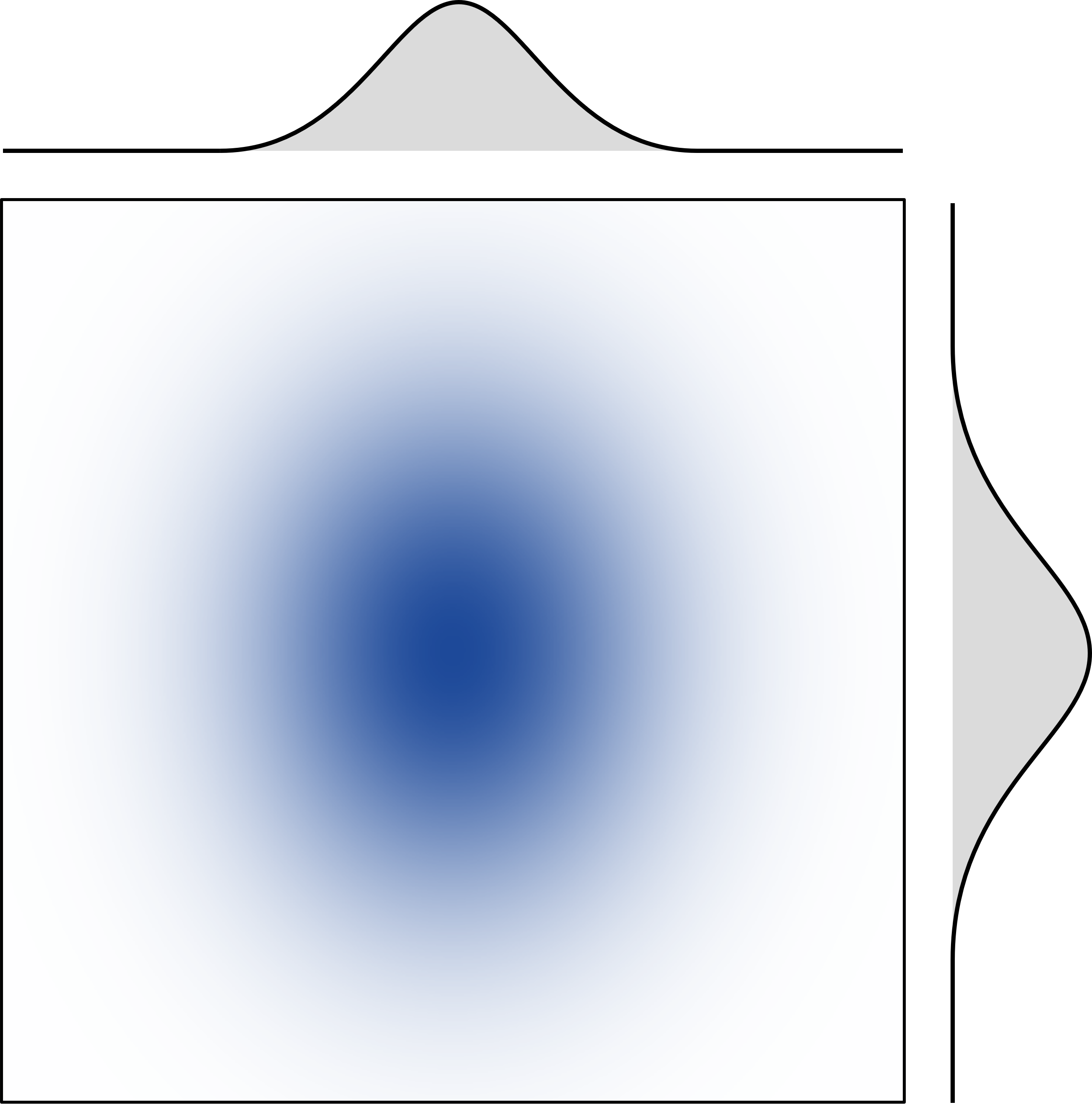}%
    \label{fig: fec}}\hfill
    \subfloat[Trapezoidal MF]{\includegraphics[width=0.11\textwidth]{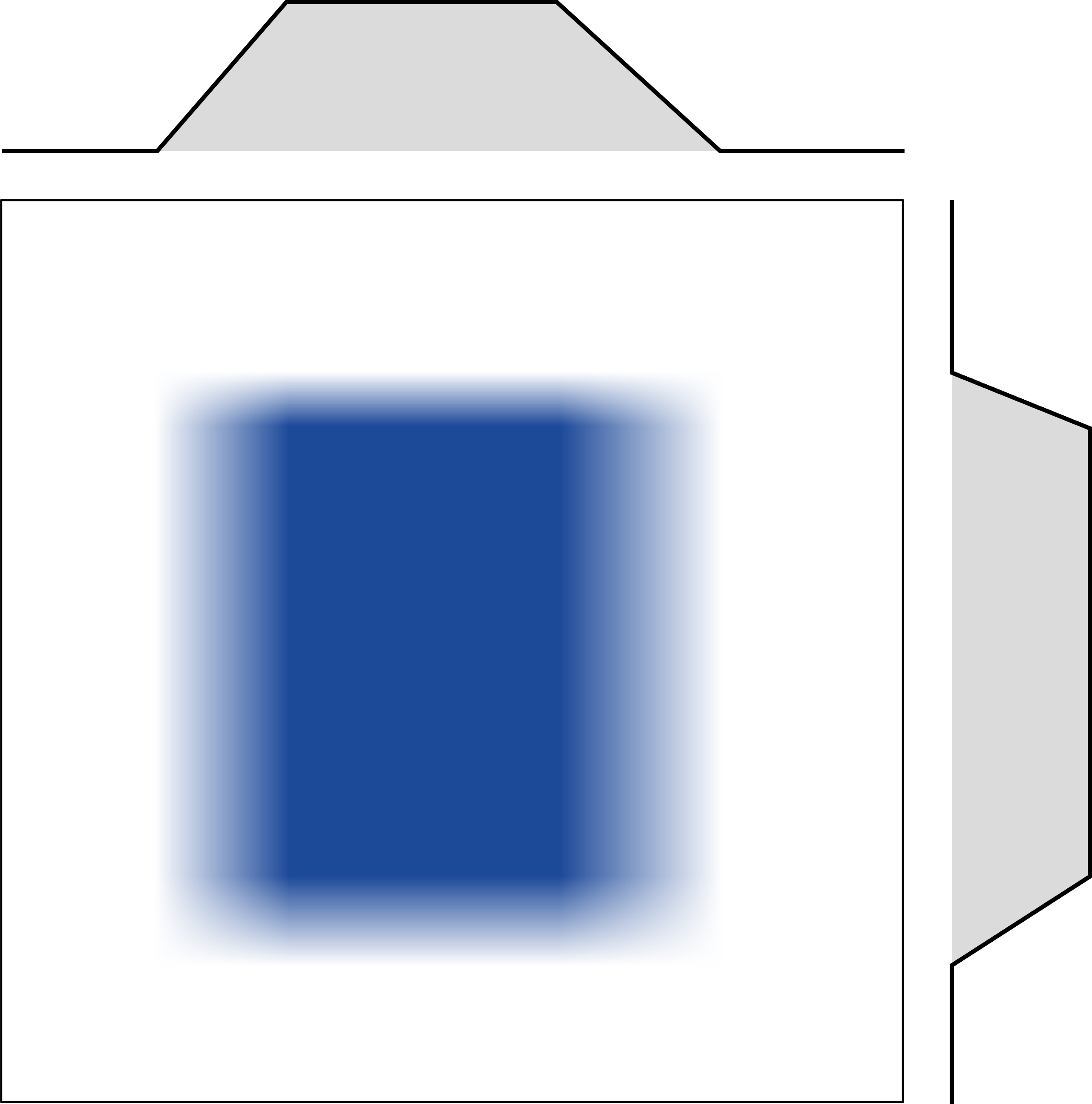}%
    \label{fig: trp}}\hfill
    \subfloat[Rectangular MF]{\includegraphics[width=0.11\textwidth]{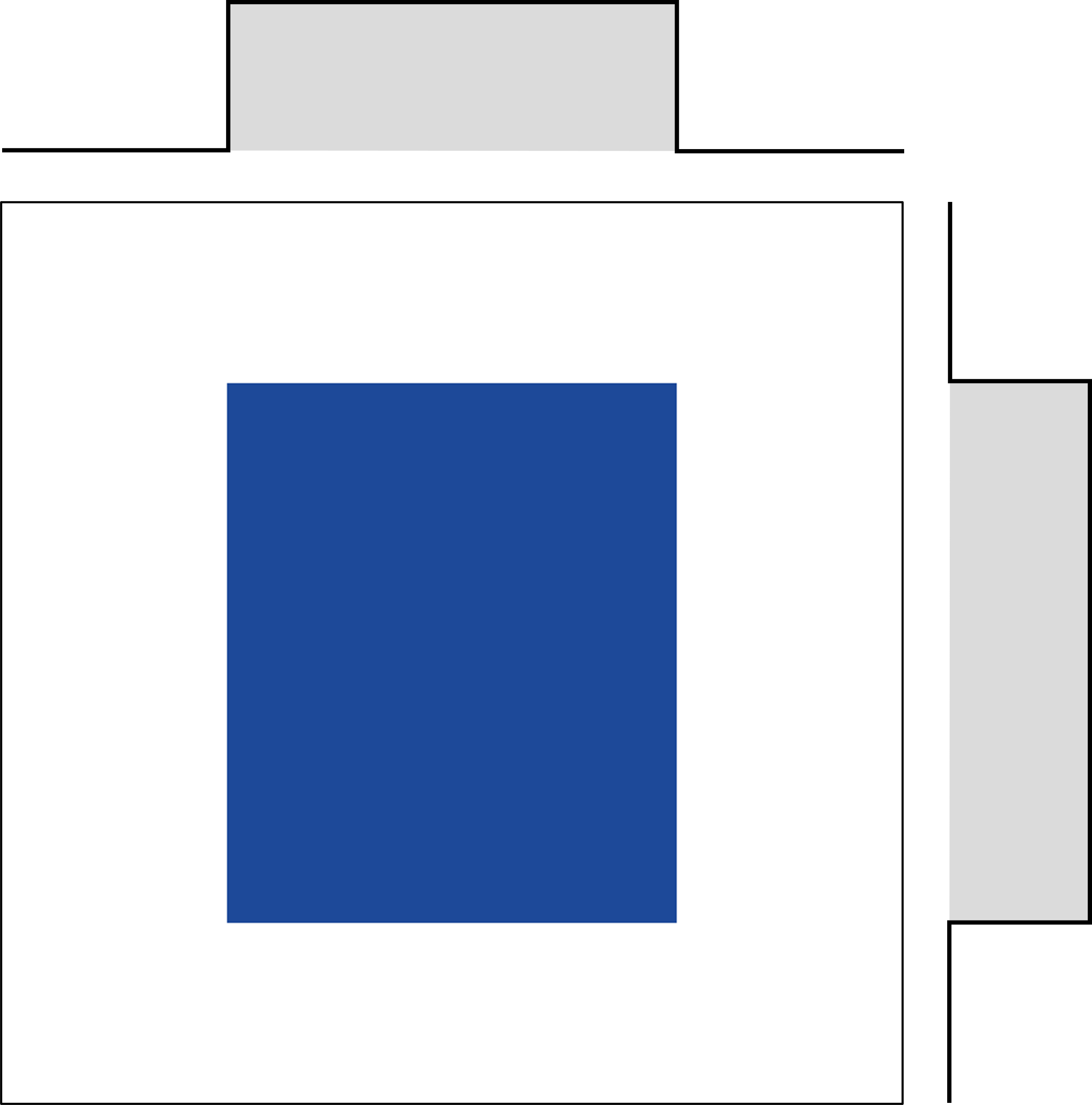}%
    \label{fig: fre}}\hfill
    \caption{Examples of the matching degree landscapes of existing rule representations for a two-dimensional input space. Blue gradients indicate matching degrees between 0 and 1, whereas blue corresponds to a matching degree of 1. MF denotes Membership Function.}
    \label{fig:exist_rule_landscape}
    \vspace{-2mm}
\end{figure*}

The most popular approach is to divide the input space with hyperrectangular shapes using crisp boundaries, as exampled in Fig.~\ref{fig:exist_rule_landscape}a. Various representations have been proposed for hyperrectangular shapes such as the center-spread \cite{wilson1999xcsr}, ordered-bound \cite{wilson2000mining}, unordered-bound \cite{stone2003real}, and min-percentage \cite{dam2005real} ones. These variants define a matching interval for each dimension of the input space in different semantics to form a hyperrectangular shape. For example, the ordered-bound representation uses lower/upper values to express an interval. Another popular approach is to utilize ellipsoid-like shapes. In \cite{butz2005kernel} and \cite{butz2008function}, a Gaussian kernel-based representation was proposed, which bounds a subspace with a hyperellipsoid shape, as exampled in Fig.~\ref{fig:exist_rule_landscape}b. The bivariate beta distribution-based representation \cite{shiraishi2022beta} adopts symmetric/asymmetric beta distribution-based kernels to represent different ellipsoid-like shapes, as exampled in Fig.~\ref{fig:exist_rule_landscape}c. In \cite{shiraishi2022can}, the bivariate beta distribution-based representation was extended to represent both hyperrectangular and hyperellipsoid-like shapes. A convex hull-based representation \cite{lanzi2006using} can form various polyhedrons by controlling their vertices (see Fig.~\ref{fig: con}). 

As more flexible representations, neural network-based representation was considered and incorporated into the XCS classifier system \cite{wilson1995xcs} (called X-NCS \cite{bull2002accuracy}). In this attempt, a rule condition was modeled as a small neural network that returns a matching degree to an input. In \cite{wilson2008classifier}, gene expression programming \cite{ferreira2006gene} was used to represent complex subspace boundaries. In recent years, the code fragment {(CF)} representation \cite{iqbal2013reusing} was extended to real-valued inputs \cite{arif2017solving}. In \cite{arif2017solving}, {a rule condition and/or action is composed of one or more binary decision trees where input features (including lower-upper intervals for continuous features) are assigned to each leaf node. This flexible representation allows CF-based LCSs to produce various decision boundaries beyond simple hyperrectangles, as any decision surface that can be encoded in tree form can be used. Furthermore, CF-based LCSs can adapt different representations to different subspaces during training.} These representations have proven to improve the LCS performance, but {they typically require 2-10 times more training iterations (compared to the hyperrectangular representation) to achieve comparable performance levels due to their more complex search spaces \cite{wilson2008classifier,arif2017solving}.}

{While the above representations are for Michigan-style LCSs, several studies explored adaptive crisp rule representations in Pittsburgh-style LCSs. In \cite{llora2002coevolving}, different knowledge representations were co-evolved, including instance sets (forming complex nonlinear shapes), orthogonal decision trees (hyperrectangles), and oblique decision trees (hyperpolygons). In \cite{bacardit2003evolving}, the adaptive discretization intervals representation was proposed to evolve multiple discretization intervals adaptively, forming combinations of hyperrectangles and allowing the evolutionary process to select appropriate discretizations for different rules and attributes.}

\subsection{Fuzzy Representations}
As mentioned in the previous section, the shape and fuzziness of the subspace boundary can be customized by changing the type of membership functions {(MFs)}. Thus, various {MFs} have been considered to boost the LCS performance. A triangular {MF} is a popular choice. Fuzzy-XCS \cite{casillas2007fuzzy} and Fuzzy-UCS \cite{orriols2008fuzzy} use this {MF} in the disjunctive and conjunctive normal forms. These forms represent the rule condition using a set of predefined linguistic terms, such as \emph{small}, \emph{medium}, and \emph{large}, providing interpretable rules. However, these forms need to design the vertices of a triangle in advance in order to define the linguistic terms, and thus, the adequacy of its design depends on the problem \cite{ishibuchi2005comparison}. To address this limitation, Fuzzy-UCS was modified by using the non-grid-oriented form in which the vertices of the triangular shapes can be adjusted for each rule (see Fig. \ref{fig: tri})\cite{orriols2008approximate,orriols2011fuzzy}. Another popular {MF} is symmetric bell-shaped (see Fig. \ref{fig: fec}). A fuzzy variant of X-NCS, called X-NFCS \cite{bull2002accuracy}, uses fuzzy neural network-based rules in which the {MF} is set to the Gaussian kernel. In \cite{tadokoro2021xcs}, a multivariate normal distribution was used as a {MF}. In \cite{shoeleh2011towards}, a trapezoidal {MF} was used to represent crisp and fuzzy regions simultaneously (see Fig. \ref{fig: trp}). To our knowledge, the adaptive selection of crisp/fuzzy forms was first attempted in Adaptive-UCS \cite{shiraishi2023fuzzy}. Specifically, it adaptively selects the rectangular and triangular {MFs} when generating a rule. Note that a rule condition with the rectangular {MF} represents a hyperrectangle with the crisp boundary, as shown in Fig.~\ref{fig: fre}. 

{It is important to note that the four-parameter beta distribution should not be confused with the \textit{beta membership function}, as they are fundamentally distinct concepts.}
The beta membership function was initially introduced in \cite{alimi1997beta} as a transfer function within beta basis function neural networks, which are structured as artificial feed-forward three-layer neural networks. Compared to Gaussian {MFs}, the beta membership functions offer enhanced linearity, asymmetry, and flexibility, and can provide more complex shapes \cite{alimi2003beta}. They have been successfully applied in neural network modeling \cite{bouaziz2013hybrid} and time series prediction \cite{baklouti2018beta}.

\section{Preliminaries} \label{sec: preliminaries}

\subsection{Fuzzy-UCS} \label{sec: Fuzzy-UCS}

{Fuzzy-UCS \cite{orriols2008fuzzy} extends the UCS framework by incorporating fuzzy logic to handle continuous inputs more effectively. Fuzzy-UCS has been widely employed as a base algorithm for fuzzy-style LCSs in solving classification tasks, as it combines the evolutionary learning capabilities of UCS with the flexibility of fuzzy logic in handling uncertainty.}
This subsection describes the framework of Fuzzy-UCS using the non-grid-oriented form \cite{orriols2008approximate}. Hereafter, this {article} deals with $d$-dimensional, $m$-class classification problems, where an input $\mathbf{x}$ is normalized as $\mathbf{x} \in [0,1]^d$, and a set of $m$ possible classes is denoted as $\mathcal{C}=\{c_i\}_{i=1}^m$.

\subsubsection{Fuzzy Rule}
Given an input $\mathbf{x}=(x_1, x_2, ..., x_d)$, a fuzzy rule, $k$, is described as:
\begin{flalign}
        {\rm \bf IF }\; x_1\; {\rm is}\; A^k_{1}\; \text{and}\;\cdots\; \text{and}\; x_d\; {\rm is}\; A_{d}^k\; {\rm \bf THEN}\; \hat{c}^k, 
    \label{eq: rule_repr}
\end{flalign}
where $A_{i}^k$ is a fuzzy set for $x_i$, and $\hat{c}^k$ is a {consequent} class.
For the non-grid-oriented form, $A_{i}^k$ consists of parameters that control the landscape of an employed membership function, $\mu_{A_{i}^k}$. \label{others-2}
The matching degree of $k$ for $\mathbf{x}$, $\mu^k(\mathbf{x})$, is defined as:
\begin{flalign}
    \mu^k(\mathbf{x})=\prod_{i=1}^d{\mu_{A^k_{i}}}(x_i), 
    \label{eq:mat}
\end{flalign}
where Fuzzy-UCS considers that $k$ can be matched to $\mathbf{x}$ if $\mu^k(\mathbf{x})>0$. For example, when using a non-grid-oriented triangular membership function, $\mu_{A_{i}^k}(x_i)$ is designed as:
\begin{flalign}
    \mu_{A_{i}^k} (x_i) = 
    \begin{cases}
        \displaystyle\frac{x_i-\alpha_{i}^k}{\beta_{i}^k-\alpha_{i}^k}   & \mbox{$\alpha_{i}^k< x_i \leq \beta_{i}^k$,} \vspace{1mm}\\
        \displaystyle\frac{\gamma_{i}^k-x_i}{\gamma_{i}^k-\beta_{i}^k}   & \mbox{$\beta_{i}^k< x_i \leq \gamma_{i}^k$,} \vspace{1mm}\\
        0                                & \mbox{otherwise.}\\
    \end{cases} 
    \label{eq: triangular membership}
\end{flalign}
The corresponding fuzzy set is denoted by its three parameters, i.e., $A_{i}^k=(\alpha_{i}^k, \beta_{i}^k, \gamma_{i}^k)$. {If $x_i$ contains a missing value, Fuzzy-UCS handles it by specifying $\mu_{{A^k_i}}(x_i)=1$.}

Moreover, a rule has the following main variables; a fitness, $F^k$, which represents the soundness of {supporting} the class $\hat{c}^k$; a correct matching array, $\mathbf{cm}^k=\{{\rm cm}_{i}^k\}_{i=1}^{m}$, where ${\rm cm}_{i}^k$ indicates a certainty grade {for} class $c_i$ and is used to calculate {$F^k$}; \label{others-3}
an experience, ${\rm exp}^k$, which calculates the accumulated contribution in classifying training samples; and a numerosity, ${\rm num}^k$, which is the number of subsumed rules.

\subsubsection{Training Phase} 

At the initial iteration $t=0$, the population, $[P]$, is initialized as an empty set, and $t$ is updated as $t\gets t+1$. 
{At iteration $t$, the system receives an input $\mathbf{x}$ with its correct class $c^*$. The system then} builds a match set, $[M]$, which contains rules matched to $\mathbf{x}$. That is, $[M] = \{k \in [P] \mid \mu^{k}(\mathbf{x}) > 0\}$. Subsequently, it builds a correct set, $[C]$, which contains rules having the correct class, i.e., $[C] = \{k \in [M] \mid \hat{c}^k = c^*\}$. If $[C]$ does not contain rules having a large matching degree, which is defined as $\sum_{k\in{ [C]}}{\mu^{k}}(\mathbf{x})<1$, the covering operator performs to generate a new rule, $k_\text{cov}$, having 
{$A_i^{k_\text{cov}}=(\alpha_i^{k_\text{cov}},\beta_i^{k_\text{cov}},\gamma_i^{k_\text{cov}})=(\mathcal{U}{[-0.5, x_i)}, x_i, \mathcal{U}{(x_i,1.5]})$ for $i = 1, 2, \cdots, d$,}
$\mu^{k_\text{cov}}(\mathbf{x}) =1$, and $\hat{c}^{k_\text{cov}}=c^*$; and $k_\text{cov}$ is then added to $[M]$ and $[C]$. 

Next, the update process of rule variables is conducted. For each rule $k$ in $[M]$, the experience and the correct matching value for each class are updated as:
\begin{flalign}
    {\rm exp}^k &\gets {\rm exp}^k+\mu^{k}(\mathbf{x}),\label{eq: update_experience}\\
    {\rm cm}_{i}^k & \gets
    \begin{cases}
        {\rm cm}_{i}^k + \mu^{k}(\mathbf{x})   &   \text{if } c_i=c^*,\\
        {\rm cm}_{i}^k & \text{otherwise,}
    \end{cases}
\end{flalign}
for $i = 1, 2, \cdots, m$. Thus, ${\rm cm}_{i}^k$ indicates a certainty grade that $k$ {supports} $c_i$. In other words, $k$ becomes more reliable if it is modified to {support} the class that has the maximum certainty grade. Accordingly, the {consequent} class, \( \hat{c}^k \), is modified to \( c_{i_{\max}} \) if \( \hat{c}^k \neq c_{i_{\max}} \), where \( c_{i_{\max}} \) is the class having the maximum value, \( {\rm cm}_{i_{\max}}^k \), in $\mathbf{cm}^k$. The fitness is calculated as the soundness of {the consequent class}, $\hat{c}^k$, given by:
\begin{equation}
    \label{eq: update_fitness}
        F^k = \frac{{\rm cm}^k_{i_{\max}} - \sum_{i\mid i\neq{\rm max}}{{\rm cm}_{i}^k}}{{\rm exp}^k}.
\end{equation}
{This fitness calculation is equivalent to the penalized certainty factor used in \cite{ishibuchi2005rule}.
}

Finally, a steady-state genetic algorithm (GA) is applied to $[C]$ to discover more promising rules. To prevent a problematic cover-delete cycle \cite{butz2004toward}, the GA is executed only when the average of the last time the GA was applied to rules in $[C]$ exceeds a threshold defined by $\theta_\text{GA}$ \cite{wilson1995xcs}. The system selects two parent rules in $[C]$ and initializes two offspring rules, $k$ and $q$, as duplicates of the corresponding parent rules. The crossover and mutation operators are applied to $k$ and $q$. For example, when using a uniform crossover operator with probability $\chi$, the fuzzy sets for the $i$th input in the two parent rules, $A_{i}^k$ and $A_{i}^q$, are swapped with the probability 0.5. For the mutation operator, each parameter contained in $A_{i}^k$ is mutated with probability $p_\text{mut}$. The two offspring rules are inserted into $[P]$ and two rules are deleted if the number of rules in $[P]$ exceeds the maximum population size $N$.

Note that the subsumption operator \cite{wilson1998generalization} may be employed to promote the generalization of rules.{\footnote{To maintain an optimal balance between rule generalization and specialization, the absumption operator \cite{shiraishi2022absumption} may be used in addition to the subsumption operator. The absumption operator helps improve accuracy by eliminating over-general rules.}} The operator is executed after $[C]$ is formed or after GA is executed, and is referred to as \textit{Correct Set Subsumption} and \textit{GA Subsumption}, respectively. Suppose that we have two rules $k_\text{sub}$ and $k_\text{tos}$, where {$k_\text{sub}$} is a sufficiently-updated and accurate rule that satisfies ${\rm exp}^{k_\text{sub}} > \theta_\text{sub}$ and $F^{k_\text{sub}}>F_0$. If $k_\text{sub}$ has a more general condition than $k_\text{tos}$, $k_\text{tos}$ is subsumed to $k_\text{sub}$ as follows: the numerosity of ${k_\text{sub}}$, ${{\rm num}}^{k_\text{sub}}$, is increased by ${{\rm num}}^{k_\text{tos}}$ and $k_\text{tos}$ is removed from $[P]$.

The above procedure is repeated for the pre-specified maximum number of iterations. 

\subsubsection{Test Phase} 
The system decides a class $\hat{c}_{[P]}$ based on a voting process with the obtained population. Specifically, $\hat{c}_{[P]}$ for a test input $\bm x$ is decided {by selecting the class $c\in\mathcal{C}$ that maximizes the weighted sum of matching rules:}
\begin{flalign}
    \hat{c}_{[P]} = \arg\max_{c \in \mathcal{C}}\; \sum_{k\in [M_c]} F^k \cdot \mu^{k}(\mathbf{x})\cdot {\rm num}^k,
\end{flalign}
where {$[M_c]\subseteq[P]$ contains sufficiently updated rules that match $\mathbf{x}$ and support class $c$}, that is:
\begin{flalign}
    [M_c]:= \{k\in[P] \mid {\rm exp}^k > \theta_\text{exp} \land \mu^{k}(\mathbf{x}) > 0 \land \hat{c}^k = c\},
\end{flalign}
where $\theta_\text{exp}$ defines the minimum experience that rules have to participate in the voting process. 

\subsection{Four-Parameter Beta Distribution} \label{sec:beta}

\begin{figure*}[!t]
\centering
\subfloat[$\alpha = \beta=1$]{\includegraphics[width=0.249\textwidth]{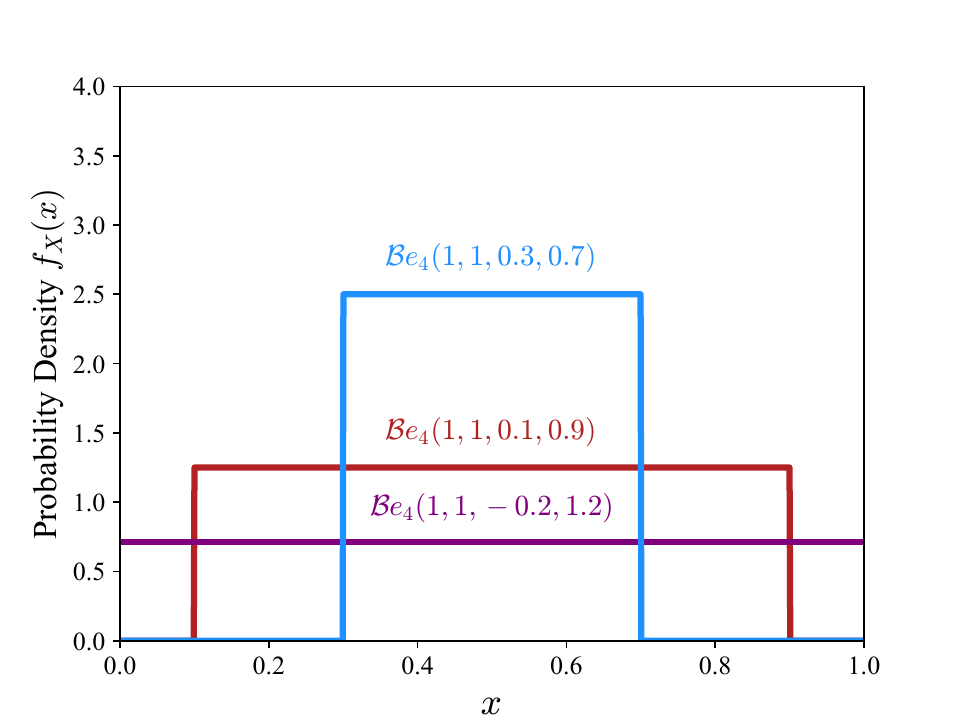} \label{fig: beta-rect}}
\subfloat[$\alpha, \beta>1$]{\includegraphics[width=0.249\textwidth]{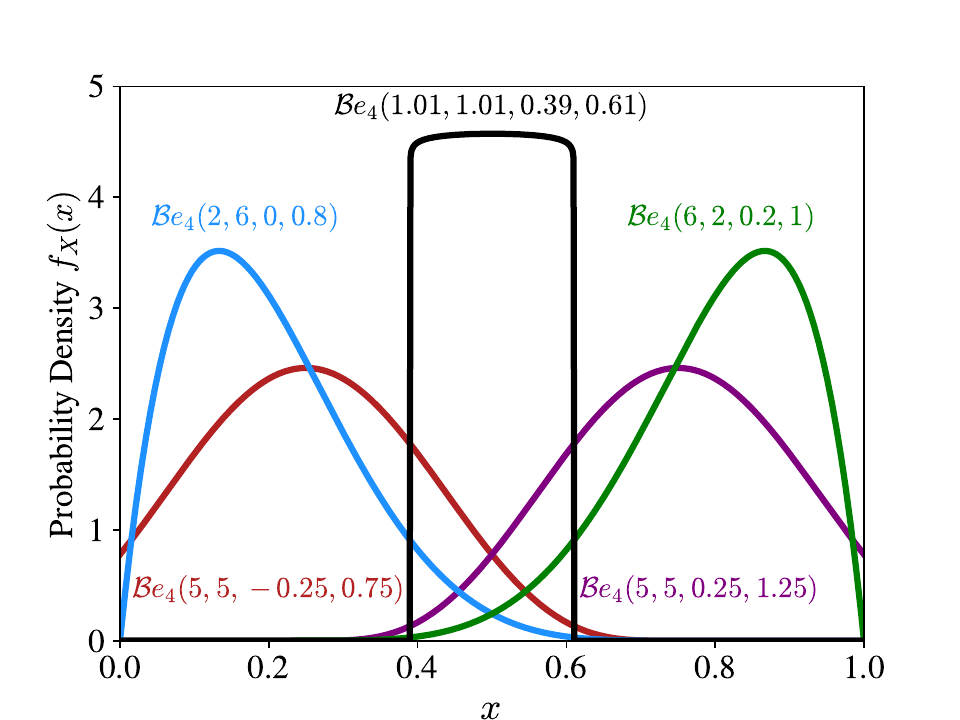} \label{fig: beta-bell}}
\subfloat[$\alpha, \beta<1$]{\includegraphics[width=0.249\textwidth]{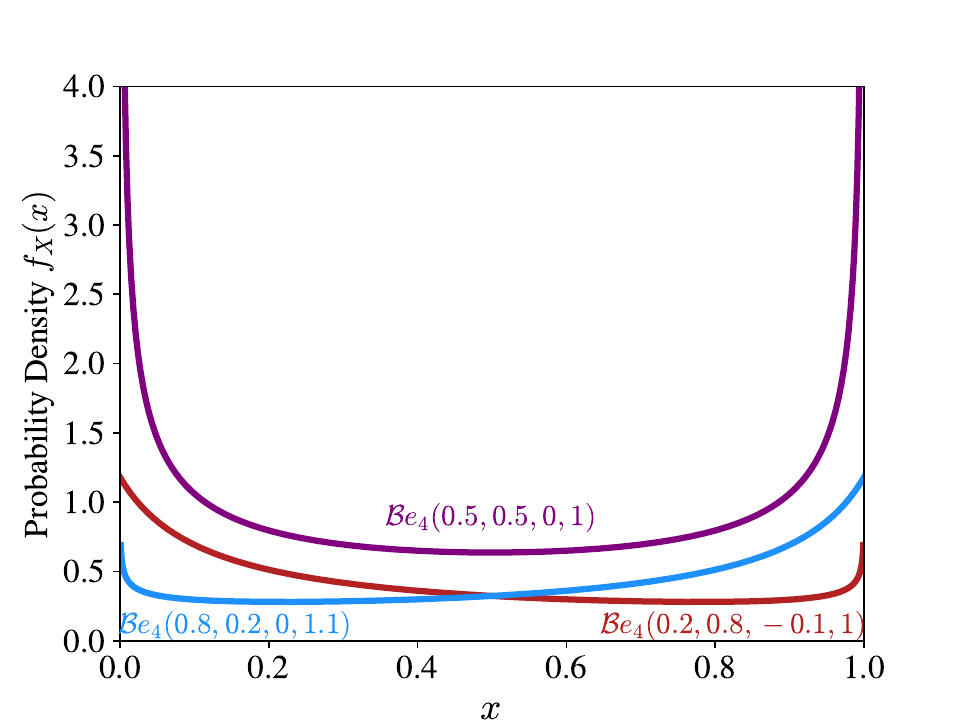} \label{fig: beta-U}}
\subfloat[$\{\alpha=1, \beta\neq 1\}$, $\{\alpha \neq 1, \beta=1\}$]{\includegraphics[width=0.249\textwidth]{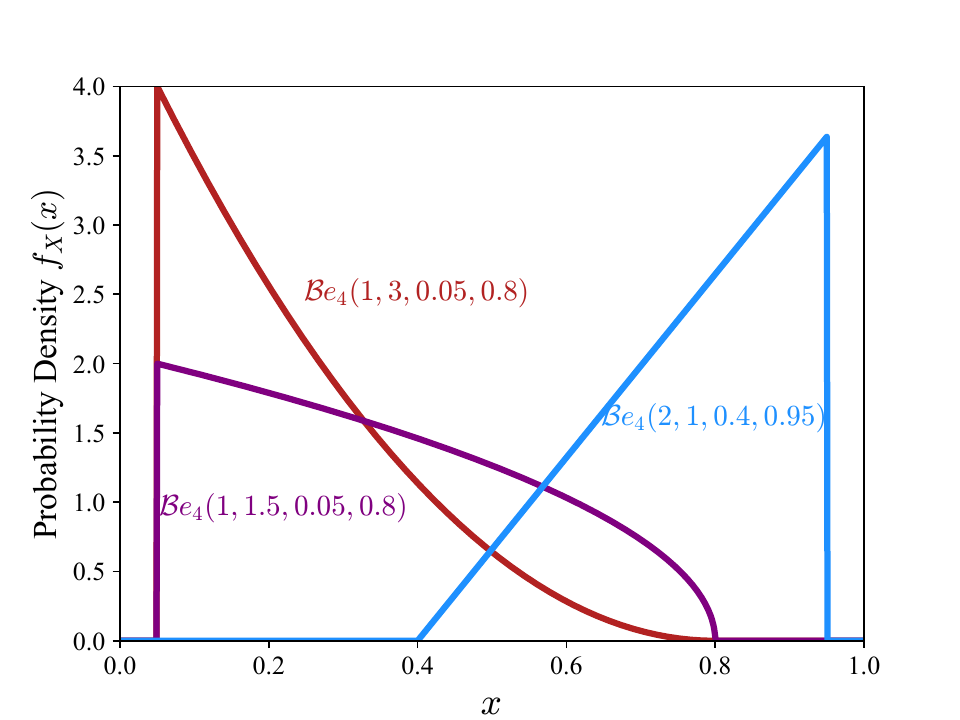} \label{fig: beta-mono}}
\caption{Examples of the PDF of the four-parameter beta distribution,  $\mathcal{B}e_4(\alpha, \beta, l, u)$. }
\label{fig: all_beta}
\vspace{-2mm}
\end{figure*}

A beta distribution is a continuous probability distribution. Its probability density function (PDF) can represent a wide variety of shapes such as linear, bell, and U shapes by controlling two shape parameters $\alpha$ and $\beta$. While the standard (i.e., bivariate) beta distribution is defined in the unit interval $[0,1]$, the four-parameter beta distribution, $\mathcal{B}e_4(\alpha, \beta, l, u)$, can be defined in any interval $[l,u]$. Specifically, the PDF of $\mathcal{B}e_4(\alpha, \beta, l, u)$ is defined as: 
\begin{flalign}
    f_X(x;\alpha, \beta, l, u)=
    \begin{dcases}
    \frac{(x-l)^{\alpha-1}(u-x)^{\beta-1}}{B(\alpha, \beta)(u-l)^{\alpha+\beta-1}} &\text{if}\; l\leq x\leq u,\\
    0&\text{otherwise,}
    \end{dcases}
    \label{eq:beta_pdf}
\end{flalign}
where $\alpha, \beta>0$; and $B(\alpha, \beta)$ is the beta function, given by:
\begin{flalign}
    B(\alpha, \beta)=\int_{0}^{1}t^{\alpha-1}(1-t)^{\beta-1}dt. 
\end{flalign}
Moreover, some important statistics of $\mathcal{B}e_4(\alpha, \beta, l, u)$, that is, mode and kurtosis are obtained as:
\begin{flalign}
    &{\rm mode}=
    \begin{dcases}
        l+\frac{\alpha-1}{\alpha+\beta-2}(u-l)  &\text{if } \alpha, \beta>1, \\
        \text{any value in } [l,u]  &\text{if } \alpha=1, \beta=1,\\
        \{l,u\}  &\text{if } \alpha, \beta < 1,\\
        l  &\text{if }\alpha \leq 1, \beta > 1, \\
        u  &\text{if }\alpha>1, \beta\leq 1,\\
    \end{dcases}\label{eq:beta-mode}\\\vspace{3mm}
    &{\rm kurtosis} = \frac{3(\alpha+\beta+1)[2(\alpha+\beta)^2+\alpha\beta(\alpha+\beta-6)]}{\alpha\beta(\alpha+\beta+2)(\alpha+\beta+3)}.\label{eq:beta-kurtosis}
\end{flalign}
As shown in Fig. \ref{fig: all_beta}, 
the PDF can form various shapes on the defined interval $[l, u]$, which can be roughly summarized as
follows:
\begin{itemize}
    \item The PDF becomes rectangular, bell-shaped, and U-shaped functions when $\{\alpha=\beta=1\}$, $\{\alpha, \beta>1\}$, and $\{\alpha, \beta<1\}$, as exampled in Figs.~\ref{fig: beta-rect}, \ref{fig: beta-bell}, and \ref{fig: beta-U}, respectively. Note that the maximum value of the PDF diverges to infinity for the U-shaped functions. Moreover, when $\{\alpha=1, \beta\neq 1\}$ and $\{\alpha \neq 1, \beta=1\}$, the PDF becomes a monotonic function for $[l, u]$, as exampled in Fig.~\ref{fig: beta-mono}. \label{others-4}
    \item Furthermore, the symmetry of the PDF can be controlled by the equivalence of $\alpha$ and $\beta$. With $\alpha = \beta$, the PDF can be a symmetric function, as exampled in the curve of $\mathcal{B}e_4(1.01, 1.01, 0.39, 0.69)$ of Fig.~\ref{fig: beta-bell}. With $\alpha \neq \beta$, it can be an asymmetric function. Specifically, the settings of $\alpha < \beta$ and $\alpha > \beta$ form left- and right-skewed functions, as exampled in $\mathcal{B}e_4(2, 6, 0, 0.8)$ and $\mathcal{B}e_4(6, 2, 0.2, 1)$ of Fig.~\ref{fig: beta-bell}, respectively. 
\end{itemize}

\section{\ours}\label{sec: proposed}

This section introduces \ours, which optimizes fuzzy rules with a novel rule representation. The main characteristics of \ours\ are three-fold; 

\begin{itemize}
    \setlength{\itemsep}{1mm}
    \item \emph{Various membership function choices.} In our representation, the membership function {(MF)} of a fuzzy rule is set to the PDF of the four-parameter beta distribution. This allows \ours\ to produce crisp and fuzzy rules by using a variety of {MFs} such as rectangular and symmetric/asymmetric bell-shaped functions. In addition, \ours\ is designed to find an appropriate {MF} for each subspace of the input space by inheriting the UCS's well-established rule discovery capability. 
    
    \item \emph{Compact search space.} Despite its high representability, our representation requires only four parameters to optimize, i.e., $\alpha, \beta, l$ and $u$, for each input dimension. This is an acceptable number of parameters for fuzzy representations as the triangular and trapezoidal {MFs} require three and four parameters, respectively. Thus, \ours\ can use different {MFs} 
    {without expanding the search space severely.}
    
    \item \emph{Generalization towards simplified rules.} Many practical LCS applications often use crisp-rectangular representations due to their interpretability and the capability of producing a concise rule set. Accordingly, \ours\ is designed to possess a generalization bias to produce crisp-rectangular rules {(see Sections \ref{ss: subsumption operator} and \ref{ss: crispification operator} for more details)}. In other words, it seeks to identify an input subspace where a simple classification logic exists. In this way, \ours\ balances the performance and the interpretability of rules.  
\end{itemize}
The remainder of this section explains a novel rule representation, termed \ourrep. Subsequently, the \ours\ framework is described.

\subsection{Four-Parameter Beta Distribution-Based Representation}

\begin{figure}[!t]
    \centering
    \includegraphics[width=0.495\textwidth]{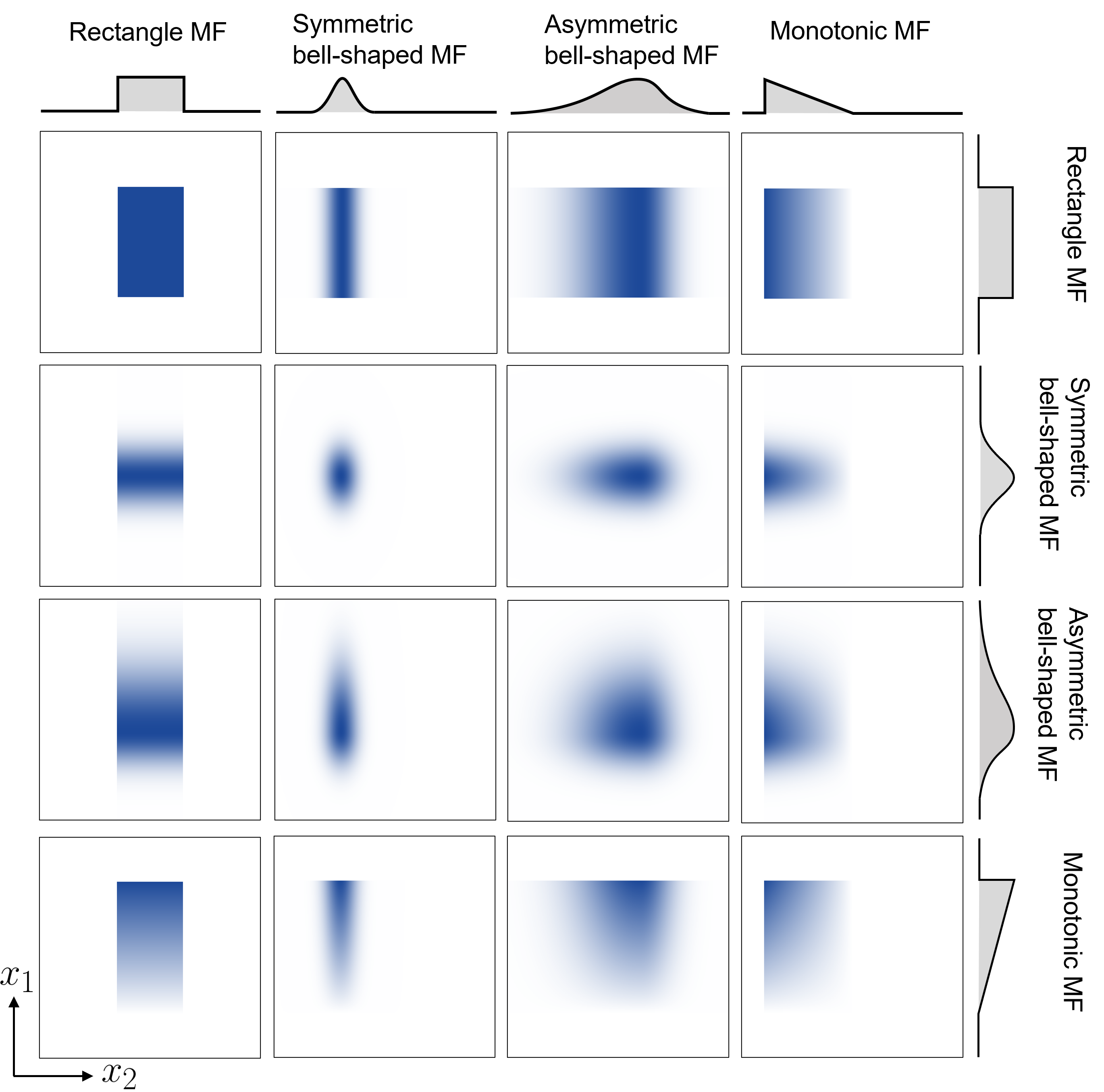}%
    \caption{Examples of the matching degree landscapes of \ourrep\ for a two-dimensional input space. Blue gradients indicate matching degrees between 0 and 1, whereas blue corresponds to a matching degree of 1.}
    \label{fig:tbr_mf}
    \vspace{-2mm}
\end{figure}

The four-parameter beta distribution-based representation (\ourrep) is used for each fuzzy set in our fuzzy rules. For a fuzzy rule $k$, we only change its fuzzy set as ${A_{i}^k} = (\alpha_{i}^k, \beta_{i}^k, l_{i}^k, u_{i}^k)$ where $(\alpha_{i}^k, \beta_{i}^k)$ and $(l_{i}^k, u_{i}^k)$ correspond to the shape and interval parameters of the four-parameter beta distribution, respectively. 

To use the PDF of the four-parameter beta distribution in \eqref{eq:beta_pdf} as a {MF}, the following two modifications are made. First, since the values of the PDF can be greater than 1, we apply the min-max normalization method {to those. This normalization ensures {MF} outputs remain bounded between 0 and 1 while preserving the distribution's relative shape.} Second, different from the original range of the shape parameters, we restrict them as $\alpha_{i}^k, \beta_{i}^k\geq 1$, meaning that the {MF} never becomes a U-shaped function. This is because the maximum value of any U-shaped function diverges to infinity, making its normalization impossible. 

Accordingly, we design the {MF} as follows:
\begin{flalign}  \label{eq:proposed_membership}
    {\mu_{A_{i}^k}}(x_i)=\frac{f_X(x_i; \alpha_{i}^k, \beta_{i}^k, l_{i}^k, u_{i}^k)}{f_X({\rm mode}_i^k; \alpha_{i}^k, \beta_{i}^k, l_{i}^k, u_{i}^k)},
\end{flalign}
where $f_X$ is the PDF given by \eqref{eq:beta_pdf}; and ${\rm mode}_i^k$ is the mode value of $f_X(x_i; \alpha_{i}^k, \beta_{i}^k, l_{i}^k, u_{i}^k)$, obtained by:
\begin{flalign}
    {\rm mode}_i^k=
    \begin{dcases}
        \text{any value in } [l_{i}^k, u_{i}^k]  &\text{if } \alpha_{i}^k=\beta_{i}^k=1,\\
        l+\frac{\alpha_{i}^k-1}{\alpha_{i}^k+\beta_{i}^k-2}(u_{i}^k-l_{i}^k)  &\text{otherwise,}
    \end{dcases}\label{eq:ours-mode}
\end{flalign}
which is much simpler than \eqref{eq:beta-mode} since $\alpha_{i}^k, \beta_{i}^k \geq 1$. Note that the matching degree, $\mu^k(\mathbf{x})$, is calculated by multiplying ${\mu_{A_{i}^k}}(x_i)$ for $i=1, 2, \cdots, d$, as designed in \eqref{eq:mat}. Fig.~\ref{fig:tbr_mf} shows some examples of the matching degree for the case of the two-dimensional input space. As shown in Fig. 3, both crisp and fuzzy matching degrees can be represented by the proposed representation method. An appropriate {MF} can be assigned to each fuzzy set in the proposed algorithm. 

In summary, \ourrep\ can represent rectangular, symmetric/asymmetric bell-shaped, and monotonic {MFs} by tuning the fuzzy sets. Furthermore, the landscape of the {MF} can differ for each input dimension. This flexibility enables \ours\ to use various matching degree landscapes. Note that \ourrep\ can represent a crisp and hyperrectangular boundary when $\alpha_i^k=\beta_i^k=1$ for all input dimensions.

\subsection{\ours\ Framework}
\label{ss: ours framework}

\begin{figure*}[!t]
    \centering
    \includegraphics[width=0.95\textwidth]{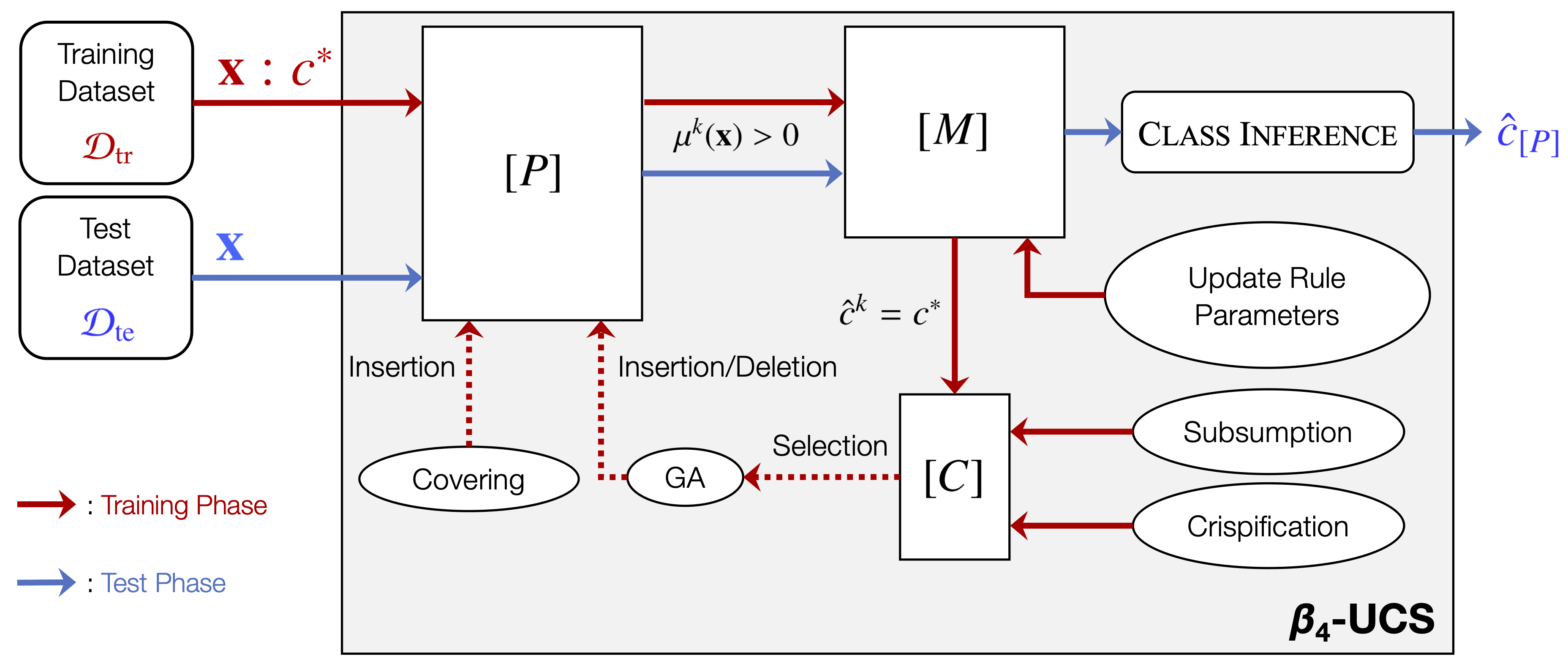}%
    \caption{Schematic illustration of \ours. The run cycle depends on the type of run: training or test. Upon receiving each data point $\mathbf{x}$, operations indicated by solid arrows are always performed, while operations indicated by dashed arrows (i.e., covering and GA) are executed only when specific conditions are met.}
    \label{fig:beta4-ucs}
    \vspace{-2mm}
\end{figure*}

Fig. \ref{fig:beta4-ucs} schematically illustrates \ours. Algorithm \ref{alg: ours training} describes the whole training procedure of \ours.
\ours\ performs in the same manner as Fuzzy-UCS, except for the covering, genetic, and subsumption operators. These operators are redesigned for \ourrep, which are described below. Furthermore, as a \ourrep-specific operation, a \textit{crispification} operator can be used if necessary. 
\begin{algorithm}[t]
\footnotesize
    \caption{{\ours\ training phase}}
    \label{alg: ours training}
    \begin{algorithmic}[1]
        \State Initialize the iteration $t$ as $t\leftarrow 0$;
        \State Initialize the population ${[P]}$ as ${[P]}\leftarrow \emptyset$;
         \While{{${t<}\text{\ the maximum number of iterations}$}}
        \State Update $t$ as $t \leftarrow t + 1$;
        \State Observe a training sample $\mathbf{x}$;
                    \State Create match set $[M] \coloneqq \{k \in {[P]} \mid \mu^k(\mathbf{x}) > 0\}$;

            \State Observe correct class $c^*\in\mathcal{C}$ associated with $\mathbf{x}$;
            \State Create correct set ${[C]}\coloneqq\{k\in{[M]} \mid \hat{c}^k=c^*\}$;
            \If{$\sum_{k\in{[C]}}\mu^{k}(\mathbf{x})<1$}
                \State Do covering; \hspace{\fill} $\triangleright$ see Section \ref{ss: covering}
            \EndIf
            
            \State Update ${{\rm exp}}^k$, $\mathbf{cm}^k$, and $F^k$ for $\forall k\in{[M]}$ as in \eqref{eq: update_experience}-\eqref{eq: update_fitness};
            \State Do correct set subsumption if necessary; \hspace{\fill} $\triangleright$ see Section \ref{ss: subsumption operator}
            \State Do correct set crispification if necessary; \hspace{\fill} $\triangleright$ see Section \ref{ss: crispification operator}
            \If{the average of the last time GA was applied to rules in $[C]>\theta_\text{GA}$}
            \State Run GA on $[C]$; \hspace{\fill} $\triangleright$ see Section \ref{ss: genetic operators}
            \State Do GA subsumption if necessary; \hspace{\fill} $\triangleright$ see Section \ref{ss: subsumption operator}
            \EndIf
            \EndWhile

    \end{algorithmic}
\end{algorithm}
\subsubsection{Covering Operator} \label{ss: covering}

Since \ours\ is motivated to cover the input space with as many crisp rules as possible, our covering operator is designed to generate initial rules in the crisp form. Specifically, after the construction of $[M]$ and $[C]$ for a training sample $\mathbf{x}$, the covering operator is executed when $\sum_{k\in{ [C]}}{\mu^{k}}(\mathbf{x})<1$, as designed in Fuzzy-UCS. The covering rule, $k_{\text{cov}}$, is then initialized as having the rectangular {MFs} with a randomly determined interval. 
That is, 
\begin{equation}
    \label{eq: beta covering}
    \begin{cases}
        \alpha^{k_{\text{cov}}}_{i} &= 1,\\
        \beta^{k_{\text{cov}}}_{i} &= 1,\\
        l^{k_{\text{cov}}}_{i} &= x_i - d,\\
        u^{k_{\text{cov}}}_{i} &= x_i + d,\\
    \end{cases} 
    \end{equation}
where $d$ is a random value sampled from a predefined range of $(0, r_0]$ with a hyperparameter $r_0$. The other rule variables, ${\rm exp}^{k_{\text{cov}}}$ and ${\rm num}^{k_{\text{cov}}}$ are initialized to zero, and $\mathbf{cm}^{k_{\text{cov}}}$ is initialized to a zero vector.

\subsubsection{Genetic Operator} \label{ss: genetic operators}

In \ours, GA plays a role in discovering plausible {MFs} by evolving the fuzzy sets. It executes almost the same GA procedure as in Fuzzy-UCS. However, unlike the existing fuzzy representations, our fuzzy set includes the shape and interval parameters, which have different semantics and thus different ranges of values. Accordingly, we slightly modify the mutation operator for the shape and interval parameters. 

The mutation operator is performed with a probability of $p_\text{mut}$ and mutates each parameter contained in $A_{i}^k$ by adding a random value to it (the mutation probability is applied to each fuzzy set, i.e., all the four parameters are mutated simultaneously, {referring to \cite{nakata2020learning}}. Specifically, the mutation operator is performed as:
\begin{flalign}
    \label{eq: beta mutation ab}
    \begin{cases}
     \alpha^{k}_i &\gets \alpha^{k}_i + \alpha^{k}_i\cdot \mathcal{U}{[-0.5, 0.5)},\\
     \beta^{k}_i &\gets \beta^{k}_i + \beta^{k}_i \cdot \mathcal{U}{[-0.5, 0.5)},\\
     l^{k}_i &\gets l^{k}_i + \mathcal{U}{[-m_0, m_0)},\\
     u^{k}_i &\gets u^{k}_i + \mathcal{U}{[-m_0, m_0)},
    \end{cases}
\end{flalign}
where $m_0$ defines the fluctuation range for the interval parameters. For the shape parameters, a relative mutation \cite{butz2008function} is used to allow large changes in parameter values, enhancing \ours\ to explore various {MFs}. In our case, the mutation allows the current value to be increased or decreased by up to 50\%. If, after mutation, $\alpha_{i}^k$ or $\beta_i^k$ is less than 1.0, it is reset to 1.0.

\subsubsection{Subsumption Operator}
\label{ss: subsumption operator}
To implement a subsumption operator for \ourrep, we define an \textit{is-more-general} operator which tests whether a rule ${k_\text{sub}}$ has a more general condition than a rule ${k_\text{tos}}$. This requires a comparison of the inclusion relation between two {MFs}.

In \ourrep, evaluating the exact inclusion relation is challenging since \ourrep\ represents various {MF} shapes.
Several methods have been proposed to determine the inclusion relation between two complex {MFs}. For example, in trapezoidal-shaped {MFs} \cite{shoeleh2010handle}, the inclusion relation is determined strictly using the piecewise quadrature method, which is computationally time-consuming. On the other hand, in normal distribution-shaped {MFs} \cite{tadokoro2021xcs}, inclusion relations can be approximated based on the mode similarity.
As in \cite{tadokoro2021xcs}, the is-more-general operator for \ourrep\ uses the concept of mode similarity. Additionally, it incorporates considerations based on kurtosis and lower/upper bounds to enhance the generalization bias towards crisp rules. The advantage of \ourrep\ lies in its ability to uniquely compute these properties for different shapes using the PDF of the four-parameter beta distribution.

In \ours, the is-more-general operator is based on the following AND Conditions 1 and 2. If neither $\mu_{A_i^{k_\text{sub}}}$ nor $\mu_{A_i^{k_\text{tos}}}$ are rectangular-shaped, Condition 3 must be satisfied in addition to Conditions 1 and 2.

\begin{figure*}[!t]
\centering
\subfloat[Condition 1]{\includegraphics[width=2.425in]{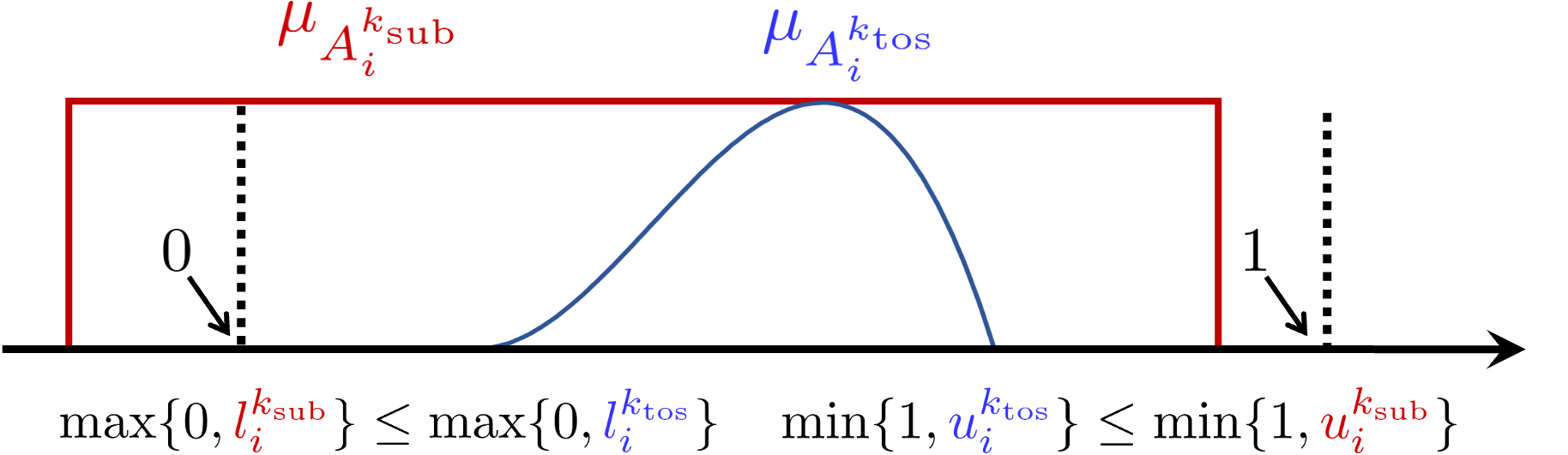}%
\label{fig: subsumption-condition1}}
\subfloat[Condition 2]{\includegraphics[width=2.425in]{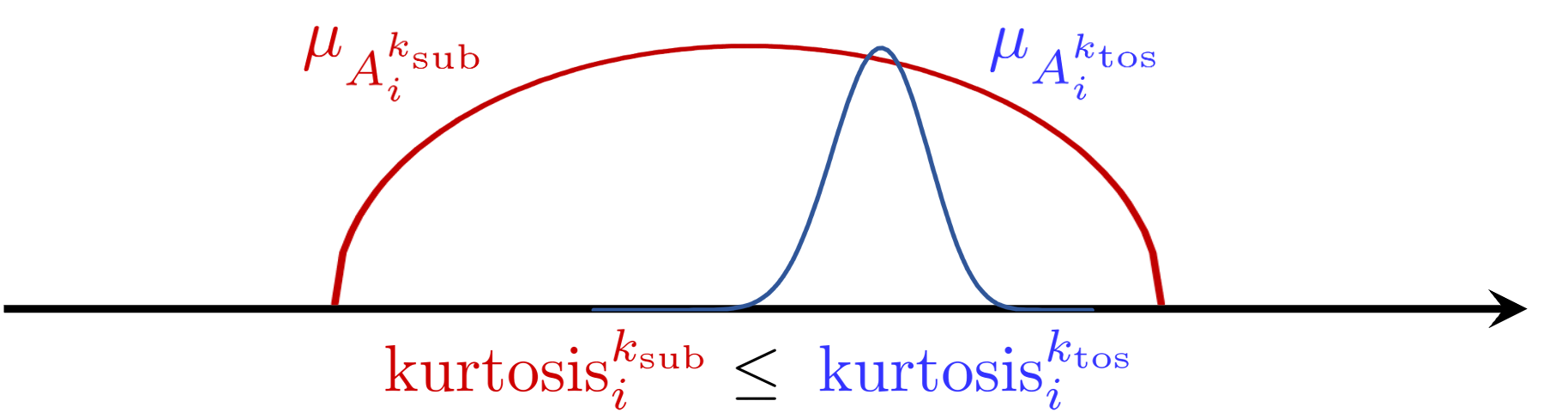}%
\label{fig: subsumption-condition2}}
\subfloat[Condition 3]{\includegraphics[width=2.425in]{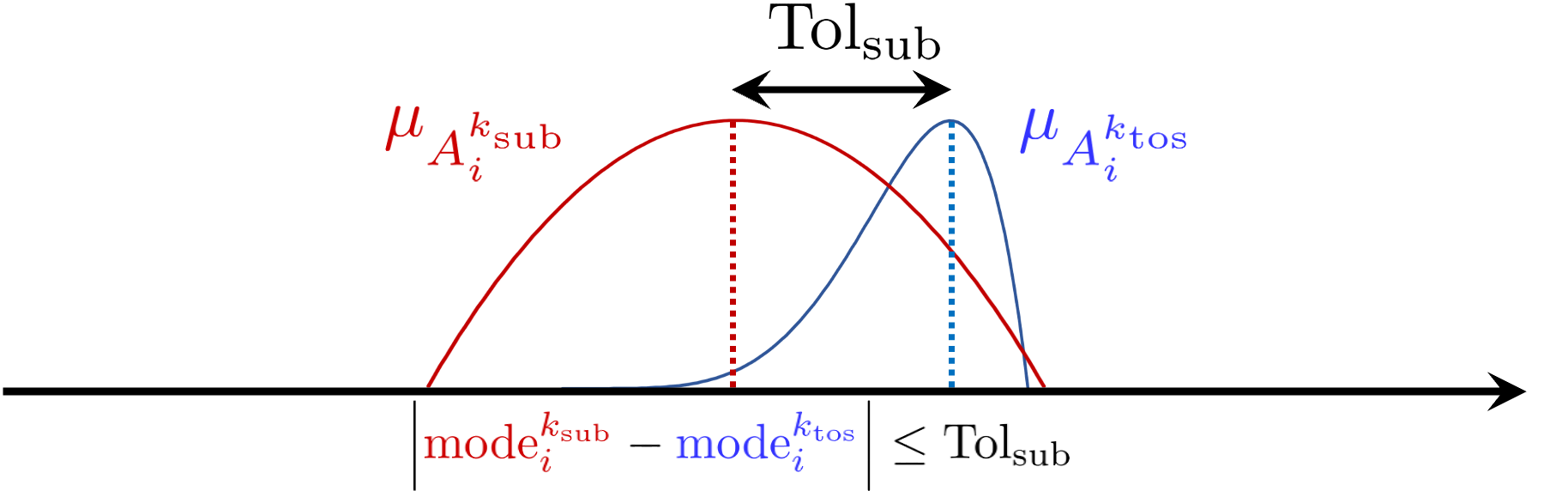}%
\label{fig: subsumption-condition3}}
\caption{
Illustration of the is-more-general operator in \ours. Rule ${k_\text{sub}}$ is deemed \textit{more general} than ${k_\text{tos}}$ if all (a) Condition 1 and (b) Condition 2 are met in every dimension $i$. If neither $\mu_{A_i^{k_\text{sub}}}$ nor $\mu_{A_i^{k_\text{tos}}}$ are rectangular-shaped, (c) Condition 3 must be satisfied in addition to (a) and (b).
}
\vspace{-2mm}
\label{fig: subsumption-condition}
\end{figure*}

\textbf{Condition 1.}
    For every dimension $i$, the criteria in \eqref{eq: subsumption-condition1} must be met. Specifically, in the domain $[0,1]$, the interval $[l_i^{k_\text{sub}},u_i^{k_\text{sub}}]$ should fully contain the interval $[l_i^{k_\text{tos}},u_i^{k_\text{tos}}]$ (see Fig. \ref{fig: subsumption-condition1}):
    \begin{flalign}
    \label{eq: subsumption-condition1}
    \begin{cases}
        \max\{0, l_i^{k_\text{sub}}\} &\leq \max\{0,l_i^{k_\text{tos}}\},\\
        \min\{1,u_i^{k_\text{tos}}\} &\leq \min\{1,u_i^{k_\text{sub}}\}.
            \end{cases}
    \end{flalign}
    
    \textbf{Condition 2.}
    For every dimension $i$, the criterion in \eqref{eq: subsumption-condition2} must be met. Specifically, the peak of ${\mu_{A_i^{k_\text{sub}}}}$ should be more rounded compared to the peak of ${\mu_{A_i^{k_\text{tos}}}}$ (see Fig. \ref{fig: subsumption-condition2}):
    \begin{equation}
    \label{eq: subsumption-condition2}
        {\rm kurtosis}_i^{k_\text{sub}} \leq \;{\rm kurtosis}_i^{k_\text{tos}}.
    \end{equation}
    
    \textbf{Condition 3.}
    For every dimension $i$, the criterion in \eqref{eq: subsumption-condition3} must be met. Specifically, the distance between the mode of ${\mu_{A_i^{k_\text{sub}}}}$ and the mode of ${\mu_{A_i^{k_\text{tos}}}}$ is within the hyperparameter ${\rm Tol}_\text{sub}$ (see Fig. \ref{fig: subsumption-condition3}):
    \begin{equation}
    \label{eq: subsumption-condition3}
        \left|{\rm mode}_i^{k_\text{sub}} - {\rm mode}_i^{k_\text{tos}}\right| \leq {\rm Tol}_\text{sub}.
    \end{equation}
{Following the same procedure as Fuzzy-UCS, when $k_\text{sub}$ is more general than $k_\text{tos}$ and $k_\text{sub}$ is sufficiently updated and accurate (i.e., ${\rm exp}^{k_\text{sub}} > \theta_\text{sub}$ and $F^{k_\text{sub}}>F_0$), $k_\text{tos}$ is removed from $[P]$ and its numerosity is added to $k_\text{sub}$.}

Fig. \ref{fig: subsumption-condition} presents a visual representation of these conditions for a specific dimension $i$.
The main goal of these conditions is to guide the system towards acquiring a more general rule set, characterized by broader intervals and smoother peaks.
More specifically, if the subspace of ${k_\text{sub}}$ includes that of ${k_\text{tos}}$ (see Condition 1, which is equivalent to the condition in the ordered-bound representation \cite{wilson2000mining}), ${k_\text{sub}}$ has a smaller kurtosis than ${k_\text{tos}}$ (see Condition 2), and their modes are similar (see Condition 3, which is equivalent to the condition in the normal distribution-shaped {MF} \cite{tadokoro2021xcs}), then ${k_\text{sub}}$ can be considered more general than ${k_\text{tos}}$. The similarity of the mode is determined by the tolerance parameter ${\rm Tol}_\text{sub}$.

\subsubsection{Crispification Operator}
\label{ss: crispification operator}

In \ours, a \textit{crispification} operator can be used if necessary. This operator further strengthens the generalization bias towards crisp-rectangular rules, aiming to simplify the rule set to the most straightforward (i.e., crisp) form. The concept of this operator is inspired by Ockham's Razor, which posits that ``Entities should not be multiplied beyond necessity'' \cite{liu2019absumption, iacca2012ockham}.

Our crispification operator is activated after the rule-parameter update in the \ours\ framework. Specifically, the proposed algorithm selects a sufficiently updated and accurate rule {$k$ (i.e., ${\rm exp}^{k} > \theta_\text{sub}$ and $F^{k}>F_0$)} and then modifies a fuzzy set representing a non-rectangular-shaped {MF} to a rectangular shape. First, \ours\ with the crispification operator builds a \textit{crispification set}, $[\mathfrak{C}]$, given by:
\begin{flalign}  \label{eq:crispification_set}
    [\mathfrak{C}] = \{k\in[C] \mid {\rm exp}^k>\theta_\text{sub} \land\; F^k>F_0\}.
\end{flalign}

Next, for each $k\in[\mathfrak{C}]$, the system randomly selects a dimension index $j$ from $\{{i\in\{1,2,...,d\}\mid \neg(\alpha_i^k = \beta_i^k=1})\}$ to decide a target fuzzy set $A_j^k$ for crispification; and then it modifies two shape parameters $\alpha_j^k, \beta_j^k$ as:
\begin{equation}
    \label{eq: crispification}
    \begin{cases}
        \alpha_j^k &\leftarrow 1,\\
        \beta_j^k &\leftarrow 1.\\
    \end{cases} 
    \end{equation}

Finally, ${\rm exp}^k$ and $\mathbf{cm}^k$ are initialized to zero and a zero vector, respectively. Thus, once a rule is modified by the crispification operator, the system temporarily removes its rule from candidates for crispification until {it becomes sufficiently updated and accurate (i.e., ${\rm exp}^k>\theta_\text{sub}$ and $F^k>F_0$) again.}

\subsubsection{Summary of \ours}

{The redesigned operators work synergistically in \ours\ to balance interpretability and adaptability. The covering operator's initialization of crisp rules provides a foundation for interpretability, while the genetic operator's relative mutation of shape parameters enables smooth adaptation to local subspace characteristics. The subsumption operator maintains consistent generalization across different rule representations, and the crispification operator further simplifies rules where possible. Together, these modifications enable \ours\ to adapt representation complexity based on local classification difficulty while maintaining a bias toward interpretable crisp rules without sacrificing accuracy.}

\section{Experiments}
\label{sec: experiment}

This section utilizes 25 datasets {from LCS literature, the UCI Machine Learning Repository \cite{dua2019uci}, and Kaggle. These datasets, listed in Table \ref{tb: dataset}, include both traditional benchmarks and modern problems.}
All the selected datasets belong to real-world domains, except for \texttt{car}, \texttt{chk}, \texttt{cmx}, \texttt{mop}, and \texttt{mux}, which are typical artificial problems used for evaluating the performance of LCSs. For the artificial problems, each dataset consists of 6,000 data points uniformly sampled at random from the input space {$[0,1]^d$ for acceptable computational time.} These datasets were selected due to their inherent challenges in data classification, such as the presence of missing values, overlapping classes, and class imbalance issues \cite{hamasaki2021minimum,shiraishi2023fuzzy}.

\subsection{Experimental Setup}

We compare the performance of 10 rule representations: four crisp rule representations using UCS, four fuzzy rule representations using Fuzzy-UCS, \ourrep\ without the crispification operator (termed ``\ourrep”) using \ours, and \ourrep\ with the crispification operator (termed ``\ourrep C”) using \ours.{\footnote{Appendix \ref{sec: sup comparison with modern machine learning methods} offers a comparison with modern machine learning methods.}} The system hyperparameters for UCS, Fuzzy-UCS, and \ours\ are set to the values established in previous works \cite{bernado2003accuracy,orriols2008fuzzy,tzima2013strength,urbanowicz2015exstracs}, as shown in Table \ref{tb: hyperparameters}. All 10 rule representations and their specific hyperparameter settings are detailed in Table \ref{tb: representation hyperparameters}. {The hyperparameter values for existing rule representations are selected based on their original papers, where these values were either used directly or determined through ablation studies to be optimal. For our proposed \ourrep C, we conduct comprehensive parameter sensitivity analysis of its key hyperparameters $\text{Tol}_\text{sub}$ and $r_0$, with results presented in Appendices \ref{sec: sup impact of the subsumption tolerance tolsub} and \ref{sec: sup impact of the covering range parameter r0}.}

Note that all of the existing rule representations listed in Table \ref{tb: representation hyperparameters}, except for \textit{Hyperrectangles}, \textit{Triangles}, and \textit{Self-Ada-RT}, were originally operated with XCS \cite{wilson1995xcs} or XCSF \cite{wilson2002classifiers}. However, to fairly compare the performance of all existing rule representations on classification tasks in this {article}, {we use UCS/Fuzzy-UCS designed to optimize crisp/fuzzy rules.}\label{others-8} Specifically, we port each representation-specific (1) matching function or matching degree calculation method, (2) covering operator, (3) mutation operator, and (4) subsumption operator to UCS and Fuzzy-UCS. For example, the performance of Fuzzy-UCS with \textit{Trapezoids} has been tested in \cite{shiraishi2023fuzzy}.

\begin{table}[!t]
\begin{center}
\caption{Properties of the 25 Datasets in the Experiment. The Columns Describe: the Identifier (ID.), the Name (Name), the Number of Instances ($\#$\textsc{Inst.}), the Total Number of Features ($\#$\textsc{Fea.}), the Number of Classes ($\#$\textsc{Cl.}), the Percentage of Missing Data Attributes ($\%$\textsc{Mis.}), and the Source (Ref.).}
\label{tb: dataset}
\normalsize
\resizebox{\linewidth}{!}{
\begin{tabular}{c l c c c c c}
\bhline{1pt}
ID. & Name  & $\#$\textsc{Inst.}  & $\#$\textsc{Fea.} &$\#$\textsc{Cl.} & $\%$\textsc{Mis.} & Ref.
\\
\bhline{1pt}
{\texttt{cae}} &{Caesarian section} & {80} & {5} & {2} & {0}&\;{\tablefootnote{\url{https://www.kaggle.com/datasets/alihasnainch/caesarian-section-dataset-for-classification} (\today)}}\\
\texttt{can} &Cancer & 569 & 30 & 2 & 0 & \;\tablefootnote{\url{https://www.kaggle.com/datasets/erdemtaha/cancer-data} (\today)} \\
\texttt{car} &$12$ {Real-valued} carry & 6000 & 12 & 2 & 0 & \cite{iqbal2013reusing}\\
\texttt{chk} &$3\times 5$ {Real-valued} checkerboard & 6000 & 3 & 2 & 0 & \cite{stone2003real} \\
\texttt{cmx} &$3\times 3$ {Real-valued} concatenated multiplexer & 6000 & 9 & 8 & 0 & \cite{butz2003analysis}\\
\texttt{col} &Column 3C weka & 310 & 6 & 3 & 0 & \cite{dua2019uci}  \\
\texttt{dbt} &Diabetes & 768 & 8 & 2 & 0 & \cite{dua2019uci} \\
\texttt{ecl} &Ecoli & 336 & 7 & 8 & 0 & \cite{dua2019uci}\\
\texttt{frt} &Fruit & 898 & 34 & 7 & 0 & \cite{koklu2021classification}\\
\texttt{gls} &Glass & 214 & 9 & 6 & 0 & \cite{dua2019uci} \\
\texttt{hcl} &Horse colic & 368 & 22 & 2 & 23.80 & \cite{dua2019uci}\\
\texttt{mam} &Mammographic masses & 961 & 5 & 2 & 3.37 & \cite{dua2019uci}\\
\texttt{mop} &$11$ {Real-valued} majority & 6000 & 11 & 2 & 0 & \cite{hamasaki2021minimum}\\
{\texttt{mul}} & {Multiple sclerosis disease} &{273}&{18}&{2}&{0.06}&
\cite{chavarria2023conversion}\\
\texttt{mux} &$20$ {Real-valued} multiplexer & 6000 & 20 & 2 & 0 & \cite{wilson1999xcsr}\\ 
{\texttt{nph}} & {National poll on healthy aging} & {714} & {14} & {2} & {0}&
\cite{malani2019npha}\\
\texttt{pdy} &Paddy leaf & 6000 & 3 & 4 & 0 & \;\tablefootnote{\url{https://www.kaggle.com/datasets/torikul140129/paddy-leaf-images-aman} (\today)} \\
\texttt{pis} &Pistachio & 2148 & 16 & 2 & 0 & \cite{singh2022classification}\\
\texttt{pmp} &Pumpkin & 2499 & 12 & 2 & 0 & \cite{koklu2021use} \\
\texttt{rsn} &Raisin & 900 & 7 & 2 & 0 & \cite{ccinar2020classification}\\
\texttt{soy} &Soybean & 683 & 35 & 19 & 9.78 & \cite{dua2019uci}  \\
\texttt{tae} &Teaching assistant evaluation & 151 & 5 & 3 & 0 & \cite{dua2019uci}\\
\texttt{wne} &Wine & 178 & 13 & 3 & 0 & \cite{dua2019uci}\\
\texttt{wpb} &Wisconsin prognostic breast cancer & 198 & 33 & 2 & 0.06 & \cite{dua2019uci} \\
\texttt{yst} &Yeast & 1484 & 8 & 10 & 0 & \cite{dua2019uci} \\
\bhline{1pt}
\end{tabular}
}
\end{center}
\vspace{-2mm}
\end{table}
The number of training iterations is set to 50 epochs, {where each epoch processes the entire training dataset once. This results in $50N_\text{tr}$ total iterations, where $N_\text{tr}$ represents the number of instances in the training dataset.} Uniform crossover and tournament selection{\footnote{Tournament selection is implemented by first randomly selecting $\tau\cdot|[C]|$ rules from $[C]$, and then choosing the rule with the highest fitness for UCS or the highest product of membership degree and fitness for Fuzzy-UCS and \ours\ from this tournament pool as the parent.}} are employed. We conduct 30 independent experiments, each with a unique random seed. Each attribute value in the data is normalized to the range $[0,1]$.{\footnote{
Each input attribute is normalized separately using min-max values from training data only. These values normalize both training and test data to [0,1], with test values clipped to this range.}} The datasets are divided using shuffle-split cross-validation, allocating 90\% of the instances for training and the remaining 10\% for testing, as in \cite{preen2021autoencoding}. The primary performance metric used to evaluate the representations is the average test classification accuracy. Additionally, we evaluate the average training classification accuracy, the population size, and the computational time. All experiments are conducted on our implementation written in Julia \cite{bezanson2017julia}\label{others-5}, using an Intel\textregistered\ Core\texttrademark\ i7-9700 CPU with 3.00 GHz and 16 GB RAM. {Julia is an efficient programming language suitable for LCS implementations due to its efficient parallel computing capabilities.} 

For the analysis of statistical significance on all metrics except for computational time, we begin by applying the Friedman test to the results of 30 runs across all 10 representations. If the Friedman test indicates a {significant} probability small enough, we then proceed with the Holm test, using the Wilcoxon signed-rank test as a post-hoc method. A difference is considered significant if the probability is less than 0.05 in both the Friedman and Holm post-hoc tests.
\subsection{Results}

\subsubsection{Test Classification Accuracy}
\label{sss: exp test accuracy}
\begin{table}[!t]
\begin{center}
\caption{Properties of the System Hyperparameter Values.}
\label{tb: hyperparameters}
\normalsize
\resizebox{\linewidth}{!}{
\begin{tabular}{l c |  c c}
\bhline{1pt}
\multicolumn{1}{l}{\multirow{2}{*}{Description}}  & \multicolumn{1}{c|}{\multirow{2}{*}{Hyperparameter}}  & \multicolumn{1}{l}{\multirow{2}{*}{UCS}} & Fuzzy-UCS\\
&&&\ours\\
\bhline{1pt}
Maximum population size & $N$ & 2000 & 2000\\
Accuracy threshold & ${\rm acc}_0$ & 0.99 & N/A\\
Fitness threshold & $F_0$ & N/A & 0.99\\
Learning rate & $\beta$ & 0.2 & N/A\\
Fitness exponent & $\nu$ & 1 & 1\\
Crossover probability & $\chi$ & 0.8 & 0.8\\
Mutation probability & $p_\text{mut}$ & 0.04 & 0.04\\
Fraction of mean fitness for rule deletion & $\delta$ & 0.01 & 0.01\\
Time threshold for GA application in $[C]$ & $\theta_\text{GA}$ & 50 & 50\\
Experience threshold for rule deletion & $\theta_\text{del}$ & 50 & 50\\
Experience threshold for subsumption & $\theta_\text{sub}$ & 50 & 50\\
Experience threshold for class inference & $\theta_\text{exp}$ & N/A & 10\\
Tournament size ratio & $\tau$ & 0.4 & 0.4\\
Probability of \textit{Don't Care} in covering & $P_\#$ & 0.33 & 0.33\\
Whether correct set subsumption is performed & $\mathit{doCSSubsumption}$ & $\mathit{yes}$ & $\mathit{yes}$\\
Whether GA subsumption is performed & $\mathit{doGASubsumption}$ & $\mathit{yes}$ & $\mathit{yes}$\\

\bhline{1pt}
\end{tabular}
}
\end{center}
\vspace{-2mm}
\end{table}

\begin{table*}[!t]
\begin{center}
\caption{Properties of the Representation-Specific Subspace Shape and Fuzziness (Type) and Hyperparameter Values.}
\vspace{-2mm}
\label{tb: representation hyperparameters}
\normalsize
\scalebox{0.725}{
\begin{tabular}{r l l | l }
\bhline{1pt}
Representation / Membership functions  & Abbreviation & Type & Hyperparameters\\
\bhline{1pt}
Unordered-bound representation \cite{stone2003real} & \textit{Hyperrectangles} & Fig. \ref{fig: rec} & $r_0=1.0$, $m_0=0.1$.\\ 
Kernel-based representation \cite{butz2008function} & \textit{Hyperellipsoids} & Fig. \ref{fig: elp} & $r_0=1.0$, $\theta_m=0.7$.\\
Bivariate beta distribution-based representation \cite{shiraishi2022beta} & \textit{CurvedPolytopes} & Fig. \ref{fig: eclipselike} & $m_0=5.0$, $\theta_m=0.7$, $s_0=5.0$.\\
Self-adaptive bivariate beta distribution-based representation \cite{shiraishi2022can} \hspace{0.65mm} & \textit{Self-Ada-RP} &  Figs. \ref{fig: rec}, \ref{fig: eclipselike} & $m_0=5.0$, $\theta_m=0.7$, $s_0=5.0$, $m_\zeta=0.5$, $\theta_\text{and}=0.5$.\\
Non-grid-oriented triangular-shaped membership functions \cite{orriols2011fuzzy} & \textit{Triangles} & Fig. \ref{fig: tri} & $m_0=0.1$.\\
Trapezoidal-shaped membership functions \cite{shoeleh2011towards} & \textit{Trapezoids} & Fig. \ref{fig: trp} & $r_0=1.0$, $m_0=0.1$, $\theta_\text{overlap}=1.0$.\\
Normal distribution-based membership functions \cite{tadokoro2021xcs} & \textit{SymmetricBells} & Fig. \ref{fig: fec} & $m_0=0.1$, $\sigma_0=0.25$, ${\rm Tol}_\text{sub}=0.01$.\\
Rectangular- and triangular-shaped membership functions \cite{shiraishi2023fuzzy} & \textit{Self-Ada-RT} & Figs. \ref{fig: tri}, \ref{fig: fre} & $r_0=1.0$, $m_0=0.1$, $\theta_\text{overlap}=1.0$.\\
Four-parameter beta distribution-based representation (Ours) & \ourrep (C)& Fig. \ref{fig:tbr_mf} & $r_0=1.0$, $m_0=0.1$, ${\rm Tol}_\text{sub}=0.01$.\\

\bhline{1pt}
\end{tabular}
}
\end{center}
\vspace{-2mm}
\end{table*}

\begin{table*}[!t]
\begin{center}

\caption{Summary of Results, Displaying Average {(a) Test Classification Accuracy, (b) Training Classification Accuracy, and (c) Population Size, }Across 30 Runs. Green-Shaded Values Denote the Best Values, While Peach-Shaded Values Indicate the Worst Values. ``Rank'' and ``Position'' Denote Each System’s Overall Average Rank Obtained by Using the Friedman Test and Its Position in the Final Ranking, Respectively. Statistical Results of the Wilcoxon Signed-Rank Test Are Summarized With Symbols Where {``$+$'' and ``$-$'' Indicate Significantly Better and Worse Performance Than FBRC, Respectively, While ``$\sim$'' Indicates No Significant Difference From FBRC. The ``$p$-Value'' Row Shows the Raw $p$-Values From These Comparisons, While the ``$p_\text{Holm}$-Value'' Row Shows the Adjusted $p$-Values Using Holm's Correction Method to Control for Multiple Comparisons.}}
\label{tb: result}
\vspace{-3mm}
\normalsize
\mbox{\small (a) Test Classification Accuracy}\\
\vspace{2mm}
\scalebox{0.725}
{
\begin{tabular}{c|cccc|cccc|cc}
\bhline{1pt}
&\multicolumn{4}{c|}{UCS} & \multicolumn{4}{c|}{Fuzzy-UCS} & \multicolumn{2}{c}{\ours}
\\
\cline{2-11}
ID.  & \textit{Hyperrectangles} & \textit{Hyperellipsoids} & \textit{CurvedPolytopes} & \textit{Self-Ada-RP}  & \textit{Triangles}&\textit{Trapezoids}&\textit{SymmetricBells}&\textit{Self-Ada-RT}& \ourrep&\ourrep C
\\
\bhline{1pt}
{\texttt{cae}} & \cellcolor{g}65.83 $\sim$ & 62.92 $\sim$ & 63.75 $\sim$ & 62.50 $\sim$ & \cellcolor{g}65.83 $\sim$ & 61.67 $\sim$ & \cellcolor{p}61.25 $\sim$ & 64.58 $\sim$ & 62.50 $\sim$ & 62.08 \\
\texttt{can} & 91.75 $-$ & 88.89 $-$ & \cellcolor{p}65.15 $-$ & 81.40 $-$ & 89.24 $-$ & 93.92 $-$ & \cellcolor{g}95.67 $\sim$ & 93.39 $-$ & 94.33 $-$ & 95.32 \\
\texttt{car} & 92.28 $-$ & 85.57 $-$ & 87.72 $-$ & 93.37 $-$ & 89.88 $-$ & 89.46 $-$ & \cellcolor{p}83.92 $-$ & 91.36 $-$ & 94.28 $\sim$ & \cellcolor{g}94.58 \\
\texttt{chk} & 54.76 $-$ & 53.44 $-$ & 53.15 $-$ & 54.44 $-$ & 58.56 $-$ & \cellcolor{p}50.89 $-$ & 54.02 $-$ & 80.63 $\sim$ & \cellcolor{g}81.61 $\sim$ & 81.32 \\
\texttt{cmx} & 71.84 $-$ & 61.01 $-$ & 51.23 $-$ & 80.80 $-$ & \cellcolor{p}7.494 $-$ & 88.01 $-$ & 12.61 $-$ & 87.91 $-$ & 92.32 $\sim$ & \cellcolor{g}92.86 \\
\texttt{col} & \cellcolor{p}70.00 $-$ & 74.84 $\sim$ & 78.17 $\sim$ & 78.28 $\sim$ & \cellcolor{g}78.82 $\sim$ & 71.29 $-$ & 73.76 $-$ & 78.49 $\sim$ & 78.17 $\sim$ & 77.63 \\
\texttt{dbt} & 72.90 $-$ & 75.54 $\sim$ & 75.41 $\sim$ & \cellcolor{g}76.02 $\sim$ & 74.11 $\sim$ & 74.46 $\sim$ & \cellcolor{p}71.13 $-$ & 75.24 $\sim$ & 74.63 $\sim$ & 74.89 \\
\texttt{ecl} & \cellcolor{p}76.18 $-$ & \cellcolor{g}86.27 $\sim$ & 82.25 $-$ & 83.82 $\sim$ & 85.20 $\sim$ & 80.98 $-$ & 85.98 $\sim$ & 84.90 $\sim$ & 83.82 $\sim$ & 85.39 \\
\texttt{frt} & 83.41 $-$ & 83.26 $-$ & \cellcolor{p}29.04 $-$ & 67.48 $-$ & 76.70 $-$ & \cellcolor{g}86.19 $\sim$ & 82.81 $-$ & 82.85 $-$ & 85.78 $\sim$ & 85.19 \\
\texttt{gls} & \cellcolor{p}53.33 $-$ & 66.21 $\sim$ & 62.12 $-$ & 63.94 $\sim$ & 64.24 $\sim$ & 63.03 $-$ & \cellcolor{g}69.24 $\sim$ & 67.42 $\sim$ & 66.52 $\sim$ & 67.73 \\
\texttt{hcl} & 63.69 $-$ & \cellcolor{p}37.66 $-$ & 63.78 $-$ & 64.32 $-$ & 68.29 $\sim$ & \cellcolor{g}72.16 $\sim$ & 68.11 $\sim$ & 68.56 $\sim$ & 71.80 $\sim$ & 70.90 \\
\texttt{mam} & 80.72 $\sim$ & 80.34 $\sim$ & 80.72 $\sim$ & 80.82 $\sim$ & 81.10 $\sim$ & 80.03 $-$ & \cellcolor{p}78.90 $-$ & 81.13 $\sim$ & 81.24 $\sim$ & \cellcolor{g}81.41 \\
\texttt{mop} & 87.23 $-$ & 85.21 $-$ & 84.91 $-$ & \cellcolor{g}92.32 $+$ & 87.88 $-$ & \cellcolor{p}83.17 $-$ & 83.76 $-$ & 87.31 $-$ & 91.52 $\sim$ & 91.71 \\
{\texttt{mul}} & 76.19 $-$ & \cellcolor{p}62.50 $-$ & 83.21 $\sim$ & \cellcolor{g}88.10 $\sim$ & 67.38 $-$ & 80.95 $-$ & 67.26 $-$ & 74.29 $-$ & 85.12 $\sim$ & 86.55 \\
\texttt{mux} & 55.36 $-$ & \cellcolor{p}49.31 $-$ & 57.73 $-$ & 66.40 $-$ & 49.34 $-$ & 63.92 $-$ & 57.69 $-$ & 87.08 $-$ & 94.66 $-$ & \cellcolor{g}96.44 \\
{\texttt{nph}} & \cellcolor{g}57.08 $+$ & 54.44 $\sim$ & 56.34 $\sim$ & 56.76 $\sim$ & 54.26 $\sim$ & 54.95 $\sim$ & 55.00 $\sim$ & \cellcolor{p}53.47 $\sim$ & 54.81 $\sim$ & 53.98 \\
\texttt{pdy} & 88.24 $-$ & 85.74 $-$ & 87.64 $-$ & 87.54 $-$ & 87.29 $-$ & \cellcolor{p}78.27 $-$ & 85.49 $-$ & 88.72 $\sim$ & 88.95 $\sim$ & \cellcolor{g}89.02 \\
\texttt{pis} & 86.12 $\sim$ & 85.97 $\sim$ & \cellcolor{p}85.50 $-$ & 85.55 $-$ & 86.42 $\sim$ & 85.60 $-$ & 86.16 $\sim$ & 85.97 $-$ & 86.57 $\sim$ & \cellcolor{g}86.60 \\
\texttt{pmp} & 86.01 $-$ & 86.83 $-$ & 86.80 $\sim$ & 86.43 $\sim$ & 86.84 $\sim$ & \cellcolor{p}85.89 $-$ & 86.43 $-$ & 86.52 $-$ & 86.71 $\sim$ & \cellcolor{g}87.19 \\
\texttt{rsn} & 84.63 $\sim$ & \cellcolor{g}86.74 $+$ & 85.67 $\sim$ & 85.93 $\sim$ & 85.67 $\sim$ & \cellcolor{p}84.19 $\sim$ & 85.48 $\sim$ & 85.81 $\sim$ & 85.52 $\sim$ & 85.52 \\
\texttt{soy} & 67.00 $-$ & \cellcolor{p}46.96 $-$ & 58.02 $-$ & 60.34 $-$ & 61.69 $-$ & 76.43 $+$ & \cellcolor{g}79.57 $+$ & 60.63 $-$ & 68.65 $\sim$ & 69.52 \\
\texttt{tae} & 49.17 $\sim$ & 54.38 $\sim$ & 55.62 $\sim$ & 52.71 $\sim$ & 52.08 $\sim$ & \cellcolor{p}48.33 $\sim$ & \cellcolor{g}56.25 $+$ & 54.79 $\sim$ & 50.42 $\sim$ & 51.04 \\
\texttt{wne} & 88.52 $\sim$ & 94.63 $\sim$ & \cellcolor{p}68.89 $-$ & 85.37 $-$ & \cellcolor{g}94.81 $\sim$ & 93.15 $\sim$ & \cellcolor{g}94.81 $+$ & 92.78 $\sim$ & 92.78 $\sim$ & 92.04 \\
\texttt{wpb} & 70.00 $\sim$ & 75.50 $\sim$ & 76.00 $\sim$ & 75.00 $\sim$ & \cellcolor{p}61.67 $-$ & \cellcolor{g}77.00 $\sim$ & 66.00 $-$ & 72.67 $\sim$ & 73.00 $\sim$ & 73.83 \\
\texttt{yst} & 56.89 $-$ & 59.19 $\sim$ & 58.21 $\sim$ & 58.48 $-$ & \cellcolor{g}59.80 $\sim$ & \cellcolor{p}50.11 $-$ & 57.99 $-$ & 58.14 $\sim$ & 59.08 $\sim$ & 59.71 \\
\bhline{1pt}
Rank & \cellcolor{p}\textit{6.72} & \textit{6.00} & \textit{6.44} & \textit{5.24} & \textit{5.40} & \textit{6.68} & \textit{6.36} & \textit{4.82} & \textit{3.96} & \cellcolor{g}\textit{3.38} \\
Position & \textit{10} & \textit{6} & \textit{8} & \textit{4} & \textit{5} & \textit{9} & \textit{7} & \textit{3} & \textit{2} & \textit{1} \\
$+/-/\sim$ & 1/17/7 & 1/12/12 & 0/14/11 & 1/12/12 & 0/11/14 & 1/15/9 & 3/14/8 & 0/10/15 & 0/2/23 & - \\
\bhline{1pt}
$p$-value & 7.50E-05 & 0.0105 & 0.00203 & 0.0147 & 0.00278 & 0.00226 & 0.0115 & 0.00964 & 0.0738 & - \\
$p_\text{Holm}$-value & 0.000675 & 0.0482 & 0.0162 & 0.0482 & 0.0167 & 0.0162 & 0.0482 & 0.0482 & 0.0738 & - \\

\bhline{1pt}
\end{tabular}}

\vspace{3mm}

\mbox{\small (b) Training Classification Accuracy}\\
\vspace{2mm}
\scalebox{0.725}
{
\begin{tabular}{c|cccc|cccc|cc}
\bhline{1pt}
&\multicolumn{4}{c|}{UCS} & \multicolumn{4}{c|}{Fuzzy-UCS} & \multicolumn{2}{c}{\ours}
\\
\cline{2-11}
ID.  & \textit{Hyperrectangles} & \textit{Hyperellipsoids} & \textit{CurvedPolytopes} & \textit{Self-Ada-RP}  & \textit{Triangles}&\textit{Trapezoids}&\textit{SymmetricBells}&\textit{Self-Ada-RT}& \ourrep&\ourrep C
\\
\bhline{1pt}
Rank & \textit{6.56} & \textit{4.76} & \textit{6.08} & \textit{4.68} & \textit{6.72} & \cellcolor{p}\textit{8.64} & \textit{6.28} & \textit{4.68} & \textit{3.64} & \cellcolor{g}\textit{2.96} \\
Position & \textit{8} & \textit{5} & \textit{6} & \textit{3} & \textit{9} & \textit{10} & \textit{7} & \textit{4} & \textit{2} & \textit{1} \\
$+/-/\sim$ & 1/20/4 & 6/13/6 & 4/16/5 & 4/13/8 & 3/20/2 & 1/23/1 & 6/17/2 & 2/15/8 & 0/5/20 & - \\
\bhline{1pt}
$p$-value & 4.54E-05 & 0.0626 & 0.0115 & 0.0342 & 3.81E-05 & 1.01E-05 & 0.00558 & 0.00278 & 0.0451 & - \\
$p_\text{Holm}$-value & 0.000318 & 0.103 & 0.0458 & 0.103 & 0.000305 & 9.07E-05 & 0.0279 & 0.0167 & 0.103 & - \\

\bhline{1pt}
\end{tabular}}

\vspace{3mm}

\mbox{\small (c) Population Size}\\
\vspace{2mm}
\scalebox{0.725}
{
\begin{tabular}{c|cccc|cccc|cc}
\bhline{1pt}
&\multicolumn{4}{c|}{UCS} & \multicolumn{4}{c|}{Fuzzy-UCS} & \multicolumn{2}{c}{\ours}
\\
\cline{2-11}
ID.  & \textit{Hyperrectangles} & \textit{Hyperellipsoids} & \textit{CurvedPolytopes} & \textit{Self-Ada-RP}  & \textit{Triangles}&\textit{Trapezoids}&\textit{SymmetricBells}&\textit{Self-Ada-RT}& \ourrep&\ourrep C
\\
\bhline{1pt}
Rank & \textit{3.88} & \textit{7.40} & \textit{5.32} & \textit{5.08} & \textit{7.32} & \textit{6.48} & \textit{3.80} & \cellcolor{p}\textit{9.20} & \textit{3.64} & \cellcolor{g}\textit{2.88} \\
Position & \textit{4} & \textit{9} & \textit{6} & \textit{5} & \textit{8} & \textit{7} & \textit{3} & \textit{10} & \textit{2} & \textit{1} \\
$+/-/\sim$ & 6/15/4 & 3/21/1 & 5/19/1 & 5/19/1 & 0/24/1 & 4/19/2 & 13/12/0 & 0/25/0 & 3/10/12 & - \\
\bhline{1pt}
$p$-value & 0.00613 & 8.17E-06 & 0.000715 & 0.000912 & 5.96E-08 & 0.00737 & 0.653 & 5.96E-08 & 0.00631 & - \\
$p_\text{Holm}$-value & 0.0245 & 5.72E-05 & 0.00429 & 0.00456 & 5.36E-07 & 0.0245 & 0.653 & 5.36E-07 & 0.0245 & - \\

\bhline{1pt}
\end{tabular}}

\end{center}
\end{table*}

Table \ref{tb: result}-a presents each representation's average test classification accuracy. Table \ref{tb: result pairwise} shows the approximate $p$-values for the pairwise comparison according to the Wilcoxon signed-rank test. {For example, in Table \ref{tb: result pairwise}, ``$\sim$({9}/{10}/{6})'' at the intersection of \textit{Hyperellipsoids} (row) and \textit{Hyperrectangles} (column) means: \textit{Hyperellipsoids} performed significantly better on {9} datasets, worse on {10}, and similar on {6}. The symbol ``$\sim$'' indicates no overall statistical difference between the representations.}

Table \ref{tb: result}-a reveals that in \ourrep\ and \ourrep C, 1 and {7} out of 25 cells are green-shaded, respectively, denoting the best values. Notably, no cells in \ourrep\ or \ourrep C are peach-shaded, which denote the worst values. \ourrep\ and \ourrep C record the second and first average ranks among all 10 representations, respectively. The Wilcoxon signed-rank test results, as summarized in Table \ref{tb: result pairwise}, show that \ourrep C records significantly better test classification accuracy than all other representations{, except for \ourrep.}\label{r1-1-1} Meanwhile, \ourrep\ records significantly higher test classification accuracy than all other representations, except for \ourrep C. Table \ref{tb: result pairwise} reveals that the test classification accuracy of \ourrep C is either {similar} or superior to the other representations across all 225 experimental cases, with the exception of {7} cases (i.e., the number of “$-$” symbols indicating that \ourrep C significantly underperformed). Similarly, the test classification accuracy of \ourrep\ is either {similar} or superior to the other representations except for \ourrep C across all 200 experimental cases, with the exception of {12} cases. These results underscore the effectiveness of \ours, its adaptive and flexible {MF}, in addressing the complex classification tasks frequently encountered in real-world scenarios. Additionally, the introduction of the crispification operator can enhance the performance of \ours\ without significant drawbacks.\label{r1-1-2}

Moreover, from Table \ref{tb: result}-a, both \textit{Self-Ada-RP} and \textit{Self-Ada-RT}, which dynamically self-adapt two types of rule representations, {achieve the fourth and third average ranks following FBRC and FBR, respectively.} This indicates that when one rule representation is inappropriate, the system can maintain its classification performance by relying on the other representation. This trend is consistent with the observations in their original works \cite{shiraishi2022can,shiraishi2023fuzzy}. Conversely, except for \textit{Self-Ada-RP} and \textit{Self-Ada-RT}, all existing representations {achieve lower average ranks than these adaptive rule representations.} 
{These results highlight how the choice of a single rule representation significantly impacts LCS performance, as discussed in Section \ref{sec: introduction}. The superior performance of adaptive approaches like \textit{Self-Ada-RP}, \textit{Self-Ada-RT}, and especially FBR(C) emphasizes the critical importance of flexible rule representation that can adapt to different problem characteristics.}

{Note that \textit{Hyperrectangles} on \texttt{mux} improves from 55.36\% to 97.05\% accuracy as training epochs increase from 50 to 300, while FBRC maintains consistently higher performance (96.44\% to 97.55\%) across both training durations. Details are provided in Appendix \ref{sec: sup extended analysis of synthetic datasets}.}

{Additionally, FBR(C) maintains its effectiveness across various evaluation settings, as detailed in the Appendices: macro F1 scores (Appendix \ref{sec: sup experimental results on macro f1 score}), different split ratios (Appendix \ref{sec: sup experimental results on 8:2 split ratio}), 10-fold cross-validation (Appendix \ref{sec: sup experimental results with 10-fold cross validation}), stratified validation (Appendix \ref{sec: sup experimental results with stratified shuffle-split cross validation}), and extended learning (Appendix \ref{sec: sup experimental results with extended learning}).}

\begin{table*}[!t]
\begin{center}
\caption{Pairwise Comparison of Average Test Classification Accuracy Over All 25 Datasets by Means of the Wilcoxon Signed-Rank Test. The Values Are the Approximated $p$-Values. The Symbols “$+$”, “$-$”, and “$\sim$” Indicate That the Representation in the Row Is Significantly Better, Worse, and {Similar} Compared to the Representation in the Column, Respectively.
}
\label{tb: result pairwise}
\normalsize
\resizebox{\textwidth}{!}
{\newlength{\onewidth}
\settowidth{\onewidth}{1}
\begin{tabular}{c|cccc|cccc|cc}
\bhline{1pt}
&\multicolumn{4}{c|}{UCS} & \multicolumn{4}{c|}{Fuzzy-UCS} & \multicolumn{2}{c}{\ours}
\\
\cline{2-11}
$+/-/\sim$& \textit{Hyperrectangles} & \textit{Hyperellipsoids} & \textit{CurvedPolytopes} & \textit{Self-Ada-RP}  & \textit{Triangles}&\textit{Trapezoids}&\textit{SymmetricBells}&\textit{Self-Ada-RT}& \ourrep&\ourrep C
\\
     \bhline{1pt}
\textit{Hyperrectangles} & - & 0.4742 & 0.8115 & 0.1267 & 0.8774 & 0.2099 & 0.7915 & 0.0115 & 0.0001 & 0.0001 \\
\textit{Hyperellipsoids} & $\sim$ (\hspace{\onewidth}9/10/\hspace{\onewidth}6) & - & 0.8949 & 0.2635 & 0.3666 & 0.8740 & 0.9158 & 0.0147 & 0.0160 & 0.0105 \\
\textit{CurvedPolytopes} & $\sim$ (\hspace{\onewidth}8/\hspace{\onewidth}8/\hspace{\onewidth}9) & $\sim$ (\hspace{\onewidth}5/\hspace{\onewidth}6/14) & - & 0.0016 & 0.2521 & 0.8325 & 0.8740 & 0.0074 & 0.0028 & 0.0020 \\
\textit{Self-Ada-RP} & $\sim$ (10/\hspace{\onewidth}4/11) & $\sim$ (\hspace{\onewidth}8/\hspace{\onewidth}5/12) & $+$ (\hspace{\onewidth}8/\hspace{\onewidth}0/17) & - & 1.0000 & 0.4908 & 0.5646 & 0.0757 & 0.0425 & 0.0147 \\
\textit{Triangles} & $\sim$ (\hspace{\onewidth}7/\hspace{\onewidth}9/\hspace{\onewidth}9) & $\sim$ (\hspace{\onewidth}6/\hspace{\onewidth}5/14) & $\sim$ (\hspace{\onewidth}9/\hspace{\onewidth}4/12) & $\sim$ (\hspace{\onewidth}6/\hspace{\onewidth}8/11) & - & 1.0000 & 0.6635 & 0.0588 & 0.0074 & 0.0028 \\
\textit{Trapezoids} & $\sim$ (11/\hspace{\onewidth}5/\hspace{\onewidth}9) & $\sim$ (\hspace{\onewidth}8/\hspace{\onewidth}7/10) & $\sim$ (\hspace{\onewidth}8/\hspace{\onewidth}6/11) & $\sim$ (\hspace{\onewidth}7/12/\hspace{\onewidth}6) & $\sim$ (\hspace{\onewidth}8/\hspace{\onewidth}8/\hspace{\onewidth}9) & - & 0.9158 & 0.0851 & 0.0013 & 0.0023 \\
\textit{SymmetricBells} & $\sim$ (\hspace{\onewidth}8/\hspace{\onewidth}7/10) & $\sim$ (\hspace{\onewidth}4/\hspace{\onewidth}6/15) & $\sim$ (\hspace{\onewidth}8/\hspace{\onewidth}9/\hspace{\onewidth}8) & $\sim$ (\hspace{\onewidth}6/10/\hspace{\onewidth}9) & $\sim$ (\hspace{\onewidth}7/10/\hspace{\onewidth}8) & $\sim$ (\hspace{\onewidth}8/\hspace{\onewidth}9/\hspace{\onewidth}8) & - & 0.0342 & 0.0173 & 0.0115 \\
\textit{Self-Ada-RT} & $+$ (11/\hspace{\onewidth}2/12) & $+$ (11/\hspace{\onewidth}2/12) & $+$ (12/\hspace{\onewidth}3/10) & $\sim$ (\hspace{\onewidth}7/\hspace{\onewidth}5/13) & $\sim$ (\hspace{\onewidth}9/\hspace{\onewidth}1/15) & $\sim$ (11/\hspace{\onewidth}4/10) & $+$ (11/\hspace{\onewidth}2/12) & - & 0.0291 & 0.0096 \\
\ourrep & $+$ (15/\hspace{\onewidth}1/\hspace{\onewidth}9) & $+$ (11/\hspace{\onewidth}3/11) & $+$ (13/\hspace{\onewidth}1/11) & $+$ (11/\hspace{\onewidth}1/13) & $+$ (12/\hspace{\onewidth}0/13) & $+$ (13/\hspace{\onewidth}2/10) & $+$ (13/\hspace{\onewidth}3/\hspace{\onewidth}9) & $+$ (\hspace{\onewidth}7/\hspace{\onewidth}1/17) & - & 0.0738 \\
\ourrep C  & $+$ (17/\hspace{\onewidth}1/\hspace{\onewidth}7) & $+$ (12/\hspace{\onewidth}1/12) & $+$ (14/\hspace{\onewidth}0/11) & $+$ (12/\hspace{\onewidth}1/12) & $+$ (11/\hspace{\onewidth}0/14) & $+$ (15/\hspace{\onewidth}1/\hspace{\onewidth}9) & $+$ (14/\hspace{\onewidth}3/\hspace{\onewidth}8) & $+$ (10/\hspace{\onewidth}0/15) & $\sim$ (\hspace{\onewidth}2/\hspace{\onewidth}0/23) & - \\
\bhline{1pt}
\end{tabular}}
\end{center}
\vspace{-2mm}
\end{table*}

\subsubsection{Training Classification Accuracy}
Table \ref{tb: result}-b presents each representation's average rank in terms of training classification accuracy across all datasets. Furthermore, the complete summary of the training classification accuracy is presented in Appendix \ref{sec: training accuracy}.
As shown in Table \ref{tb: result}-b, \ourrep\ and \ourrep C achieve the second and first average ranks, respectively, among all 10 representations. Moreover, as mentioned in Section \ref{sss: exp test accuracy}, \ourrep\ and \ourrep C also rank second and first, respectively, on average in test classification accuracy. These results confirm the effectiveness of \ourrep\ and \ourrep C, as no overfitting of rules, which could be a concern with improved training classification accuracy, is observed.
It is important to note that the average rank in test classification accuracy for all four crisp rule representations is lower than the average rank in training classification accuracy. As observed in \cite{shiraishi2023fuzzy}, the classification that relies solely on crisp rule representations is {prone to overfitting the training data.}

\subsubsection{Population Size}
Table \ref{tb: result}-c presents each representation's average rank in terms of population size across all datasets. Furthermore, the complete summary of the population size is presented in Appendix \ref{sec: population size}.
As shown in Table \ref{tb: result}-c, \ourrep\ and \ourrep C record the second and first average ranks, respectively, among all 10 representations. \ourrep C recorded a significantly smaller population size than \ourrep. This is because the crispification operator increased the number of crisp rules and eased the difficulty of satisfying subsumption conditions, thereby promoting population generalization.

\subsubsection{Computational Time}

\begin{figure}[!t]
\centering
\vspace{-5mm}
\includegraphics[width=\linewidth]{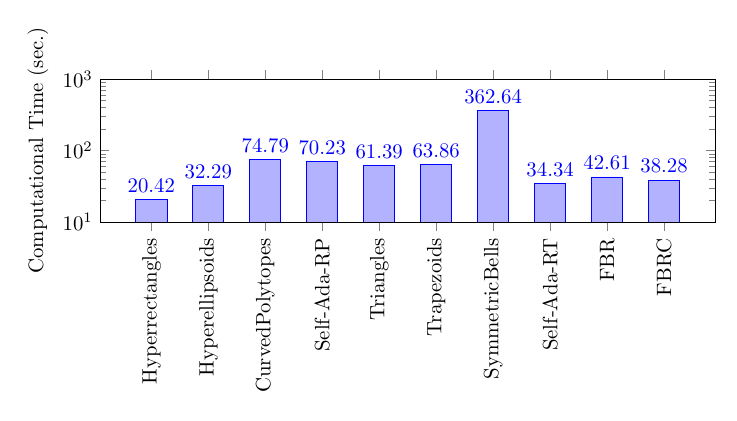}
\vspace{-8mm}
\caption{The average computational time for all 10 rule representations per run (i.e., 50 epochs) in 25 classification problems.}
\label{fig: time}
\vspace{-2mm}
\end{figure}

Fig. \ref{fig: time} presents the average computational time needed to complete one run.

An inspection of Fig. \ref{fig: time} reveals that the choice of rule representation plays a significant role in influencing the computational time of the system. It is well known that $[M]$ formation typically accounts for the majority of the computational effort in LCS execution \cite{rosenbauer2020generic}. Therefore, the complexity associated with computing the matching degree for each rule on the input, essentially the complexity of the rule representation, can cause delays in the execution speed of the LCS.

From Fig. \ref{fig: time}, we can see
that \textit{Hyperrectangles} is significantly faster, mainly because they use computationally efficient interval-based representation. In contrast, \textit{CurvedPolytopes} and \textit{Self-Ada-RP}, which frequently call expensive beta functions during $[M]$ formation, are the slowest among the crisp rule representations. \textit{SymmetricBells} is the slowest of all representations, attributed to frequent calls to the computationally intensive PDF of the multivariate normal distribution during $[M]$ formation. In general, fuzzy rule representations require more computation time compared to crisp representations. \ourrep\ and \ourrep C rank fifth and fourth in speed among all 10 representations, and they are the third and second fastest among the fuzzy representations, respectively. This efficiency is because within \ours, when $\alpha_{i}^k=\beta_{i}^k=1$, the rule $k$ behaves similarly to a hyperrectangular representation (with a rectangular {MF}). Consequently, the membership degree can be determined based on whether the input is within the interval $[l_{i}^k, u_{i}^k]$ without having to call the time-consuming beta function $B(\alpha_i^k, \beta_i^k)$ since $B(1,1)=1$, making the computation highly efficient. Furthermore, the use of the crispification operator increases the number of crisp rules, so even considering the execution time of that operator, the computational time is slightly lower than without the crispification operator. It should be noted, however, that \ours\ still takes almost twice as long to compute as the widely used hyperrectangular representation. Nonetheless, given the rapid advances in computing power, this increased time may not be a significant challenge.

\section{Analysis}
\label{sec: analysis}
This section provides further analytical results to investigate the performance of \ours. Unless otherwise noted, the setup of hyperparameters and training iterations is consistent with the experiments described in Section \ref{sec: experiment}.

\subsection{Impact of the Initial Rule Form on \ours}
\label{ss: sensitivity analysis s0}

Since \ours\ aims to cover the input space with as many crisp rules as possible, our covering operator is designed to generate initial rules in the crisp form (see Section \ref{ss: covering}). 
In this subsection, we evaluate the implications of this design strategy by extending our covering operator to generate initial rules in a fuzzy form. Specifically, we alter the initial values of the shape parameters defined in \eqref{eq: beta covering} from $\alpha_i^{k_\text{cov}}=\beta_i^{k_\text{cov}}=1$ to $\alpha_i^{k_\text{cov}}=\beta_i^{k_\text{cov}}=U_{[1,s_0]}$. Here, $s_0$ serves as a hyperparameter that dictates the sharpness of the {MF}; a larger $s_0$ yields a sharper {MF}. With this modification, when $s_0>1.0$, the covering operator consistently generates symmetric bell-shaped {MFs} as the initial rule. Conversely, when $s_0=1.0$, {this} operator is functionally equivalent to our original operator.\label{others-9}

The objective of this subsection is to investigate the impact of the covering shape parameter $s_0$ on the test classification accuracy of \ours\ with \ourrep C. Table \ref{tb: s0} presents the results of the Wilcoxon signed-rank test for various $s_0$ settings across the 25 problems used in Section \ref{sec: experiment}. We set $s_0\in\{1.0,1.5,2.0,2.5,3.0,3.5,4.0,4.5,5.0\}$. Additionally, the complete summary of the test classification accuracy is presented in Appendix \ref{sec: sup experimental results on ours with s0}.

\begin{table}[!t]
\begin{center}
\caption{Results of the Wilcoxon Signed-Rank Test for \ours\ With Various $s_0$ Settings: ``$+$'', ``$-$'', and ``$\sim$'' for Significantly Better, Worse, and {Similar} Compared to the Default $s_0=1.0$ Setting in Test Classification Accuracy, Respectively. {Other Notations Follow Table \ref{tb: result}.}}
\label{tb: s0}
\label{others-6}
\normalsize
\resizebox{\linewidth}{!}{
\begin{tabular}{c|c|cccccccc}
\bhline{1pt}
$s_0$ & 1.0 & 1.5 & 2.0 & 2.5 & 3.0 & 3.5 & 4.0 & 4.5 & 5.0
\\
\bhline{1pt}
Rank & \cellcolor{g}\textit{2.74} & \textit{3.32} & \textit{3.46} & \textit{4.56} & \textit{5.26} & \textit{5.32} & \textit{6.64} & \cellcolor{p}\textit{7.14} & \textit{6.56} \\
Position & \textit{1} & \textit{2} & \textit{3} & \textit{4} & \textit{5} & \textit{6} & \textit{8} & \textit{9} & \textit{7} \\
$+/-/\sim$ & - & 0/9/16 & 1/9/15 & 0/11/14 & 0/10/15 & 0/13/12 & 0/11/14 & 0/10/15 & 1/13/11 \\
\bhline{1pt}
$p$-value & - & 0.00390 & 0.00673 & 0.00964 & 0.000963 & 0.0059 & 8.80E-05 & 0.000108 & 0.00281 \\
$p_\text{Holm}$-value & - & 0.0156 & 0.0177 & 0.0177 & 0.00578 & 0.0177 & 0.000704 & 0.000754 & 0.0141 \\
\bhline{1pt}
\end{tabular}
}
\end{center}
\vspace{-2mm}
\end{table}
Table \ref{tb: s0} indicates that the $s_0 = 1.0$ setting achieved the highest average rank. It is also evident that as the value of $s_0$ increases, thereby increasing the sharpness of the {MF} of the initial rule, the average rank tends to decrease.

The hyperparameter $s_0$ determines the shape of the {MF} for the initial rule generated by the covering operator. Specifically, a crisp form corresponds to $s_0 = 1.0$, while a fuzzy form corresponds to $s_0 > 1.0$. Consequently, when $s_0 = 1.0$, crisp rules dominate the rule set in the initial training phase, with fuzzy rules being generated only by genetic operators. Conversely, when $s_0 > 1.0$, fuzzy rules prevail within the rule set, and crisp rules are generated either by {the genetic or crispification operators.} \label{others-10}Notably, the larger the value of $s_0$, the less likely a crisp rule will be generated, as \ours\ utilizes the relative mutation (see Section \ref{ss: genetic operators}). 
{In summary,}
{values of $s_0$ closer to 1.0 increase crisp rule generation, improving \ours's test classification accuracy.}

The analysis suggests a hypothesis: the superior ranking of $s_0=1.0$ is due to its reliance on crisp rules for the majority of the classification, supplemented by fuzzy rules for handling specific difficult subspaces. The next subsection attempts to validate this hypothesis.

\subsection{Visualization of Decision Boundaries and Rule Discovery Strategy in \ours}
\label{ss: visualization of decision boundaries and rule discovery strategy of ours}

\begin{figure*}[!t]
\centering
\subfloat[\scriptsize Original Data]{\includegraphics[width=0.2\linewidth]{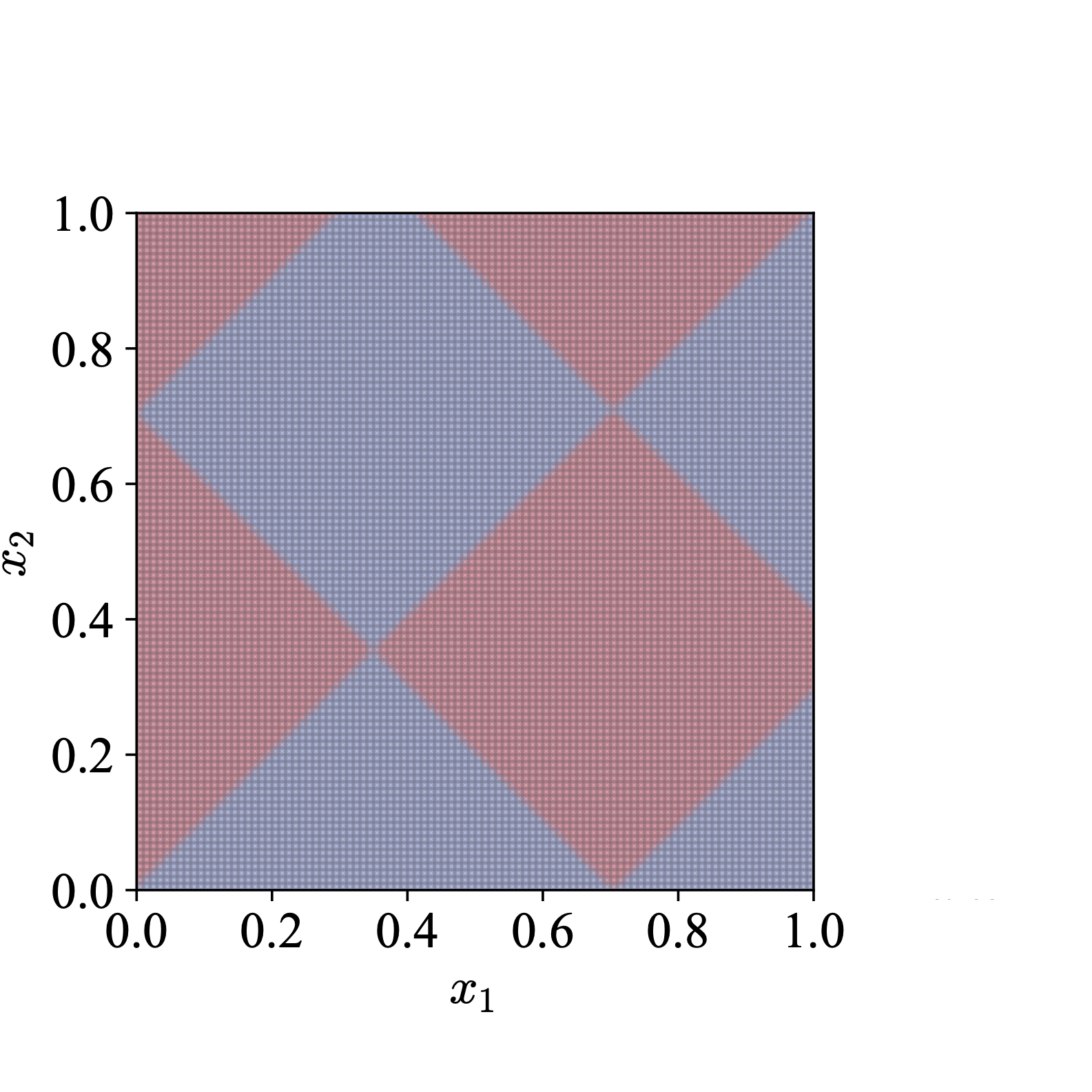}%
\label{fig: rcb_landscape}}
\subfloat[\scriptsize Kurtosis at the 1st epoch]{\includegraphics[width=0.2\linewidth]{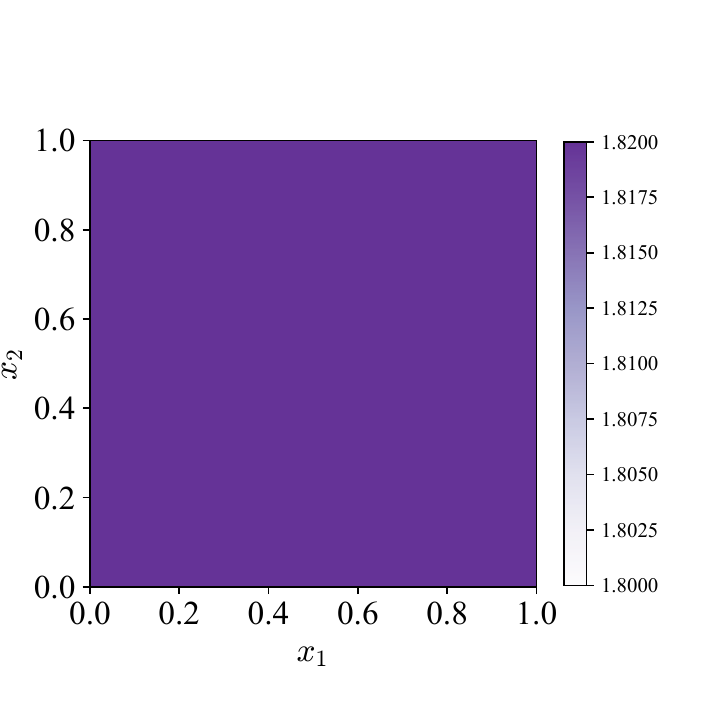}%
\label{fig: rcb_kurtosis_1epoch}}
\subfloat[\scriptsize Kurtosis at the 10th epoch]{\includegraphics[width=0.2\linewidth]{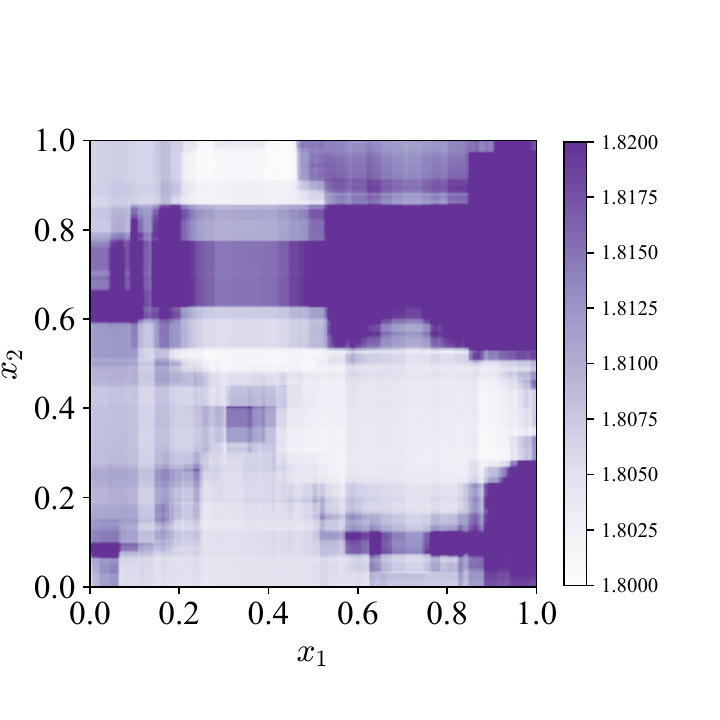}%
\label{fig: rcb_kurtosis_10epoch}}
\subfloat[\scriptsize Kurtosis at the 50th epoch]{\includegraphics[width=0.2\linewidth]{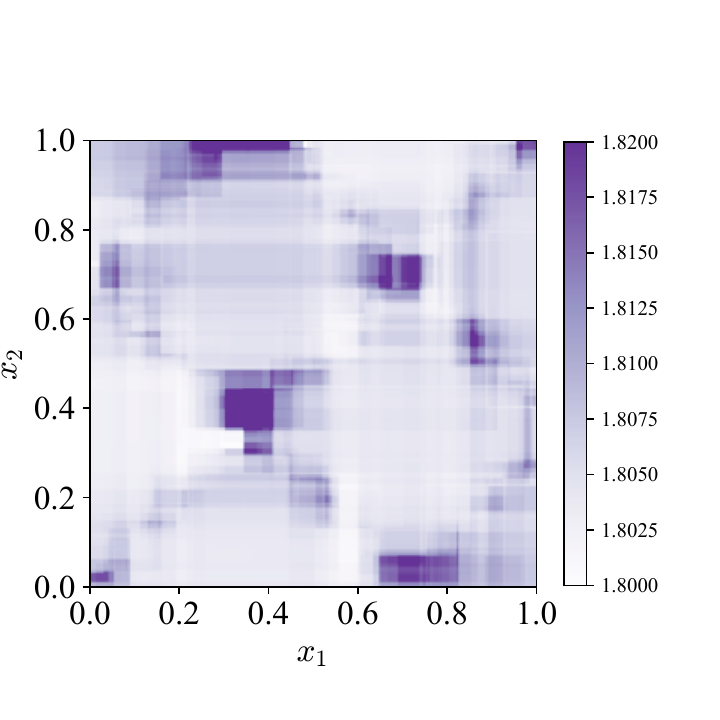}%
\label{fig: rcb_kurtosis_50epoch}}
\subfloat[\scriptsize Boundaries at the 50th epoch]{\includegraphics[width=0.2\linewidth]{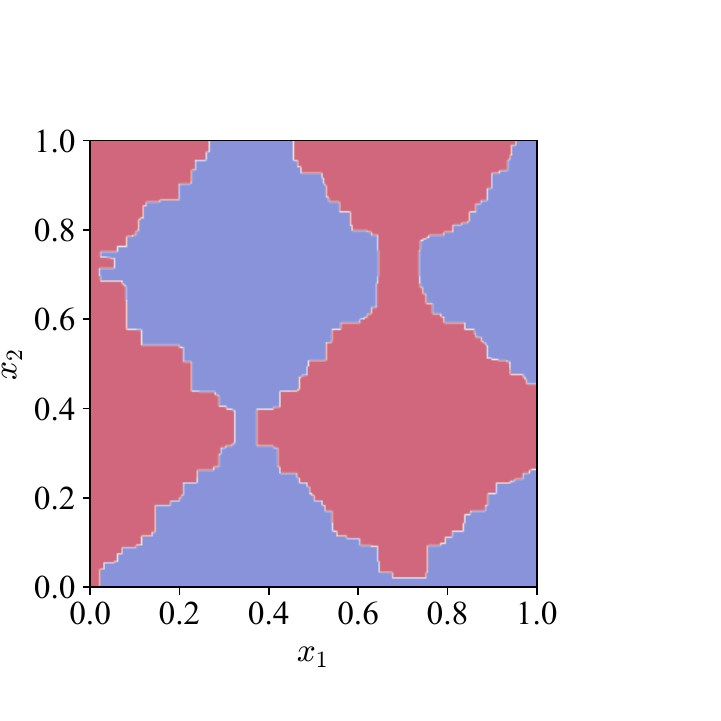}%
\label{fig: rcb_answer}}
\caption{(a) shows the rotated checkerboard problem. (b), (c) and (d) display the rule-kurtosis landscapes at the 1st, 10th, and 50th epoch, respectively, obtained by \ours. (e) displays the decision boundaries at the 50th epoch, created using \ours.}
\label{fig: all_analysis_landscape}
\vspace{-2mm}
\end{figure*}

This subsection focuses on the visual analysis of the decision boundaries created by \ours\ with \ourrep C. We utilize an artificial dataset known as the rotated checkerboard problem \cite{shiraishi2022absumption}. As depicted in Fig. \ref{fig: rcb_landscape}, the dataset is designed for a binary classification task and consists of 10,201 two-dimensional data points. Points in class $c_1$ are colored blue, and points in class $c_2$ are colored red. This dataset includes a mix of subspaces: some that can be easily classified using the hyperrectangular crisp rules generated by \ours, particularly those distant from the oblique decision boundaries, and others that cannot, especially those near these oblique boundaries. Therefore, this dataset is ideal for testing the hypothesis introduced in Section \ref{ss: sensitivity analysis s0} about the effectiveness of the $s_0=1.0$ setting.

Figs.~\ref{fig: rcb_kurtosis_1epoch},~\ref{fig: rcb_kurtosis_10epoch}, and \ref{fig: rcb_kurtosis_50epoch} illustrate the average kurtosis of all rules used during class inference in a randomly selected run at the 1st, 10th, and 50th epoch of training, respectively. Here, kurtosis is computed using the following equation:
\begin{equation}
    {\rm kurtosis}=\frac{\sum_{k\in[M_\text{exp}]}{({\rm kurtosis}_1^k+{\rm kurtosis}_2^k)\cdot {\rm num}^k}}{2\cdot \sum_{k\in[M_\text{exp}]}{{\rm num}^k}},
\end{equation}
where $[M_\text{exp}]=\{k\in[M] \mid exp^k>\theta_\text{exp}\}$ is the set of rules used during class inference. If all rules are crisp-rectangular, the kurtosis is 1.8. As the number of fuzzy rules with high sharpness increases, the kurtosis also increases. Fig. \ref{fig: rcb_answer} shows the class assignments $\hat{c}_{[P]}\in\{c_1, c_2\}$ in {the}\label{others-7} run, with blue and red indicating the two classes (resolution: $1000\times 1000$). Over 30 runs with different seeds, \ours\ achieved an average training accuracy of 95.68\%.
As mentioned in Section \ref{ss: sensitivity analysis s0}, 
under the $s_0=1.0$ setting, 
most of the rule set comprises crisp-rectangular rules.
Thus, as Fig. \ref{fig: rcb_answer} illustrates, most of the decision boundaries formed by \ours\ are axis-parallel.

Figs. \ref{fig: rcb_kurtosis_1epoch}, \ref{fig: rcb_kurtosis_10epoch}, and \ref{fig: rcb_kurtosis_50epoch} provide an overview of the rule discovery strategy of \ours. Initially, in the early stage of learning, \ours\ performs a global search for rules using the genetic operator. During this phase, \ours\ aims to cover the entire input space with crisp rules generated by the covering operator and fuzzy rules generated by the genetic operator. As a result, the kurtosis of all subspaces is greater than 1.8, as shown in Fig. \ref{fig: rcb_kurtosis_1epoch} (i.e., the 1st epoch). In the middle stage of learning, after the entire input space is sufficiently covered by rules, \ours\ attempts to cover as many subspaces as possible with crisp rules that are easy to classify, using both the genetic operator and the crispification operator. Consequently, as shown in Fig. \ref{fig: rcb_kurtosis_10epoch} (i.e., the 10th epoch), the kurtosis in those subspaces (e.g., near $(0.7, 0.4)$, away from the diagonal class boundary) tends to approach 1.8. Finally, at the end of the learning phase, the kurtosis of rules covering subspaces that are difficult to classify with crisp rules (e.g., on or near the diagonal class boundary) tends to be greater than 1.8, as shown in Fig. \ref{fig: rcb_kurtosis_50epoch} (i.e., the 50th epoch).

\begin{figure}[t]
\centerline{\includegraphics[width=\linewidth]{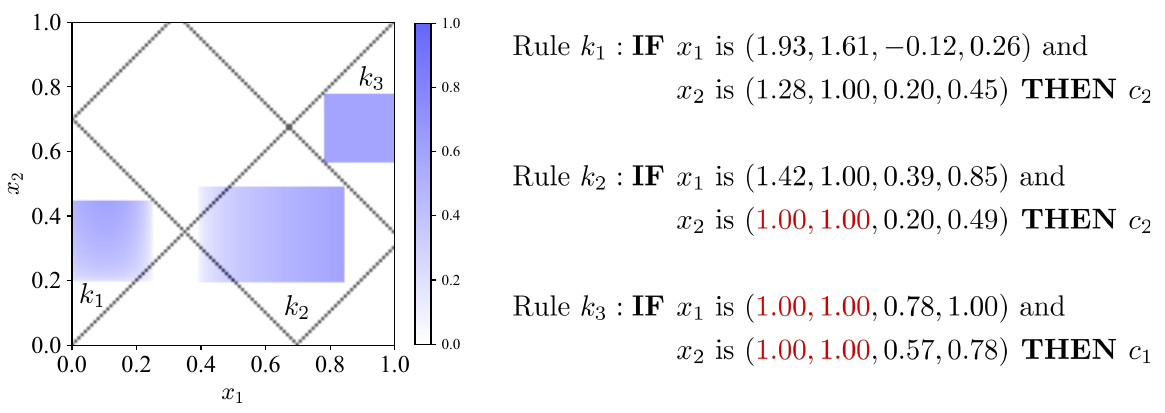}}
\caption{Examples of the matching degree landscapes of rules generated by \ours\ for the rotated checkerboard problem. The diagonal line represents the class boundary. Rectangular MFs (i.e., $\alpha_i^k=\beta_i^k=1.00$) are in red.}
\label{fig: rules in RCB}
\vspace{-2mm}
\end{figure}
{Fig. \ref{fig: rules in RCB} shows rules generated by \ours\ for the problem, demonstrating adaptive rule representation. The system uses fuzzy rules ($k_1$, $k_2$) near diagonal class boundaries where classification is challenging, and crisp rules ($k_3$) in subspaces far from boundaries. This adaptation balances accuracy and interpretability by using non-rectangular {MFs} ($\alpha_i^k, \beta_i^k>1$) for complex subspaces and rectangular functions ($\alpha_i^k=\beta_i^k=1$) for simpler subspaces.}

These results support the hypotheses presented in the previous subsection. Specifically, we can confirm that \ours\ can effectively self-adapt rule representations according to the classification difficulty of each subspace. {Furthermore, Appendix \ref{sec: sup analysis of rule representation adaptation across dataset} demonstrates that \ours\ appropriately adapts its rule representations (choosing between crisp and fuzzy) across different datasets.}

\section{Concluding Remarks}
\label{sec: concluding remarks}

In this {article}, we proposed a novel rule representation for LCSs using a four-parameter beta distribution, termed \ourrep. Subsequently, we introduced \ours, an LCS that optimizes fuzzy rules employing \ourrep.
\ourrep\ defines the membership function ({MF}) of a fuzzy set as a probability density function derived from a four-parameter beta distribution, normalized to the range $[0,1]$. This function can represent various {MF} shapes, from crisp-rectangular to fuzzy-monotonic and fuzzy-symmetric/asymmetric bell-shaped. The genetic operator in \ours\ optimizes the parameters of each fuzzy set in the condition part (i.e., if part) of each fuzzy rule, adjusting both shape and fuzziness simultaneously. 
{Furthermore, we proposed a crispification operator for \ours\ to enhance the generalization bias toward crisp rules.}

We evaluated the performance of \ourrep\ and \ourrep\ with the crispification operator, comparing them with eight existing rule representations. While the performance of existing representations varied depending on the problem, \ourrep\ consistently outperformed them. This superiority is attributed to its adaptive and flexible {MF} representation, which allows smooth transitions between crisp rectangular and various fuzzy shapes. Furthermore, the crispification operator enhances the robustness of \ourrep. Essentially, \ours\ serves as an intermediary between crisp and fuzzy rule-based systems. Previously, users faced the challenge of selecting an appropriate rule representation for unknown problems through trial and error. Our experimental results demonstrate that the proposed algorithm alleviates this difficulty in fuzzy system design.

{Future work will focus on several key areas. First, improving computational efficiency is essential since \ours\ was tested on relatively small datasets due to computational limitations.}\label{r1-2}
{Second, we investigate the balance between crisp and fuzzy rules during evolution and in final solutions, particularly how absumption and/or generational compaction affect evolutionary progress and address over-generalization in crisp rules. 
Third, drawing on the framework of the scalable LCS (i.e., ExSTraCS 2.0 \cite{urbanowicz2015exstracs}), we integrate expert knowledge to enhance rule discovery and optimization in \ours.
{Finally, \ourrep\ shows potential for online learning scenarios with evolving data distributions (e.g., concept drift problems \cite{hinder2023model} such as stock markets). In these cases, FBR can dynamically adapt to changing patterns. This ability could help balance precision and robustness through both crisp and fuzzy rules, in applications such as robotic control \cite{omrane2016fuzzy}.}

\begin{appendices}
\section{Comparison of \ours\ to Modern Machine Learning Techniques}
\label{sec: sup comparison with modern machine learning methods}
{To further evaluate our proposed algorithm's performance against modern machine learning methods, we conduct additional experiments comparing \ours\ with FBRC, Random Forest, SVM, and XGBoost. The experimental setup follows the conditions outlined in Section \ref{sec: experiment}. We use standard implementations of Random Forest\footnote{\url{https://scikit-learn.org/dev/modules/generated/sklearn.ensemble.RandomForestClassifier.html}}, SVM\footnote{\url{https://scikit-learn.org/1.5/modules/generated/sklearn.svm.SVC.html}}, and XGBoost\footnote{\url{https://xgboost.readthedocs.io/en/stable/python/index.html}}, ensuring consistent dataset splits for fair comparison. For Random Forest, SVM, and XGBoost, we use default hyperparameter settings. For \ours, the number of training iterations is set to 50 and 200 epochs.

Table \ref{tb: sup sota} presents each system's average training and test classification accuracy, respectively, and average rank across all datasets. From Table \ref{tb: sup sota}, Random Forest, XGBoost, and \ours\ with 200 epochs ranked first, second, and third in average rank during both training and testing phases. Notably, during training, Random Forest and XGBoost achieved 100\% classification accuracy on many datasets. During testing, \ours\ with 200 epochs showed no significant differences compared to {Random Forest ($p=0.0588$)}, SVM ($p={0.210}$), and XGBoost ($p={0.210}$). 

While these modern standard ML methods achieved similar or higher accuracy compared to \ours, they lack the interpretability offered by our rule-based approach. For example, on the Wine dataset, \ours\ produced clear crisp rules such as:
\begin{align*}
    \textbf{IF} &\textit{ Alkalinity of ash} \in [28.35,28.66] \text{ and}\\ &\textit{ Nonflavanoid phenols}\in[0.240,0.660] \textbf{ THEN} \text{ Class 1},
\end{align*}
providing users with interpretable insights. Note that this rule was obtained by \ours\ (with numerical values shown in their denormalized form). In contrast, the black-box nature of Random Forest, SVM, and XGBoost makes their decisions difficult to interpret and validate, particularly in sensitive applications requiring transparency. For reference, a recent survey on explainable artificial intelligence (XAI) \cite{ortigossa2024explainable} emphasizes that while methods like Random Forest and XGBoost offer superior learning performance compared to rule-based systems (such as LCSs), they provide significantly lower model interpretability. This underscores the trade-off between performance and interpretability in machine learning models.

Our \ours\ strikes a balance between these competing objectives. While it may not achieve the same classification accuracy as Random Forest or XGBoost on some datasets, it offers enhanced interpretability by crisp rules. These features are crucial in domains where decision transparency is crucial, such as healthcare or finance. }

 \begin{table*}[ht]
\begin{center}
\caption{Summary of Results, Displaying Average Training and Test Classification Accuracy Across 30 Runs. Green-Shaded Values Denote the Best Values, While Peach-Shaded Values Indicate the Worst Values. ``Rank'' and ``Position'' Denote Each System’s Overall Average Rank Obtained by Using the Friedman Test and Its Position in the Final Ranking, Respectively. Statistical Results of the Wilcoxon Signed-Rank Test Are Summarized With Symbols Where ``$+$'' and ``$-$'' Indicate Significantly Better and Worse Performance Than \ours\ With 200 Epochs, Respectively, While ``$\sim$'' Indicates No Significant Difference From \ours\ With 200 Epochs. The ``$p$-Value'' Row Shows the Raw $p$-Values From These Comparisons, While the ``$p_\text{Holm}$-Value'' Row Shows the Adjusted $p$-Values Using Holm's Correction Method to Control for Multiple Comparisons.
}
\label{tb: sup sota}
\normalsize
\resizebox{\textwidth}{!}
{
\begin{tabular}{c|ccccc|ccccc}
\bhline{1pt}
& \multicolumn{5}{c|}{\textsc{Training Accuracy} (\%)} & \multicolumn{5}{c}{\textsc{Test Accuracy (\%)}}\\
&\multirow{2}{*}{Random Forest} & \multirow{2}{*}{SVM} & \multirow{2}{*}{XGBoost} & \multicolumn{2}{c|}{\ours} &\multirow{2}{*}{Random Forest} & \multirow{2}{*}{SVM} & \multirow{2}{*}{XGBoost} & \multicolumn{2}{c}{\ours}\\
& & & & 50 Epochs & 200 Epochs & & & & 50 Epochs & 200 Epochs\\
 \bhline{1pt}
{\texttt{cae}} & \cellcolor{g}95.51 $+$ & \cellcolor{p}66.57 $-$ & 92.08 $+$ & 74.12 $-$ & 77.69 & 57.92 $\sim$ & \cellcolor{g}62.50 $\sim$ & \cellcolor{p}52.08 $\sim$ & 62.08 $\sim$ & 61.67 \\
\texttt{can} & \cellcolor{g}100.0 $+$ & 98.08 $-$ & \cellcolor{g}100.0 $+$ & \cellcolor{p}97.54 $-$ & 98.83 & 96.20 $\sim$ & \cellcolor{g}97.49 $+$ & 96.78 $+$ & 95.32 $\sim$ & \cellcolor{p}95.03 \\
\texttt{car} & \cellcolor{g}100.0 $+$ & \cellcolor{p}83.99 $-$ & \cellcolor{g}100.0 $+$ & 95.29 $-$ & 96.10 & 96.98 $+$ & \cellcolor{p}82.98 $-$ & \cellcolor{g}98.56 $+$ & 94.58 $-$ & 95.11 \\
\texttt{chk} & \cellcolor{g}100.0 $+$ & \cellcolor{p}50.82 $-$ & 93.85 $+$ & 85.02 $-$ & 89.37 & 76.87 $-$ & \cellcolor{p}49.62 $-$ & 69.22 $-$ & 81.32 $-$ & \cellcolor{g}86.43 \\
\texttt{cmx} & \cellcolor{g}100.0 $+$ & \cellcolor{p}42.76 $-$ & \cellcolor{g}100.0 $+$ & 95.17 $-$ & 96.59 & \cellcolor{g}99.31 $+$ & \cellcolor{p}42.83 $-$ & 98.98 $+$ & 92.86 $-$ & 94.19 \\
\texttt{col} & \cellcolor{g}100.0 $+$ & \cellcolor{p}77.71 $-$ & \cellcolor{g}100.0 $+$ & 83.56 $-$ & 91.64 & \cellcolor{g}83.44 $\sim$ & 78.06 $\sim$ & \cellcolor{g}83.44 $\sim$ & \cellcolor{p}77.63 $-$ & 80.86 \\
\texttt{dbt} & \cellcolor{g}100.0 $+$ & \cellcolor{p}77.75 $-$ & \cellcolor{g}100.0 $+$ & 81.05 $-$ & 87.81 & \cellcolor{g}75.67 $\sim$ & 75.37 $\sim$ & \cellcolor{p}73.07 $\sim$ & 74.89 $\sim$ & 74.85 \\
\texttt{ecl} & 99.98 $+$ & \cellcolor{p}86.35 $-$ & \cellcolor{g}100.0 $+$ & 89.91 $-$ & 95.08 & \cellcolor{g}86.47 $\sim$ & \cellcolor{p}82.84 $\sim$ & 84.61 $\sim$ & 85.39 $\sim$ & 86.08 \\
\texttt{frt} & \cellcolor{g}100.0 $+$ & \cellcolor{p}92.33 $-$ & \cellcolor{g}100.0 $+$ & 93.90 $-$ & 96.27 & 89.00 $+$ & \cellcolor{g}91.41 $+$ & 89.81 $+$ & \cellcolor{p}85.19 $\sim$ & 85.41 \\
\texttt{gls} & 99.98 $+$ & \cellcolor{p}60.31 $-$ & \cellcolor{g}100.0 $+$ & 85.38 $-$ & 97.67 & \cellcolor{g}80.00 $+$ & \cellcolor{p}55.61 $-$ & 77.58 $+$ & 67.73 $-$ & 71.52 \\
\texttt{hcl} & \cellcolor{g}100.0 $+$ & \cellcolor{p}86.79 $-$ & \cellcolor{g}100.0 $+$ & 91.58 $-$ & 96.35 & 84.32 $+$ & 83.87 $+$ & \cellcolor{g}84.41 $+$ & \cellcolor{p}70.90 $-$ & 74.23 \\
\texttt{mam} & \cellcolor{g}94.65 $+$ & \cellcolor{p}78.82 $-$ & 92.99 $+$ & 82.80 $-$ & 84.07 & 80.21 $\sim$ & \cellcolor{p}78.14 $-$ & 79.45 $-$ & 81.41 $\sim$ & \cellcolor{g}81.86 \\
\texttt{mop} & \cellcolor{g}100.0 $+$ & \cellcolor{p}84.16 $-$ & \cellcolor{g}100.0 $+$ & 93.73 $-$ & 94.26 & 92.49 $\sim$ & \cellcolor{p}84.71 $-$ & \cellcolor{g}97.34 $+$ & 91.71 $\sim$ & 92.02 \\
{\texttt{mul}} & \cellcolor{g}100.0 $\sim$ & \cellcolor{g}100.0 $\sim$ & \cellcolor{g}100.0 $\sim$ & \cellcolor{p}99.48 $-$ & 99.99 & \cellcolor{g}100.0 $+$ & \cellcolor{g}100.0 $+$ & \cellcolor{g}100.0 $+$ & \cellcolor{p}86.55 $-$ & 94.17 \\
\texttt{mux} & \cellcolor{g}100.0 $+$ & \cellcolor{p}58.88 $-$ & \cellcolor{g}100.0 $+$ & 97.59 $-$ & 98.69 & 72.14 $-$ & \cellcolor{p}58.12 $-$ & 90.81 $-$ & 96.44 $-$ & \cellcolor{g}97.50 \\
{\texttt{nph}} & \cellcolor{g}96.23 $+$ & \cellcolor{p}57.59 $-$ & 91.33 $+$ & 76.67 $-$ & 83.15 & 53.15 $\sim$ & 52.82 $\sim$ & \cellcolor{p}51.34 $\sim$ & 53.98 $\sim$ & \cellcolor{g}54.21 \\
\texttt{pdy} & \cellcolor{g}100.0 $+$ & \cellcolor{p}86.66 $-$ & 99.95 $+$ & 90.26 $-$ & 90.53 & \cellcolor{g}95.19 $+$ & \cellcolor{p}86.68 $-$ & 94.69 $+$ & 89.02 $\sim$ & 89.17 \\
\texttt{pis} & 99.99 $+$ & 87.40 $-$ & \cellcolor{g}100.0 $+$ & \cellcolor{p}87.38 $-$ & 88.44 & 86.88 $\sim$ & 86.82 $\sim$ & \cellcolor{g}87.35 $\sim$ & \cellcolor{p}86.60 $\sim$ & 86.70 \\
\texttt{pmp} & \cellcolor{g}100.0 $+$ & \cellcolor{p}87.25 $-$ & 99.96 $+$ & 88.11 $-$ & 88.79 & \cellcolor{g}89.05 $+$ & 87.87 $\sim$ & 87.95 $\sim$ & \cellcolor{p}87.19 $\sim$ & 87.27 \\
\texttt{rsn} & \cellcolor{g}100.0 $+$ & 86.84 $\sim$ & 99.91 $+$ & \cellcolor{p}86.00 $\sim$ & 86.44 & 85.15 $\sim$ & \cellcolor{g}85.74 $\sim$ & \cellcolor{p}84.74 $\sim$ & 85.52 $\sim$ & 85.56 \\
\texttt{soy} & \cellcolor{g}99.87 $+$ & 96.32 $+$ & \cellcolor{g}99.87 $+$ & \cellcolor{p}84.74 $-$ & 86.00 & 93.24 $+$ & \cellcolor{g}93.77 $+$ & 92.85 $+$ & \cellcolor{p}69.52 $\sim$ & 69.76 \\
\texttt{tae} & \cellcolor{g}97.04 $+$ & \cellcolor{p}55.48 $-$ & 96.54 $+$ & 64.00 $-$ & 89.21 & 60.21 $\sim$ & \cellcolor{p}49.58 $-$ & 59.38 $\sim$ & 51.04 $-$ & \cellcolor{g}62.29 \\
\texttt{wne} & \cellcolor{g}100.0 $+$ & 99.38 $\sim$ & \cellcolor{g}100.0 $+$ & \cellcolor{p}97.42 $-$ & 99.12 & \cellcolor{g}97.59 $+$ & \cellcolor{g}97.59 $+$ & 96.67 $\sim$ & \cellcolor{p}92.04 $-$ & 94.63 \\
\texttt{wpb} & \cellcolor{g}100.0 $+$ & \cellcolor{p}79.78 $-$ & \cellcolor{g}100.0 $+$ & 96.57 $-$ & 99.49 & 79.17 $\sim$ & 77.67 $\sim$ & \cellcolor{g}79.50 $\sim$ & \cellcolor{p}73.83 $\sim$ & 76.17 \\
\texttt{yst} & \cellcolor{g}100.0 $+$ & \cellcolor{p}57.80 $-$ & 99.90 $+$ & 68.24 $-$ & 72.69 & \cellcolor{g}61.59 $+$ & \cellcolor{p}57.25 $\sim$ & 58.97 $\sim$ & 59.71 $\sim$ & 58.43 \\
\bhline{1pt}
Rank & \cellcolor{g}\textit{1.40} & \cellcolor{p}\textit{4.56} & \textit{1.64} & \textit{4.24} & \textit{3.16} & \cellcolor{g}\textit{2.12} & \textit{3.42} & \textit{2.70} & \cellcolor{p}\textit{3.76} & \textit{3.00} \\
Position & \textit{1} & \textit{5} & \textit{2} & \textit{4} & \textit{3} & \textit{1} & \textit{4} & \textit{2} & \textit{5} & \textit{3} \\
$+/-/\sim$ & 24/0/1 & 1/21/3 & 24/0/1 & 0/24/1 & - & 11/2/12 & 6/9/10 & 10/3/12 & 0/10/15 & - \\
\bhline{1pt}
$p$-value & 5.96E-08 & 2.21E-05 & 5.96E-08 & 5.96E-08 & - & 0.0588 & 0.210 & 0.210 & 0.000376 & - \\
$p_\text{Holm}$-value & 2.38E-07 & 2.21E-05 & 2.38E-07 & 2.38E-07 & - & 0.176 & 0.420 & 0.420 & 0.00151 & - \\

\bhline{1pt}
\end{tabular}}
\end{center}
\end{table*}

\clearpage
\onecolumn

\section{Impact of the Subsumption Tolerance ${\rm Tol}_\text{\rm sub}$ on \ours}
\label{sec: sup impact of the subsumption tolerance tolsub}

{The subsumption tolerance parameter, ${\rm Tol}_\text{sub}$, is employed to compare the similarity of the modes of two rules (see Section \ref{ss: subsumption operator}). The higher its value, the more likely a rule $k_\text{sub}$ is judged to be more general than a rule $k_\text{tos}$, even if the similarity between their modes is minimal \cite{tadokoro2021xcs}. 

The objective of this subsection is to investigate the impact of ${\rm Tol}_\text{sub}$ on the test classification accuracy of \ours\ with \ourrep C. Table \ref{tb: tolsub_test} presents the test classification accuracy of \ours\ with \ourrep C for various ${\rm Tol}_\text{\rm sub}$ settings (${\rm Tol}_\text{\rm sub}\in\{0.0, 0.01, 0.05, 0.1, 0.2, 0.4, 0.6, 0.8, 1.0\}$) across all datasets.

The Friedman test indicates no significant differences among the settings. Two reasons can explain this observation. First, \ours\ predominantly employs crisp rules. As discussed in Section \ref{ss: subsumption operator}, ${\rm Tol}_\text{sub}$ is utilized only for Condition 3 in the is-more-general operator, applicable exclusively when both rules are fuzzy. Consequently, in \ours, where most of the rule set comprises crisp rules, Condition 3 is infrequently used.
Second, Conditions 1 and 2 are sufficient to evaluate the inclusion relation between two rules. Specifically, Condition 1 ensures that the subspace covered by $k_\text{sub}$ encompasses the subspace covered by $k_\text{tos}$ (see \eqref{eq: subsumption-condition1}), and Condition 2 ensures that the membership function of $k_\text{sub}$ has a more rounded peak than that of $k_\text{tos}$ (see \eqref{eq: subsumption-condition2}). Thus, even with a low mode similarity between $ {k_\text{sub}} $ and $ {k_\text{tos}} $, as long as Conditions 1 and 2 are satisfied, removing $ {k_\text{tos}} $ by subsumption does not significantly affect system performance. This is because $ {k_\text{sub}} $ can completely cover the subspace formerly covered by $ {k_\text{tos}} $ and maintain a certain level of matching degree formerly held by $ {k_\text{tos}} $, as illustrated in Fig. \ref{fig: subsumption-condition3}. This perspective is consistent with the results in Table \ref{tb: tolsub_test}, where varying ${\rm Tol}_\text{sub}$ does not lead to significant changes in test classification accuracy.

The conclusion in this section is that ${\rm Tol}_\text{sub}$ is a hyperparameter with low sensitivity. Changing ${\rm Tol}_\text{sub}$ from its default value of 0.01 does not significantly impact classification performance, as indicated by the limited occurrence of significant differences (denoted by “$+$” and “$-$” symbols) in only up to 3 of the 25 problems.}

\begin{table*}[ht]
\begin{center}
\caption{Summary of Results, Displaying Average Test Classification Accuracy of \ours\ With Various ${\rm Tol}_\text{\rm sub}$ Settings Across 30 Runs. ``$+$'', ``$-$'', and ``$\sim$'' for Significantly Better, Worse, and Competitive Compared to the Default ${\rm Tol}_\text{\rm sub}=0.01$ Setting in Test Classification Accuracy, Respectively. ``Rank''; ``Position''; ``$p$-Value''; ``$p_\text{Holm}$-Value''; and Green- and Peach-Shaded Values Should be Interpreted as in Table \ref{tb: sup sota}.}
\label{tb: tolsub_test}
\normalsize
\scalebox{0.825}{
\begin{tabular}{c|c|cccccccc}
\bhline{1pt}
${\rm Tol}_\text{sub}$ & 0.01 & 0.0 & 0.05 & 0.1 & 0.2 & 0.4 & 0.6 & 0.8 & 1.0
\\
\bhline{1pt}
{\texttt{cae}} & \cellcolor{g}62.08 & \cellcolor{g}62.08 $\sim$ & \cellcolor{g}62.08 $\sim$ & \cellcolor{g}62.08 $\sim$ & \cellcolor{p}61.67 $\sim$ & \cellcolor{g}62.08 $\sim$ & \cellcolor{p}61.67 $\sim$ & \cellcolor{p}61.67 $\sim$ & \cellcolor{p}61.67 $\sim$ \\
\texttt{can} & 95.32 & 95.20 $\sim$ & \cellcolor{g}95.67 $\sim$ & \cellcolor{g}95.67 $\sim$ & 95.56 $\sim$ & 95.56 $\sim$ & 95.20 $\sim$ & 95.15 $\sim$ & \cellcolor{p}94.85 $\sim$ \\
\texttt{car} & 94.58 & 94.56 $\sim$ & 94.39 $\sim$ & \cellcolor{p}94.23 $\sim$ & 94.34 $\sim$ & 94.27 $\sim$ & \cellcolor{g}94.70 $\sim$ & 94.69 $\sim$ & 94.46 $\sim$ \\
\texttt{chk} & 81.32 & 81.96 $\sim$ & \cellcolor{p}81.28 $\sim$ & 81.85 $\sim$ & \cellcolor{g}82.16 $\sim$ & 81.94 $\sim$ & 82.09 $\sim$ & 82.10 $\sim$ & 82.14 $\sim$ \\
\texttt{cmx} & 92.86 & 92.80 $\sim$ & \cellcolor{g}93.16 $\sim$ & 92.83 $\sim$ & 92.81 $\sim$ & 92.77 $\sim$ & 92.78 $\sim$ & 93.06 $\sim$ & \cellcolor{p}92.73 $\sim$ \\
\texttt{col} & 77.63 & \cellcolor{p}77.31 $\sim$ & 77.63 $\sim$ & 77.42 $\sim$ & 77.42 $\sim$ & 77.53 $\sim$ & 77.63 $\sim$ & \cellcolor{g}77.96 $\sim$ & \cellcolor{g}77.96 $\sim$ \\
\texttt{dbt} & 74.89 & 74.20 $-$ & 74.94 $\sim$ & \cellcolor{g}75.02 $\sim$ & 74.29 $\sim$ & 74.16 $\sim$ & \cellcolor{p}73.98 $\sim$ & 74.50 $\sim$ & 74.72 $\sim$ \\
\texttt{ecl} & \cellcolor{p}85.39 & 85.78 $\sim$ & 85.88 $\sim$ & \cellcolor{g}86.57 $\sim$ & 86.08 $\sim$ & 86.08 $\sim$ & 85.78 $\sim$ & 85.69 $\sim$ & 85.88 $\sim$ \\
\texttt{frt} & 85.19 & 85.70 $\sim$ & 85.37 $\sim$ & \cellcolor{g}85.81 $\sim$ & 85.78 $\sim$ & \cellcolor{p}85.07 $\sim$ & 85.33 $\sim$ & 85.78 $\sim$ & 85.59 $\sim$ \\
\texttt{gls} & \cellcolor{p}67.73 & 68.33 $\sim$ & 68.18 $\sim$ & 68.18 $\sim$ & 68.64 $\sim$ & \cellcolor{p}67.73 $\sim$ & 68.79 $\sim$ & \cellcolor{g}70.00 $+$ & \cellcolor{g}70.00 $+$ \\
\texttt{hcl} & 70.90 & 71.71 $\sim$ & 70.90 $\sim$ & \cellcolor{p}70.81 $\sim$ & 71.71 $\sim$ & 73.06 $\sim$ & \cellcolor{g}73.15 $+$ & 72.52 $\sim$ & 71.62 $\sim$ \\
\texttt{mam} & \cellcolor{g}81.41 & 81.37 $\sim$ & 81.31 $\sim$ & \cellcolor{p}81.20 $\sim$ & 81.34 $\sim$ & 81.37 $\sim$ & 81.34 $\sim$ & 81.34 $\sim$ & 81.31 $\sim$ \\
\texttt{mop} & \cellcolor{g}91.71 & 91.67 $\sim$ & 91.58 $\sim$ & 91.54 $\sim$ & 91.52 $\sim$ & 91.44 $\sim$ & \cellcolor{p}91.24 $\sim$ & 91.44 $\sim$ & 91.42 $\sim$ \\
{\texttt{mul}} & \cellcolor{g}86.55 & 86.31 $\sim$ & 85.36 $\sim$ & 85.48 $\sim$ & 85.83 $\sim$ & \cellcolor{p}84.52 $\sim$ & 85.36 $\sim$ & 85.12 $\sim$ & 85.83 $\sim$ \\
\texttt{mux} & 96.44 & \cellcolor{p}96.13 $\sim$ & 96.62 $\sim$ & 96.49 $\sim$ & 96.37 $\sim$ & 96.58 $\sim$ & \cellcolor{g}96.77 $\sim$ & 96.54 $\sim$ & 96.67 $\sim$ \\
{\texttt{nph}} & 53.98 & 54.21 $\sim$ & 53.89 $\sim$ & \cellcolor{p}53.80 $\sim$ & 54.49 $\sim$ & 54.86 $\sim$ & 54.49 $\sim$ & \cellcolor{g}55.09 $\sim$ & 54.40 $\sim$ \\
\texttt{pdy} & 89.02 & 88.96 $\sim$ & 89.08 $\sim$ & 89.17 $\sim$ & 89.27 $\sim$ & 88.98 $\sim$ & 89.24 $\sim$ & \cellcolor{g}89.33 $\sim$ & \cellcolor{p}88.89 $\sim$ \\
\texttt{pis} & 86.60 & 86.48 $\sim$ & 86.16 $-$ & 86.14 $\sim$ & 86.25 $\sim$ & \cellcolor{g}86.74 $\sim$ & 86.45 $\sim$ & 86.28 $\sim$ & \cellcolor{p}86.05 $\sim$ \\
\texttt{pmp} & 87.19 & \cellcolor{g}87.23 $\sim$ & 87.05 $\sim$ & 86.97 $\sim$ & 87.00 $\sim$ & \cellcolor{p}86.59 $-$ & 86.69 $\sim$ & 87.04 $\sim$ & 87.19 $\sim$ \\
\texttt{rsn} & 85.52 & \cellcolor{g}85.74 $\sim$ & \cellcolor{p}85.26 $\sim$ & 85.41 $\sim$ & 85.59 $\sim$ & 85.44 $\sim$ & 85.44 $\sim$ & 85.44 $\sim$ & 85.48 $\sim$ \\
\texttt{soy} & 69.52 & 69.66 $\sim$ & \cellcolor{g}69.66 $\sim$ & 68.99 $\sim$ & 69.28 $\sim$ & \cellcolor{p}68.74 $\sim$ & 69.52 $\sim$ & 68.94 $\sim$ & 69.13 $\sim$ \\
\texttt{tae} & \cellcolor{p}51.04 & 52.08 $\sim$ & 52.08 $\sim$ & 52.08 $\sim$ & 52.08 $\sim$ & 52.08 $\sim$ & 52.08 $\sim$ & 52.71 $\sim$ & \cellcolor{g}52.92 $+$ \\
\texttt{wne} & 92.04 & 92.04 $\sim$ & \cellcolor{p}91.85 $\sim$ & 92.04 $\sim$ & 92.41 $\sim$ & 92.78 $\sim$ & 92.96 $\sim$ & \cellcolor{g}93.15 $+$ & \cellcolor{g}93.15 $+$ \\
\texttt{wpb} & \cellcolor{g}73.83 & \cellcolor{g}73.83 $\sim$ & \cellcolor{g}73.83 $\sim$ & 73.67 $\sim$ & 73.33 $\sim$ & \cellcolor{p}73.00 $\sim$ & \cellcolor{p}73.00 $\sim$ & \cellcolor{g}73.83 $\sim$ & 73.50 $\sim$ \\
\texttt{yst} & 59.71 & 59.51 $\sim$ & 59.64 $\sim$ & 59.75 $\sim$ & 58.48 $\sim$ & \cellcolor{p}58.43 $-$ & 59.24 $\sim$ & \cellcolor{g}59.84 $\sim$ & 59.40 $\sim$ \\
\bhline{1pt}
Rank & \textit{4.72} & \textit{4.72} & \textit{5.00} & \textit{5.50} & \textit{4.88} & \cellcolor{p}\textit{5.90} & \textit{5.18} & \cellcolor{g}\textit{4.08} & \textit{5.02} \\
Position & \textit{2.5} & \textit{2.5} & \textit{5} & \textit{8} & \textit{4} & \textit{9} & \textit{7} & \textit{1} & \textit{6} \\
$+/-/\sim$ & - & 0/1/24 & 0/1/24 & 0/0/25 & 0/0/25 & 0/2/23 & 1/0/24 & 2/0/23 & 3/0/22 \\
\bhline{1pt}
$p$-value & - & 0.483 & 0.812 & 0.846 & 0.692 & 0.800 & 0.705 & 0.197 & 0.546 \\
$p_\text{Holm}$-value & - & 1.000 & 1.000 & 1.000 & 1.000 & 1.000 & 1.000 & 1.000 & 1.000 \\

\bhline{1pt}
\end{tabular}}
\end{center}
\end{table*}

\clearpage
\section{Impact of the Covering Range Parameter $r_0$ on \ours}
\label{sec: sup impact of the covering range parameter r0}
{The hyperparameter $r_0$ determines the maximum range for initializing intervals in the covering operator, affecting how much input space each newly created rule can cover. When $r_0$ is small, rules cover smaller subspaces of the input space, potentially requiring more rules to achieve good coverage but enabling more precise approximation of decision boundaries with finer granularity.
In contrast, when $r_0$ is large, rules cover larger subspaces of the input space, which may enhance generalization capability while potentially sacrificing accuracy where decision boundaries are complex. 

To investigate the impact of $r_0$, we conduct experiments with various $r_0$ settings across the 25 problems used in Section \ref{sec: experiment}. We set $r_0\in\{0.1, 0.2, 0.4, 0.6, 0.8, 1.0\}$. 

Table \ref{tb: sup result r0} presents the average test classification accuracy and population size of \ours\ with FBRC. The Friedman test indicates no significant differences in test classification accuracy among the settings ($p={0.128}$) but reveals significant differences in population size ($p={3.25\text{E-7}}$). Notably, 
$r_0=1$ ranks highest for population size and demonstrated statistically significant differences compared to all other settings (i.e., $p_\text{Holm}<0.05$). Moreover, a trend is observed where the average rank in population size decreased as the value of 
$r_0$ decreased. These results confirm our understanding of the trade-off: smaller $r_0$ values lead to more precise but numerous rules, while larger values promote generalization with fewer rules. The $r_0=1$ setting achieves the best balance in this trade-off.}

 \begin{table*}[ht]
\begin{center}
\caption{Summary of Results, Displaying Average Test Classification Accuracy of \ours\ With Various $r_0$ Settings Across 30 Runs. Statistical Results of the Wilcoxon Signed-Rank Test Are Summarized With Symbols Where ``$+$'' and ``$-$'' Indicate Significantly Better and Worse Performance Than the Default $r_0=1.0$ Setting, Respectively, While ``$\sim$'' Indicates No Significant Difference From the Setting. 
``Rank''; ``Position''; ``$p$-Value''; ``$p_\text{Holm}$-Value''; and Green- and Peach-Shaded Values Should be Interpreted as in Table \ref{tb: sup sota}.
}
\vspace{2mm}
\label{tb: sup result r0}
\normalsize
\resizebox{\textwidth}{!}{

\begin{tabular}{c|cccccc|cccccc}
\bhline{1pt}
 $r_0$& 0.1 & 0.2 & 0.4 & 0.6 & 0.8 & 1.0 & 0.1 & 0.2 & 0.4 & 0.6 & 0.8 & 1.0 \\
 \bhline{1pt}
{\texttt{cae}} & \cellcolor{p}61.67 $\sim$ & 66.67 $\sim$ & 62.92 $\sim$ & \cellcolor{g}67.50 $\sim$ & \cellcolor{g}67.50 $\sim$ & 62.08 & \cellcolor{p}914.5 $-$ & 882.9 $-$ & 834.0 $-$ & 644.3 $-$ & 486.8 $\sim$ & \cellcolor{g}457.5 \\
\texttt{can} & \cellcolor{p}62.75 $-$ & 63.98 $-$ & 92.16 $-$ & 94.44 $-$ & 94.74 $-$ & \cellcolor{g}95.32 & 1793 $-$ & \cellcolor{p}1846 $-$ & 1817 $-$ & 1781 $-$ & 1759 $-$ & \cellcolor{g}1722 \\
\texttt{car} & \cellcolor{p}94.01 $\sim$ & 94.63 $\sim$ & 94.81 $\sim$ & \cellcolor{g}95.15 $+$ & 94.77 $\sim$ & 94.58 & \cellcolor{p}1511 $-$ & 1483 $-$ & 1471 $-$ & 1430 $-$ & 1411 $\sim$ & \cellcolor{g}1399 \\
\texttt{chk} & \cellcolor{p}81.22 $\sim$ & 81.54 $\sim$ & 81.37 $\sim$ & 81.62 $\sim$ & \cellcolor{g}81.79 $\sim$ & 81.32 & 1360 $\sim$ & \cellcolor{p}1363 $-$ & 1361 $\sim$ & 1359 $\sim$ & 1355 $\sim$ & \cellcolor{g}1354 \\
\texttt{cmx} & \cellcolor{p}12.79 $-$ & 13.28 $-$ & 62.20 $-$ & 89.01 $-$ & 92.21 $-$ & \cellcolor{g}92.86 & \cellcolor{p}1887 $-$ & 1800 $-$ & 1612 $-$ & 1570 $-$ & 1548 $-$ & \cellcolor{g}1525 \\
\texttt{col} & \cellcolor{g}81.83 $+$ & 81.18 $+$ & 81.51 $+$ & 79.46 $\sim$ & 78.39 $\sim$ & \cellcolor{p}77.63 & \cellcolor{p}1458 $-$ & 1425 $-$ & 1387 $-$ & 1343 $-$ & 1335 $\sim$ & \cellcolor{g}1309 \\
\texttt{dbt} & 74.63 $\sim$ & \cellcolor{p}73.90 $\sim$ & \cellcolor{g}76.10 $\sim$ & 74.50 $\sim$ & 74.46 $\sim$ & 74.89 & \cellcolor{p}1568 $-$ & 1565 $-$ & 1552 $-$ & 1530 $\sim$ & 1523 $\sim$ & \cellcolor{g}1519 \\
\texttt{ecl} & \cellcolor{p}83.53 $\sim$ & 86.76 $\sim$ & \cellcolor{g}87.16 $\sim$ & 85.10 $\sim$ & 85.00 $\sim$ & 85.39 & \cellcolor{p}1518 $-$ & 1482 $-$ & 1423 $\sim$ & \cellcolor{g}1405 $\sim$ & 1407 $\sim$ & 1408 \\
\texttt{frt} & \cellcolor{p}7.074 $-$ & 15.93 $-$ & 78.26 $-$ & 83.41 $\sim$ & 84.96 $\sim$ & \cellcolor{g}85.19 & \cellcolor{g}1820 $+$ & 1837 $\sim$ & \cellcolor{p}1875 $-$ & 1867 $-$ & 1850 $\sim$ & 1845 \\
\texttt{gls} & \cellcolor{p}63.33 $-$ & 67.88 $\sim$ & 68.03 $\sim$ & 66.67 $\sim$ & \cellcolor{g}69.09 $\sim$ & 67.73 & \cellcolor{p}1579 $-$ & 1566 $-$ & 1530 $-$ & 1493 $-$ & 1470 $-$ & \cellcolor{g}1440 \\
\texttt{hcl} & \cellcolor{p}58.47 $-$ & 61.35 $-$ & 67.21 $\sim$ & 70.90 $\sim$ & \cellcolor{g}71.35 $\sim$ & 70.90 & 1807 $\sim$ & 1810 $-$ & 1809 $-$ & \cellcolor{p}1812 $-$ & 1800 $\sim$ & \cellcolor{g}1798 \\
\texttt{mam} & \cellcolor{g}82.06 $\sim$ & 81.37 $\sim$ & 81.48 $\sim$ & \cellcolor{p}81.24 $\sim$ & 81.44 $\sim$ & 81.41 & \cellcolor{g}1416 $+$ & 1444 $+$ & 1440 $+$ & 1476 $\sim$ & 1489 $\sim$ & \cellcolor{p}1490 \\
\texttt{mop} & 91.74 $\sim$ & \cellcolor{g}91.81 $\sim$ & 91.79 $\sim$ & \cellcolor{p}91.51 $\sim$ & 91.71 $\sim$ & 91.71 & \cellcolor{p}1617 $-$ & 1611 $-$ & 1602 $\sim$ & 1599 $\sim$ & 1604 $\sim$ & \cellcolor{g}1597 \\
{\texttt{mul}} & 80.00 $-$ & \cellcolor{p}79.76 $-$ & 80.48 $-$ & 83.93 $\sim$ & 83.93 $\sim$ & \cellcolor{g}86.55 & \cellcolor{p}1745 $-$ & 1743 $-$ & 1743 $-$ & 1732 $-$ & 1725 $-$ & \cellcolor{g}1709 \\
\texttt{mux} & \cellcolor{p}49.31 $-$ & \cellcolor{p}49.31 $-$ & 70.64 $-$ & 90.24 $-$ & 95.23 $-$ & \cellcolor{g}96.44 & 1943 $-$ & \cellcolor{p}1945 $-$ & 1805 $-$ & 1737 $-$ & 1634 $-$ & \cellcolor{g}1559 \\
{\texttt{nph}} & 54.44 $\sim$ & 54.03 $\sim$ & 54.49 $\sim$ & \cellcolor{g}54.95 $\sim$ & \cellcolor{p}53.38 $\sim$ & 53.98 & \cellcolor{g}1698 $+$ & 1706 $\sim$ & 1709 $\sim$ & 1704 $\sim$ & 1708 $\sim$ & \cellcolor{p}1712 \\
\texttt{pdy} & 89.22 $\sim$ & \cellcolor{g}89.48 $+$ & 89.31 $\sim$ & 89.17 $\sim$ & 89.27 $\sim$ & \cellcolor{p}89.02 & 1346 $\sim$ & \cellcolor{g}1344 $\sim$ & 1351 $\sim$ & 1347 $\sim$ & 1353 $\sim$ & \cellcolor{p}1353 \\
\texttt{pis} & 86.36 $\sim$ & 86.54 $\sim$ & 86.84 $\sim$ & \cellcolor{g}87.04 $\sim$ & \cellcolor{p}86.28 $\sim$ & 86.60 & \cellcolor{p}1719 $-$ & 1709 $-$ & 1690 $-$ & 1681 $-$ & 1652 $-$ & \cellcolor{g}1636 \\
\texttt{pmp} & 87.60 $\sim$ & \cellcolor{g}87.71 $\sim$ & 87.55 $\sim$ & 87.25 $\sim$ & \cellcolor{p}86.97 $\sim$ & 87.19 & 1628 $-$ & \cellcolor{p}1629 $-$ & 1622 $-$ & 1622 $-$ & 1589 $-$ & \cellcolor{g}1561 \\
\texttt{rsn} & \cellcolor{g}86.70 $+$ & 86.63 $+$ & 86.07 $\sim$ & 86.00 $\sim$ & 85.74 $\sim$ & \cellcolor{p}85.52 & \cellcolor{p}1478 $-$ & 1474 $-$ & 1464 $-$ & 1445 $-$ & 1410 $\sim$ & \cellcolor{g}1394 \\
\texttt{soy} & \cellcolor{p}44.98 $-$ & 47.68 $-$ & 51.06 $-$ & 58.79 $-$ & 63.86 $-$ & \cellcolor{g}69.52 & \cellcolor{g}1858 $+$ & 1873 $\sim$ & 1877 $\sim$ & \cellcolor{p}1880 $-$ & 1874 $\sim$ & 1871 \\
\texttt{tae} & \cellcolor{g}61.67 $+$ & 60.21 $+$ & 57.71 $+$ & 55.00 $\sim$ & 54.58 $\sim$ & \cellcolor{p}51.04 & \cellcolor{p}1321 $-$ & 1298 $-$ & 1262 $-$ & 1210 $-$ & 1125 $\sim$ & \cellcolor{g}1077 \\
\texttt{wne} & \cellcolor{p}37.78 $-$ & 65.19 $-$ & 94.26 $\sim$ & \cellcolor{g}95.74 $+$ & 95.37 $+$ & 92.04 & \cellcolor{p}1683 $-$ & 1673 $-$ & 1619 $-$ & 1525 $-$ & 1435 $-$ & \cellcolor{g}1328 \\
\texttt{wpb} & \cellcolor{g}76.00 $\sim$ & \cellcolor{g}76.00 $\sim$ & 75.83 $\sim$ & 75.17 $\sim$ & 74.50 $\sim$ & \cellcolor{p}73.83 & 1876 $-$ & \cellcolor{p}1883 $-$ & 1880 $-$ & 1876 $-$ & 1864 $-$ & \cellcolor{g}1842 \\
\texttt{yst} & \cellcolor{p}57.52 $-$ & 58.32 $\sim$ & 59.06 $\sim$ & \cellcolor{g}59.78 $\sim$ & 59.42 $\sim$ & 59.71 & \cellcolor{g}1431 $+$ & 1482 $+$ & 1537 $+$ & 1560 $\sim$ & \cellcolor{p}1575 $\sim$ & 1568 \\
 \bhline{1pt}
Rank & \cellcolor{p}\textit{4.32} & \textit{3.68} & \cellcolor{g}\textit{2.92} & \textit{3.16} & \textit{3.32} & \textit{3.60} & \textit{4.40} & \cellcolor{p}\textit{4.56} & \textit{4.12} & \textit{3.24} & \textit{2.72} & \cellcolor{g}\textit{1.96} \\
Position & \textit{6} & \textit{5} & \textit{1} & \textit{2} & \textit{3} & \textit{4} & \textit{5} & \textit{6} & \textit{4} & \textit{3} & \textit{2} & \textit{1} \\
$+/-/\sim$ & 3/10/12 & 4/8/13 & 2/6/17 & 2/4/19 & 1/4/20 & - & 5/17/3 & 2/19/4 & 2/17/6 & 0/17/8 & 0/9/16 & - \\
 \bhline{1pt}
$p$-value & 0.0367 & 0.339 & 0.958 & 1.000 & 0.916 & - & 0.00145 & 0.000287 & 5.39E-05 & 6.37E-05 & 5.25E-06 & - \\
$p_\text{Holm}$-value & 0.183 & 1.000 & 1.000 & 1.000 & 1.000 & - & 0.00145 & 0.000574 & 0.000216 & 0.000216 & 2.62E-05 & - \\

\bhline{1pt}
\end{tabular}}
\end{center}
\end{table*}

\clearpage

\section{Extended Analysis of Synthetic Datasets}
\label{sec: sup extended analysis of synthetic datasets}

{This section clarifies our experimental design choices regarding synthetic datasets in Section \ref{sec: experiment} and provides additional analysis of system performance with extended training.

Unlike binary-valued problems where the input space contains $2^d$ discrete instances, our synthetic datasets use real-valued inputs from $[0,1]^d$, containing infinitely many possible instances. We sampled 6000 instances to match our largest real-world dataset (Paddy leaf) while maintaining reasonable computational costs per epoch.

To investigate achievable performance limits, we conduct experiments with extended training (300 epochs) on the 20-dimensional real-valued multiplexer problem using 6000 samples from $[0,1]^{20}$. 

Fig. \ref{fig: 20mux learning curve} shows the learning curves (training accuracy, test accuracy, and population size) for both UCS with \textit{Hyperrectangles} and \ours\ with FBRC. While UCS reaches 98.82\% training and 97.05\% test accuracy, with a population size of 1642 rules at the 300th epoch, \ours\ achieves 98.55\% training and 97.55\% test accuracy, with a population size of 1476 rules.

An essential observation from this experiment's outcomes is the effectiveness of fuzzy rules in FBRC for the real-valued multiplexer problem, which ideally requires only crisp rules. This is because, compared to the crisp form, the fuzzy form can reduce the negative impact of over-general rules on misclassification and evolutionary propagation of rules.  Specifically, in the rule generation phase (i.e., when applying the covering and genetic operators), LCSs frequently generate over-general rules that cover multi-class subspaces. General LCSs (e.g., XCS, UCS, Fuzzy-UCS) lack explicit methods to eliminate these over-general rules \cite{wagner2022mechanisms}. Such rules are capable of degrading the performance of a system since they may result in the inference of an incorrect class and propagate incorrect information \cite{wagner2022mechanisms}. As elaborated in Section \ref{sec: Fuzzy-UCS}, the higher the matching degree, the more influence a rule has on parent selection and class inference. Over-general crisp rules, thus, have a more pronounced negative impact on parent selection and class inference than their fuzzy counterparts. This distinction arises because fuzzy rules have matching degrees that vary between 0 and 1 for the entire domain defined by the rule-antecedent, while crisp rules consistently possess a matching degree of 1. Given that fuzzy rules have a matching degree near zero in proximity to the rule boundary, a slightly over-generalized fuzzy rule exerts a less negative influence on the system compared to crisp rules in aspects like parent selection and class inference. This suggests that minor over-general fuzzy rules can serve as less sensitive rules, aiding the system in efficiently optimizing each rule's membership function to attain the desired shape and interval range.

For instance, consider a 1D binary classification problem on the input space $[0,1]$ where $x_1$ belongs to class 1 in $[0,0.5)$ and class 2 in $[0.5,1.0]$, as shown in Fig. \ref{fig: overgeneral comparison}. An over-general hyperrectangular crisp rule with antecedent ``$x_1\in[0,0.6]$'' can be contrasted with an over-general FBR fuzzy rule with antecedent ``$x_1 \text{ is }(1.1,1.6,0.0,0.6)$.'' The latter exhibits a smaller rule matching degree near the class boundary of 0.5 compared to the former, making it a less sensitive rule.
}

\begin{figure*}[ht]
\centering
{\includegraphics[width=0.3\textwidth]{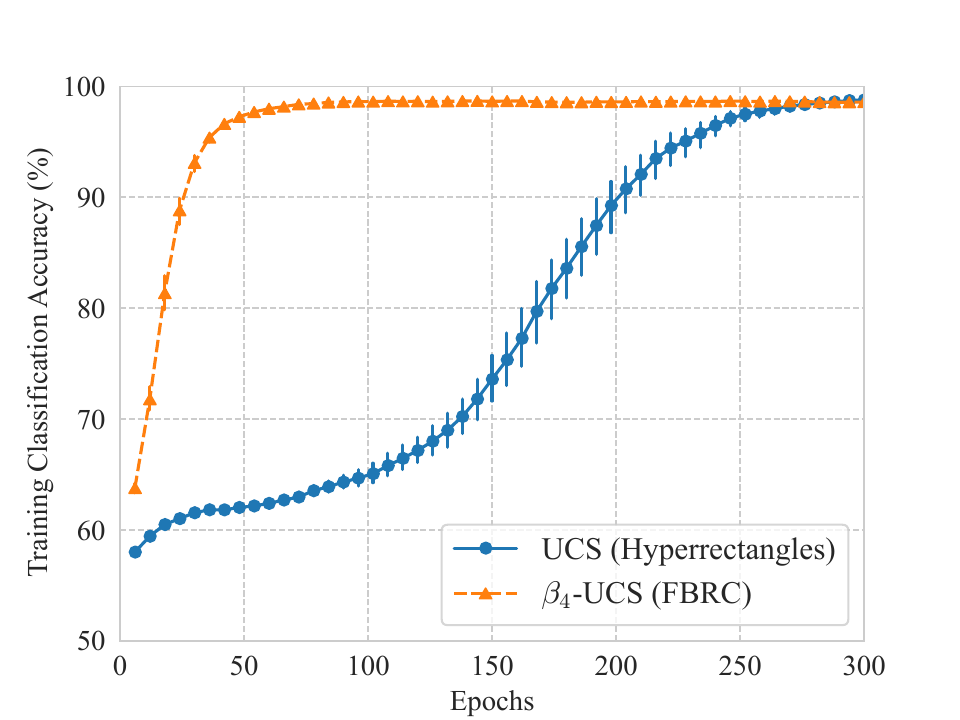}}
{\includegraphics[width=0.3\textwidth]{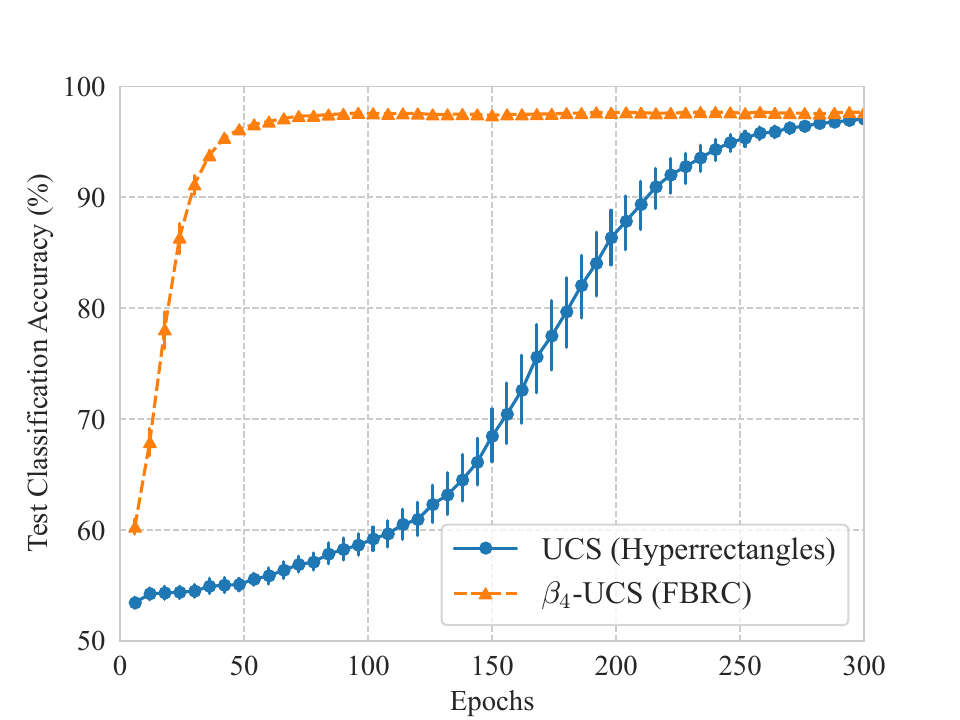}}
{\includegraphics[width=0.3\textwidth]{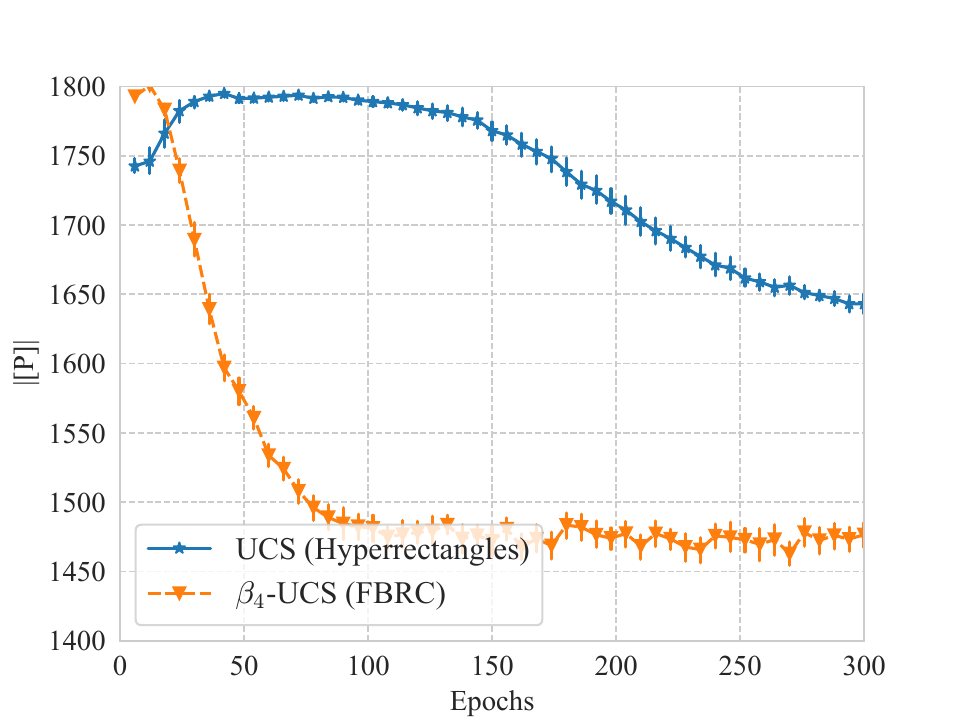}}
\caption{Average training accuracy (left), test accuracy (center), and population size (right) of UCS and \ours\ on the 20-dimensional real-valued multiplexer problem over 30 runs. Error bars represent the 95\% confidence interval.}
\label{fig: 20mux learning curve}
\end{figure*}

\begin{figure*}[ht]
\centerline{\includegraphics[width=\linewidth]{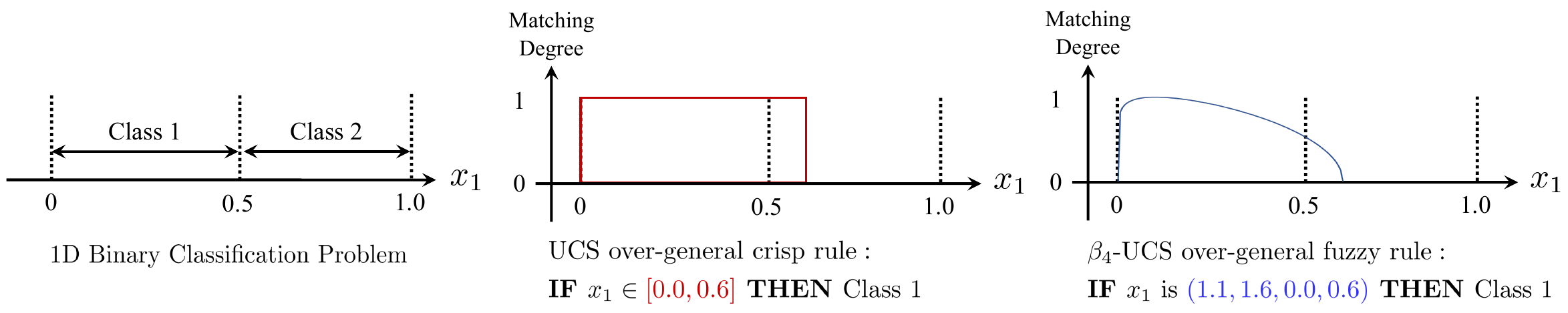}}
\caption{An example of over-general rules generated by UCS (center) and \ours\ (right) on the 1D binary classification problem (left).}
\label{fig: overgeneral comparison}
\end{figure*}

\clearpage

\section{Experimental Results on Macro F1 Score}
\label{sec: sup experimental results on macro f1 score}

We conduct additional experiments to evaluate the proposed \ours\ on a more robust metric, i.e., the macro F1 score.
The datasets and settings used in our experiments have adhered to those outlined in Section \ref{sec: experiment}.

Table \ref{tb: sup f1} presents each representation's average test macro F1 score and average rank across all datasets. The table reveals that FBR and FBRC record the second and first average ranks among all ten representations, respectively. Therefore, even when using macro F1 score instead of accuracy as an evaluation metric, \ours\ maintains its effectiveness.

\begin{table*}[ht]
\begin{center}
\caption{Summary of Results, Displaying Average Test Macro F1 Score Across 30 Runs. Statistical Results of the Wilcoxon Signed-Rank Test Are Summarized With Symbols Where ``$+$'' and ``$-$'' Indicate Significantly Better and Worse Performance Than FBRC, Respectively, While ``$\sim$'' Indicates No Significant Difference From FBRC. ``Rank''; ``Position''; ``$p$-Value''; ``$p_\text{Holm}$-Value''; and Green- and Peach-Shaded Values Should be Interpreted as in Table \ref{tb: sup sota}.}
\label{tb: sup f1}
\normalsize
\resizebox{\textwidth}{!}{

\begin{tabular}{c|cccc|cccc|cc}
\bhline{1pt}
&\multicolumn{4}{c|}{UCS} & \multicolumn{4}{c|}{Fuzzy-UCS} & \multicolumn{2}{c}{\ours}
\\
\cline{2-11}
ID.  & \textit{Hyperrectangles} & \textit{Hyperellipsoids} & \textit{CurvedPolytopes} & \textit{Self-Ada-RP}  & \textit{Triangles}&\textit{Trapezoids}&\textit{SymmetricBells}&\textit{Self-Ada-RT}& \ourrep&\ourrep C
\\
\bhline{1pt}
{\texttt{cae}} & 61.47 $\sim$ & 60.18 $\sim$ & 59.57 $\sim$ & 59.57 $\sim$ & \cellcolor{g}62.93 $\sim$ & 57.74 $\sim$ & \cellcolor{p}57.51 $\sim$ & 61.04 $\sim$ & 59.34 $\sim$ & 58.34 \\
\texttt{can} & 91.77 $-$ & 87.87 $-$ & \cellcolor{p}45.01 $-$ & 76.15 $-$ & 83.97 $-$ & 92.83 $\sim$ & 90.10 $-$ & 92.69 $\sim$ & 94.41 $\sim$ & \cellcolor{g}94.42 \\
\texttt{car} & 91.96 $-$ & 84.34 $-$ & 86.96 $-$ & 92.71 $-$ & 88.67 $-$ & 88.78 $-$ & \cellcolor{p}82.79 $-$ & 90.40 $-$ & 94.54 $\sim$ & \cellcolor{g}94.77 \\
\texttt{chk} & 53.86 $-$ & 53.40 $-$ & 53.33 $-$ & 53.76 $-$ & 58.53 $-$ & \cellcolor{p}50.03 $-$ & 54.33 $-$ & 81.39 $\sim$ & 81.16 $\sim$ & \cellcolor{g}81.62 \\
\texttt{cmx} & 71.29 $-$ & 60.93 $-$ & 50.85 $-$ & 79.98 $-$ & \cellcolor{p}6.011 $-$ & 86.77 $-$ & 10.10 $-$ & 87.42 $-$ & 92.10 $-$ & \cellcolor{g}93.30 \\
\texttt{col} & 61.02 $-$ & 68.70 $\sim$ & 69.55 $\sim$ & 71.37 $\sim$ & 70.74 $\sim$ & 62.37 $\sim$ & \cellcolor{p}59.92 $-$ & \cellcolor{g}71.67 $\sim$ & 66.73 $\sim$ & 67.75 \\
\texttt{dbt} & 68.27 $\sim$ & \cellcolor{g}70.65 $\sim$ & 70.00 $\sim$ & 69.87 $\sim$ & 68.09 $\sim$ & 68.19 $\sim$ & \cellcolor{p}57.48 $-$ & 69.37 $\sim$ & 69.52 $\sim$ & 69.99 \\
\texttt{ecl} & \cellcolor{p}57.77 $-$ & \cellcolor{g}73.48 $\sim$ & 66.17 $\sim$ & 68.95 $\sim$ & 70.94 $\sim$ & 64.81 $\sim$ & 66.19 $\sim$ & 70.45 $\sim$ & 66.12 $\sim$ & 69.58 \\
\texttt{frt} & 75.42 $-$ & 74.01 $-$ & \cellcolor{p}27.72 $-$ & 58.95 $-$ & 58.48 $-$ & \cellcolor{g}80.25 $\sim$ & 63.05 $-$ & 76.06 $-$ & 78.00 $-$ & 79.83 \\
\texttt{gls} & \cellcolor{p}44.91 $-$ & 54.98 $\sim$ & 54.53 $\sim$ & 52.42 $\sim$ & 55.05 $\sim$ & 47.85 $-$ & 52.59 $\sim$ & \cellcolor{g}55.44 $\sim$ & 53.03 $\sim$ & 54.35 \\
\texttt{hcl} & 61.21 $\sim$ & \cellcolor{p}37.19 $-$ & 62.10 $\sim$ & \cellcolor{g}64.97 $\sim$ & 56.62 $\sim$ & 57.59 $\sim$ & 50.02 $-$ & 57.30 $\sim$ & 60.98 $\sim$ & 60.50 \\
\texttt{mam} & \cellcolor{g}82.26 $\sim$ & 79.34 $\sim$ & 81.02 $\sim$ & 81.73 $\sim$ & 80.28 $\sim$ & \cellcolor{p}78.77 $-$ & 79.80 $\sim$ & 80.64 $\sim$ & 81.11 $\sim$ & 81.47 \\
\texttt{mop} & 87.47 $-$ & 84.96 $-$ & 85.01 $-$ & \cellcolor{g}92.55 $+$ & 87.55 $-$ & \cellcolor{p}83.23 $-$ & 83.87 $-$ & 87.45 $-$ & 91.88 $\sim$ & 91.71 \\
{\texttt{mul}} & 74.38 $-$ & \cellcolor{p}58.20 $-$ & 82.08 $\sim$ & \cellcolor{g}87.34 $\sim$ & 65.54 $-$ & 80.15 $-$ & 64.60 $-$ & 73.66 $-$ & 84.51 $\sim$ & 85.82 \\
\texttt{mux} & 55.58 $-$ & 33.45 $-$ & 57.07 $-$ & 66.53 $-$ & \cellcolor{p}33.44 $-$ & 64.49 $-$ & 56.79 $-$ & 86.00 $-$ & 94.96 $-$ & \cellcolor{g}96.33 \\
{\texttt{nph}} & 53.42 $\sim$ & 53.46 $\sim$ & 55.03 $+$ & \cellcolor{g}55.41 $+$ & 52.72 $\sim$ & 53.38 $\sim$ & 52.39 $\sim$ & 51.74 $\sim$ & 52.56 $\sim$ & \cellcolor{p}51.67 \\
\texttt{pdy} & 88.06 $-$ & 86.74 $-$ & 87.21 $-$ & 87.10 $-$ & 87.04 $-$ & \cellcolor{p}77.89 $-$ & 85.82 $-$ & 88.76 $\sim$ & \cellcolor{g}88.92 $\sim$ & 88.89 \\
\texttt{pis} & 86.04 $\sim$ & 86.12 $\sim$ & 85.78 $\sim$ & \cellcolor{p}84.88 $-$ & 86.14 $\sim$ & 85.45 $\sim$ & 84.92 $-$ & 85.88 $\sim$ & 86.23 $\sim$ & \cellcolor{g}86.23 \\
\texttt{pmp} & 87.34 $\sim$ & 87.82 $+$ & 87.35 $\sim$ & 87.37 $+$ & \cellcolor{g}87.94 $+$ & 86.91 $\sim$ & 87.38 $\sim$ & 87.45 $\sim$ & 86.84 $\sim$ & \cellcolor{p}86.49 \\
\texttt{rsn} & 83.03 $-$ & 85.10 $\sim$ & 84.67 $\sim$ & 84.89 $\sim$ & 84.47 $\sim$ & \cellcolor{p}82.32 $\sim$ & 83.35 $\sim$ & 84.76 $\sim$ & \cellcolor{g}85.41 $\sim$ & 85.34 \\
\texttt{soy} & 63.41 $\sim$ & \cellcolor{p}41.66 $-$ & 53.46 $-$ & 55.53 $-$ & 46.04 $-$ & \cellcolor{g}75.29 $+$ & 69.74 $+$ & 55.82 $-$ & 62.27 $\sim$ & 63.26 \\
\texttt{tae} & 43.96 $\sim$ & 45.91 $\sim$ & 47.08 $\sim$ & 48.54 $\sim$ & \cellcolor{g}51.64 $\sim$ & \cellcolor{p}38.98 $-$ & 49.85 $\sim$ & 48.10 $\sim$ & 49.58 $\sim$ & 49.01 \\
\texttt{wne} & 84.60 $-$ & \cellcolor{g}94.97 $\sim$ & \cellcolor{p}66.35 $-$ & 86.31 $-$ & 93.17 $\sim$ & 89.85 $\sim$ & 89.15 $\sim$ & 91.47 $\sim$ & 91.14 $\sim$ & 90.94 \\
\texttt{wpb} & \cellcolor{g}57.49 $+$ & 43.58 $\sim$ & \cellcolor{p}43.16 $-$ & 45.28 $\sim$ & 53.13 $\sim$ & 52.98 $+$ & 53.92 $+$ & 47.98 $\sim$ & 48.58 $\sim$ & 48.15 \\
\texttt{yst} & 45.62 $\sim$ & 48.79 $\sim$ & 46.50 $\sim$ & 46.84 $\sim$ & 47.82 $\sim$ & \cellcolor{p}36.71 $-$ & 46.63 $\sim$ & 45.94 $\sim$ & 48.44 $\sim$ & \cellcolor{g}48.87 \\
\bhline{1pt}
Rank & \textit{5.88} & \textit{5.88} & \textit{6.40} & \textit{4.96} & \textit{5.40} & \textit{7.00} & \cellcolor{p}\textit{7.16} & \textit{4.68} & \textit{4.00} & \cellcolor{g}\textit{3.64} \\
Position & \textit{6} & \textit{7} & \textit{8} & \textit{4} & \textit{5} & \textit{9} & \textit{10} & \textit{3} & \textit{2} & \textit{1} \\
$+/-/\sim$ & 1/14/10 & 1/11/13 & 1/11/13 & 3/10/12 & 1/10/14 & 2/11/12 & 2/13/10 & 0/7/18 & 0/3/22 & - \\
\bhline{1pt}
$p$-value & 0.00226 & 0.00807 & 0.00115 & 0.0342 & 0.0551 & 0.000912 & 0.000329 & 0.0187 & 0.0903 & - \\
$p_\text{Holm}$-value & 0.0135 & 0.0403 & 0.00808 & 0.103 & 0.110 & 0.00730 & 0.00296 & 0.0750 & 0.110 & - \\

\bhline{1pt}
\end{tabular}}
\end{center}
\end{table*}

\clearpage

\section{Experimental Results on 8:2 Split Ratio}
\label{sec: sup experimental results on 8:2 split ratio}
{To verify the robustness of our results against different data splitting protocols, we conduct experiments using an 8:2 training-test split ratio. The datasets and settings used in our experiments have adhered to those outlined in Section \ref{sec: experiment}.

Table \ref{tb: sup shuffle split 8:2} presents each representation's average test classification accuracy and average rank across all datasets, using an 8:2 training-test split ratio. The table reveals that our proposed FBR and FBRC achieve the first (average rank {3.46}) and second (average rank {3.64}) positions, respectively, among all representations. Statistical analysis shows that while FBR slightly outranks FBRC overall, FBRC demonstrates superior performance on more individual datasets (significantly better on three datasets versus one for FBR). These results indicate the general effectiveness of \ours\ and the crispification operator, even with a different training-test data splitting protocol.}

	\begin{table*}[ht]
\begin{center}
\caption{Summary of Results Using Shuffle-Split Cross Validation With an 8:2 Training-Test Split Ratio, Displaying Average Test Classification Accuracy Across 30 Runs. ``$+$'', ``$-$'', and ``$\sim$''; ``Rank''; ``Position''; ``$p$-Value''; ``$p_\text{Holm}$-Value''; and Green- and Peach-Shaded Values Should be Interpreted as in Table \ref{tb: sup f1}.
}
\label{tb: sup shuffle split 8:2}
\normalsize
\resizebox{\textwidth}{!}{

\begin{tabular}{c|cccc|cccc|cc}
\bhline{1pt}
&\multicolumn{4}{c|}{UCS} & \multicolumn{4}{c|}{Fuzzy-UCS} & \multicolumn{2}{c}{\ours}
\\
\cline{2-11}
ID.  & \textit{Hyperrectangles} & \textit{Hyperellipsoids} & \textit{CurvedPolytopes} & \textit{Self-Ada-RP}  & \textit{Triangles}&\textit{Trapezoids}&\textit{SymmetricBells}&\textit{Self-Ada-RT}& \ourrep&\ourrep C
\\
\bhline{1pt}
{\texttt{cae}} & 63.54 $\sim$ & \cellcolor{p}63.33 $\sim$ & 64.38 $\sim$ & 63.75 $\sim$ & 63.75 $\sim$ & 64.38 $\sim$ & 64.58 $\sim$ & \cellcolor{g}67.92 $\sim$ & 65.62 $\sim$ & 65.42 \\
\texttt{can} & 91.17 $-$ & 87.22 $-$ & \cellcolor{p}63.27 $-$ & 78.74 $-$ & 88.68 $-$ & 93.71 $-$ & 92.19 $-$ & 92.54 $-$ & 94.62 $\sim$ & \cellcolor{g}95.15 \\
\texttt{car} & 91.79 $-$ & 85.49 $-$ & 87.88 $-$ & 92.65 $-$ & 89.32 $-$ & 89.13 $-$ & \cellcolor{p}84.31 $-$ & 91.29 $-$ & \cellcolor{g}94.44 $\sim$ & 94.41 \\
\texttt{chk} & 54.50 $-$ & 53.03 $-$ & 53.06 $-$ & 53.99 $-$ & 57.97 $-$ & \cellcolor{p}51.24 $-$ & 53.40 $-$ & \cellcolor{g}80.65 $\sim$ & 79.67 $\sim$ & 79.82 \\
\texttt{cmx} & 68.31 $-$ & 60.36 $-$ & 49.91 $-$ & 78.22 $-$ & \cellcolor{p}7.997 $-$ & 85.53 $-$ & 13.38 $-$ & 86.72 $-$ & 91.05 $-$ & \cellcolor{g}91.90 \\
\texttt{col} & 72.85 $-$ & 76.88 $\sim$ & 78.71 $\sim$ & \cellcolor{g}79.78 $\sim$ & 79.14 $\sim$ & \cellcolor{p}71.18 $-$ & 74.62 $-$ & 78.71 $\sim$ & 77.31 $-$ & 78.76 \\
\texttt{dbt} & 73.35 $-$ & 76.71 $+$ & 76.62 $+$ & \cellcolor{g}77.06 $\sim$ & 74.52 $-$ & 73.74 $-$ & \cellcolor{p}69.48 $-$ & 75.91 $\sim$ & 75.11 $\sim$ & 75.43 \\
\texttt{ecl} & \cellcolor{p}75.93 $-$ & \cellcolor{g}85.83 $+$ & 81.47 $\sim$ & 82.65 $\sim$ & 85.15 $+$ & 79.95 $-$ & 81.52 $\sim$ & 84.75 $+$ & 84.46 $+$ & 82.79 \\
\texttt{frt} & 81.57 $-$ & 78.24 $-$ & \cellcolor{p}24.56 $-$ & 63.69 $-$ & 73.54 $-$ & \cellcolor{g}84.85 $\sim$ & 77.17 $-$ & 81.70 $-$ & 84.15 $\sim$ & 84.81 \\
\texttt{gls} & \cellcolor{p}58.14 $-$ & 65.66 $-$ & 64.26 $-$ & 61.32 $-$ & 68.84 $\sim$ & 63.80 $-$ & 66.59 $-$ & 68.53 $\sim$ & 68.91 $\sim$ & \cellcolor{g}69.38 \\
\texttt{hcl} & 62.03 $-$ & \cellcolor{p}40.00 $-$ & 62.03 $-$ & 64.19 $-$ & 68.24 $\sim$ & \cellcolor{g}70.95 $\sim$ & 67.52 $\sim$ & 67.75 $\sim$ & 70.81 $\sim$ & 69.50 \\
\texttt{mam} & \cellcolor{g}81.61 $\sim$ & 80.57 $-$ & 80.64 $-$ & 81.57 $\sim$ & 81.42 $\sim$ & \cellcolor{p}79.71 $-$ & 80.28 $-$ & 81.33 $\sim$ & 81.55 $\sim$ & 81.54 \\
\texttt{mop} & 86.85 $-$ & 84.88 $-$ & 84.45 $-$ & \cellcolor{g}92.33 $+$ & 87.41 $-$ & \cellcolor{p}83.00 $-$ & 83.22 $-$ & 86.70 $-$ & 91.26 $\sim$ & 90.97 \\
{\texttt{mul}} & 73.58 $-$ & \cellcolor{p}60.55 $-$ & \cellcolor{g}84.67 $\sim$ & 83.45 $\sim$ & 67.70 $-$ & 78.36 $\sim$ & 68.91 $-$ & 72.55 $-$ & 83.64 $\sim$ & 81.70 \\
\texttt{mux} & 55.15 $-$ & \cellcolor{p}49.80 $-$ & 56.72 $-$ & 64.96 $-$ & 50.12 $-$ & 63.72 $-$ & 56.83 $-$ & 81.78 $-$ & 94.19 $-$ & \cellcolor{g}96.08 \\
{\texttt{nph}} & 54.83 $\sim$ & 55.94 $\sim$ & 56.39 $\sim$ & 55.62 $\sim$ & 55.24 $\sim$ & 56.25 $+$ & \cellcolor{g}56.78 $+$ & 55.52 $\sim$ & 54.62 $\sim$ & \cellcolor{p}54.59 \\
\texttt{pdy} & 88.34 $-$ & 85.42 $-$ & 87.53 $-$ & 86.99 $-$ & 87.17 $-$ & \cellcolor{p}69.15 $-$ & 85.76 $-$ & 88.86 $\sim$ & \cellcolor{g}88.90 $\sim$ & 88.85 \\
\texttt{pis} & 85.76 $\sim$ & 86.44 $\sim$ & 85.51 $-$ & 85.71 $\sim$ & \cellcolor{g}86.63 $\sim$ & 85.55 $-$ & \cellcolor{p}85.26 $-$ & 86.33 $\sim$ & 86.47 $\sim$ & 86.29 \\
\texttt{pmp} & \cellcolor{p}86.34 $-$ & \cellcolor{g}88.05 $+$ & 87.81 $\sim$ & 86.78 $\sim$ & 87.57 $\sim$ & 86.61 $-$ & 87.16 $\sim$ & 87.40 $\sim$ & 87.24 $\sim$ & 87.48 \\
\texttt{rsn} & \cellcolor{p}83.85 $\sim$ & \cellcolor{g}85.96 $\sim$ & 85.83 $\sim$ & 85.59 $\sim$ & 85.89 $\sim$ & 84.30 $-$ & 85.15 $\sim$ & 85.39 $\sim$ & 85.22 $\sim$ & 85.41 \\
\texttt{soy} & 65.45 $\sim$ & \cellcolor{p}44.84 $-$ & 56.91 $-$ & 58.98 $-$ & 61.00 $-$ & \cellcolor{g}76.91 $+$ & 75.89 $+$ & 61.12 $-$ & 66.86 $\sim$ & 67.08 \\
\texttt{tae} & 46.99 $\sim$ & 50.65 $\sim$ & 49.78 $\sim$ & 49.57 $\sim$ & 51.29 $\sim$ & \cellcolor{p}45.91 $\sim$ & \cellcolor{g}55.48 $+$ & 52.80 $+$ & 49.68 $\sim$ & 49.35 \\
\texttt{wne} & 84.72 $-$ & 95.19 $\sim$ & \cellcolor{p}64.44 $-$ & 79.72 $-$ & \cellcolor{g}95.37 $\sim$ & 92.04 $\sim$ & 89.44 $-$ & 92.41 $\sim$ & 93.43 $\sim$ & 94.17 \\
\texttt{wpb} & 69.92 $-$ & 74.92 $\sim$ & 75.17 $\sim$ & 76.83 $\sim$ & \cellcolor{p}61.33 $-$ & \cellcolor{g}77.33 $+$ & 64.67 $-$ & 71.75 $-$ & 75.17 $\sim$ & 74.83 \\
\texttt{yst} & 55.58 $-$ & \cellcolor{g}58.96 $\sim$ & 56.05 $-$ & 57.91 $\sim$ & 58.02 $\sim$ & \cellcolor{p}47.45 $-$ & 56.29 $-$ & 57.27 $\sim$ & 58.18 $\sim$ & 57.95 \\
\bhline{1pt}
Rank & \cellcolor{p}\textit{7.12} & \textit{6.00} & \textit{6.44} & \textit{5.42} & \textit{5.22} & \textit{6.54} & \textit{6.80} & \textit{4.36} & \cellcolor{g}\textit{3.46} & \textit{3.64} \\
Position & \textit{10} & \textit{6} & \textit{7} & \textit{5} & \textit{4} & \textit{8} & \textit{9} & \textit{3} & \textit{1} & \textit{2} \\
$+/-/\sim$ & 0/18/7 & 3/13/9 & 1/15/9 & 1/11/13 & 1/12/12 & 3/16/6 & 3/17/5 & 2/9/14 & 1/3/21 & - \\
\bhline{1pt}
$p$-value & 2.98E-07 & 0.00613 & 0.00145 & 0.0236 & 0.00737 & 0.00145 & 0.000808 & 0.0255 & 0.615 & - \\
$p_\text{Holm}$-value & 2.68E-06 & 0.0306 & 0.0102 & 0.0709 & 0.0306 & 0.0102 & 0.00646 & 0.0709 & 0.615 & - \\

\bhline{1pt}
\end{tabular}}
\end{center}
\end{table*}

\clearpage
	
\section{Experimental Results With 10-Fold Cross-Validation}
\label{sec: sup experimental results with 10-fold cross validation}
{While our experiments in Section \ref{sec: experiment} use shuffle-split validation following recent LCS studies (e.g., \cite{preen2021autoencoding}, \cite{heider2023suprb}, and \cite{shiraishi2024variable}), we conduct experiments using 10-fold cross-validation to verify the robustness of our results. The datasets and settings used in our experiments have adhered to those outlined in Section \ref{sec: experiment}.

Table \ref{tb: sup 10-fold cv} presents each representation's average test classification accuracy and average rank across all datasets, using 10-fold cross-validation. The table reveals that our proposed FBR and FBRC achieve the first (average rank {3.16}) and second (average rank {3.84}) positions, respectively, among all representations. Statistical analysis shows that while FBR slightly outranks FBRC overall, FBRC demonstrates superior performance on more individual datasets (significantly better on one dataset versus zero for FBR). These results indicate the general effectiveness of \ours\ and the crispification operator, even with a different cross-validation protocol.}

\begin{table*}[ht]
\begin{center}
\caption{Summary of Results Using 10-Fold Cross Validation, Displaying Average Test Classification Accuracy. ``$+$'', ``$-$'', and ``$\sim$''; ``Rank''; ``Position''; ``$p$-Value''; ``$p_\text{Holm}$-Value''; and Green- and Peach-Shaded Values Should be Interpreted as in Table \ref{tb: sup f1}.}
\label{tb: sup 10-fold cv}
\normalsize
\resizebox{\textwidth}{!}{

\begin{tabular}{c|cccc|cccc|cc}
\bhline{1pt}
&\multicolumn{4}{c|}{UCS} & \multicolumn{4}{c|}{Fuzzy-UCS} & \multicolumn{2}{c}{\ours}
\\
\cline{2-11}
ID.  & \textit{Hyperrectangles} & \textit{Hyperellipsoids} & \textit{CurvedPolytopes} & \textit{Self-Ada-RP}  & \textit{Triangles}&\textit{Trapezoids}&\textit{SymmetricBells}&\textit{Self-Ada-RT}& \ourrep&\ourrep C
\\
\bhline{1pt}
{\texttt{cae}} & 62.50 $\sim$ & 61.25 $\sim$ & 60.00 $\sim$ & 58.75 $\sim$ & 61.25 $\sim$ & \cellcolor{g}67.50 $\sim$ & 61.25 $\sim$ & 60.00 $\sim$ & \cellcolor{p}55.00 $\sim$ & 57.50 \\
\texttt{can} & 90.89 $-$ & 89.11 $-$ & \cellcolor{p}64.64 $-$ & 80.71 $-$ & 88.75 $-$ & 92.50 $\sim$ & 90.18 $\sim$ & 92.50 $\sim$ & \cellcolor{g}95.54 $\sim$ & 95.00 \\
\texttt{car} & 92.72 $-$ & 84.78 $-$ & 87.22 $-$ & 93.43 $\sim$ & 89.38 $-$ & 88.68 $-$ & \cellcolor{p}83.20 $-$ & 90.78 $-$ & \cellcolor{g}94.43 $\sim$ & 94.08 \\
\texttt{chk} & 55.45 $-$ & 53.90 $-$ & 54.17 $-$ & 55.13 $-$ & 58.88 $-$ & \cellcolor{p}50.65 $-$ & 54.23 $-$ & 81.25 $\sim$ & 80.12 $\sim$ & \cellcolor{g}81.72 \\
\texttt{cmx} & 72.02 $-$ & 60.28 $-$ & 51.65 $-$ & 79.60 $-$ & \cellcolor{p}7.933 $-$ & 88.02 $-$ & 13.35 $-$ & 87.37 $-$ & 92.00 $\sim$ & \cellcolor{g}92.50 \\
\texttt{col} & \cellcolor{p}73.55 $\sim$ & 78.39 $\sim$ & 80.65 $\sim$ & 76.77 $\sim$ & 78.71 $\sim$ & 76.77 $\sim$ & 74.19 $\sim$ & \cellcolor{g}80.97 $\sim$ & 79.68 $\sim$ & 78.71 \\
\texttt{dbt} & 73.42 $\sim$ & 77.76 $\sim$ & 74.47 $\sim$ & 77.50 $\sim$ & 75.92 $\sim$ & 73.42 $\sim$ & \cellcolor{p}70.13 $\sim$ & 74.74 $\sim$ & \cellcolor{g}77.89 $\sim$ & 74.87 \\
\texttt{ecl} & \cellcolor{p}74.55 $-$ & \cellcolor{g}84.24 $\sim$ & 79.70 $\sim$ & 81.52 $\sim$ & 80.61 $\sim$ & 80.30 $\sim$ & 80.30 $\sim$ & 82.42 $\sim$ & \cellcolor{g}84.24 $\sim$ & 82.42 \\
\texttt{frt} & 81.80 $\sim$ & 81.35 $\sim$ & \cellcolor{p}29.55 $-$ & 65.84 $-$ & 74.49 $-$ & 84.49 $\sim$ & 73.48 $-$ & 80.79 $-$ & 84.83 $\sim$ & \cellcolor{g}85.51 \\
\texttt{gls} & \cellcolor{p}57.62 $\sim$ & 63.33 $\sim$ & 63.33 $\sim$ & 68.57 $\sim$ & 68.10 $\sim$ & 58.57 $\sim$ & 64.76 $\sim$ & 68.10 $\sim$ & \cellcolor{g}70.48 $\sim$ & 66.67 \\
\texttt{hcl} & 63.61 $\sim$ & \cellcolor{p}41.11 $-$ & 69.44 $\sim$ & 66.39 $\sim$ & 68.61 $\sim$ & 67.78 $-$ & 65.83 $\sim$ & 70.56 $\sim$ & 70.28 $\sim$ & \cellcolor{g}71.67 \\
\texttt{mam} & 81.15 $\sim$ & 80.94 $\sim$ & 81.77 $\sim$ & \cellcolor{g}82.71 $\sim$ & \cellcolor{p}79.90 $\sim$ & 80.10 $-$ & 80.52 $\sim$ & 80.94 $\sim$ & 81.15 $\sim$ & 82.29 \\
\texttt{mop} & 86.47 $-$ & 84.45 $-$ & 83.92 $-$ & \cellcolor{g}91.88 $+$ & 87.20 $-$ & \cellcolor{p}82.95 $-$ & 84.25 $-$ & 87.07 $-$ & 91.15 $\sim$ & 90.95 \\
{\texttt{mul}} & 74.44 $-$ & \cellcolor{p}59.63 $-$ & \cellcolor{g}88.15 $\sim$ & 85.93 $\sim$ & 68.89 $-$ & 83.33 $\sim$ & 63.33 $-$ & 76.67 $-$ & 87.78 $\sim$ & 86.30 \\
\texttt{mux} & 55.85 $-$ & \cellcolor{p}49.82 $-$ & 58.35 $-$ & 66.47 $-$ & 50.27 $-$ & 63.83 $-$ & 57.10 $-$ & 85.13 $-$ & 94.68 $-$ & \cellcolor{g}96.88 \\
{\texttt{nph}} & 56.76 $\sim$ & 54.93 $\sim$ & \cellcolor{g}57.04 $\sim$ & 55.35 $\sim$ & 54.65 $\sim$ & 56.06 $\sim$ & 56.20 $\sim$ & 53.94 $\sim$ & \cellcolor{p}51.97 $\sim$ & 53.94 \\
\texttt{pdy} & 88.03 $\sim$ & 87.48 $\sim$ & 87.58 $\sim$ & 87.58 $\sim$ & 86.95 $\sim$ & \cellcolor{p}74.30 $-$ & 85.77 $-$ & 88.68 $\sim$ & \cellcolor{g}88.78 $\sim$ & 88.33 \\
\texttt{pis} & 86.21 $\sim$ & 87.06 $\sim$ & 86.36 $\sim$ & 86.21 $\sim$ & 86.78 $\sim$ & 86.78 $\sim$ & \cellcolor{p}85.61 $\sim$ & 86.59 $\sim$ & \cellcolor{g}87.29 $\sim$ & 86.64 \\
\texttt{pmp} & 86.12 $\sim$ & \cellcolor{g}87.92 $\sim$ & 87.28 $\sim$ & 87.48 $\sim$ & 87.48 $\sim$ & \cellcolor{p}84.84 $-$ & 87.12 $\sim$ & 87.56 $\sim$ & 87.36 $\sim$ & 87.16 \\
\texttt{rsn} & \cellcolor{p}84.00 $\sim$ & 85.89 $\sim$ & \cellcolor{g}86.78 $\sim$ & 85.44 $\sim$ & 85.44 $\sim$ & 84.33 $\sim$ & 84.67 $\sim$ & 85.78 $\sim$ & 85.67 $\sim$ & 84.78 \\
\texttt{soy} & 67.35 $\sim$ & \cellcolor{p}49.85 $-$ & 61.18 $-$ & 58.68 $-$ & 62.06 $\sim$ & 76.32 $+$ & \cellcolor{g}77.79 $+$ & 63.24 $\sim$ & 70.00 $\sim$ & 69.71 \\
\texttt{tae} & 47.33 $\sim$ & 52.00 $\sim$ & 53.33 $\sim$ & 55.33 $\sim$ & 53.33 $\sim$ & \cellcolor{p}44.67 $\sim$ & \cellcolor{g}58.67 $\sim$ & 58.00 $\sim$ & 54.00 $\sim$ & 50.00 \\
\texttt{wne} & 87.06 $\sim$ & \cellcolor{g}95.88 $\sim$ & \cellcolor{p}71.76 $-$ & 85.29 $\sim$ & 95.29 $\sim$ & 94.12 $\sim$ & 88.82 $\sim$ & 90.59 $\sim$ & 93.53 $\sim$ & 92.94 \\
\texttt{wpb} & 72.11 $\sim$ & 77.37 $\sim$ & 76.84 $\sim$ & 75.79 $\sim$ & \cellcolor{p}58.95 $-$ & 75.79 $\sim$ & 70.53 $\sim$ & \cellcolor{g}77.37 $\sim$ & 75.26 $\sim$ & \cellcolor{g}77.37 \\
\texttt{yst} & 55.00 $\sim$ & 58.18 $\sim$ & 57.57 $\sim$ & 58.72 $\sim$ & 58.11 $\sim$ & \cellcolor{p}49.53 $-$ & 57.09 $\sim$ & 57.64 $\sim$ & \cellcolor{g}59.46 $\sim$ & 58.51 \\
\bhline{1pt}
Rank & \textit{6.88} & \textit{5.82} & \textit{6.16} & \textit{5.22} & \textit{6.00} & \textit{6.46} & \cellcolor{p}\textit{7.20} & \textit{4.26} & \cellcolor{g}\textit{3.16} & \textit{3.84} \\
Position & \textit{9} & \textit{5} & \textit{7} & \textit{4} & \textit{6} & \textit{8} & \textit{10} & \textit{3} & \textit{1} & \textit{2} \\
$+/-/\sim$ & 0/8/17 & 0/9/16 & 0/9/16 & 1/6/18 & 0/9/16 & 1/10/14 & 1/8/16 & 0/6/19 & 0/1/24 & - \\
\bhline{1pt}
$p$-value & 5.39E-05 & 0.0367 & 0.0128 & 0.0710 & 0.0126 & 0.00226 & 0.00278 & 0.0587 & 0.634 & - \\
$p_\text{Holm}$-value & 0.000485 & 0.147 & 0.0756 & 0.176 & 0.0756 & 0.0180 & 0.0195 & 0.176 & 0.634 & - \\

\bhline{1pt}
\end{tabular}}
\end{center}
\end{table*}

\clearpage
\section{Experimental Results With Stratified Shuffle-Split Cross-Validation}
\label{sec: sup experimental results with stratified shuffle-split cross validation}

{In Section \ref{sec: experiment}, we intentionally chose not to use stratification to evaluate system performance under real-world
conditions where class distributions may differ between training and test sets \cite{guan2022prediction}. Employing stratification could potentially lead to an artificial alteration of these class distributions, which might not accurately reflect real-world scenarios. However, to evaluate system performance under balanced class distributions, we conduct additional experiments using stratified shuffle-split cross-validation, maintaining the same class proportions in both training (90\%) and test (10\%) sets. The datasets and settings used in our experiments have adhered to those outlined in Section \ref{sec: experiment}.

Table \ref{tb: sup result 9:1 stratified shuffle-split cv} presents the average test classification accuracy and average rank across all datasets used in Section \ref{sec: experiment}. The table reveals that FBR and FBRC record the {first} and {second} average ranks among all ten representations, respectively. We can see that \ours\ maintains its effectiveness even with balanced class distributions in both training and test sets. }

\begin{table*}[ht]
\begin{center}
\caption{Summary of Results Using Stratified Shuffle-Split Cross Validation, Displaying Average Test Classification Accuracy Across 30 Runs. ``$+$'', ``$-$'', and ``$\sim$''; ``Rank''; ``Position''; ``$p$-Value''; ``$p_\text{Holm}$-Value''; and Green- and Peach-Shaded Values Should be Interpreted as in Table \ref{tb: sup f1}.}
\label{tb: sup result 9:1 stratified shuffle-split cv}
\normalsize
\resizebox{\textwidth}{!}{

\begin{tabular}{c|cccc|cccc|cc}
\bhline{1pt}
&\multicolumn{4}{c|}{UCS} & \multicolumn{4}{c|}{Fuzzy-UCS} & \multicolumn{2}{c}{\ours}
\\
\cline{2-11}
ID.  & \textit{Hyperrectangles} & \textit{Hyperellipsoids} & \textit{CurvedPolytopes} & \textit{Self-Ada-RP}  & \textit{Triangles}&\textit{Trapezoids}&\textit{SymmetricBells}&\textit{Self-Ada-RT}& \ourrep&\ourrep C
\\
\bhline{1pt}
{\texttt{cae}} & 65.00 $\sim$ & 66.67 $\sim$ & 65.00 $\sim$ & 64.58 $\sim$ & 67.92 $\sim$ & 65.42 $\sim$ & 64.58 $\sim$ & \cellcolor{g}69.17 $\sim$ & 64.58 $\sim$ & \cellcolor{p}63.75 \\
\texttt{can} & 91.75 $-$ & 89.24 $-$ & \cellcolor{p}65.03 $-$ & 81.46 $-$ & 88.65 $-$ & 94.27 $\sim$ & 91.81 $-$ & 93.45 $-$ & \cellcolor{g}95.32 $\sim$ & 95.15 \\
\texttt{car} & 92.34 $-$ & 85.27 $-$ & 87.78 $-$ & 93.17 $-$ & 89.30 $-$ & 89.03 $-$ & \cellcolor{p}84.19 $-$ & 91.17 $-$ & 93.98 $\sim$ & \cellcolor{g}94.19 \\
\texttt{chk} & 55.07 $-$ & 53.90 $-$ & 53.79 $-$ & 53.78 $-$ & 58.58 $-$ & \cellcolor{p}51.45 $-$ & 54.93 $-$ & 80.54 $-$ & 81.46 $\sim$ & \cellcolor{g}82.37 \\
\texttt{cmx} & 70.66 $-$ & 60.51 $-$ & 51.18 $-$ & 79.57 $-$ & \cellcolor{p}7.067 $-$ & 86.92 $-$ & 12.12 $-$ & 87.96 $-$ & 91.88 $\sim$ & \cellcolor{g}92.44 \\
\texttt{col} & \cellcolor{p}73.98 $-$ & 78.39 $\sim$ & 80.86 $\sim$ & 80.43 $\sim$ & 79.89 $\sim$ & 76.24 $-$ & 75.91 $-$ & \cellcolor{g}81.29 $+$ & 80.43 $+$ & 78.28 \\
\texttt{dbt} & 74.42 $\sim$ & 77.06 $\sim$ & 76.88 $\sim$ & 76.75 $\sim$ & 75.45 $\sim$ & 75.37 $\sim$ & \cellcolor{p}69.57 $-$ & \cellcolor{g}77.10 $\sim$ & 75.67 $\sim$ & 76.32 \\
\texttt{ecl} & \cellcolor{p}77.47 $-$ & \cellcolor{g}85.76 $\sim$ & 83.23 $\sim$ & 83.13 $\sim$ & 83.94 $\sim$ & 81.52 $-$ & 82.93 $\sim$ & 84.85 $\sim$ & 84.65 $\sim$ & 85.15 \\
\texttt{frt} & 83.30 $\sim$ & 82.04 $\sim$ & \cellcolor{p}28.22 $-$ & 65.26 $-$ & 75.41 $-$ & \cellcolor{g}85.96 $+$ & 75.85 $-$ & 82.63 $-$ & 85.22 $\sim$ & 84.37 \\
\texttt{gls} & \cellcolor{p}57.88 $-$ & 64.85 $\sim$ & 66.52 $\sim$ & 65.45 $\sim$ & 67.42 $\sim$ & 61.97 $-$ & 64.70 $-$ & 66.36 $\sim$ & 67.88 $\sim$ & \cellcolor{g}68.03 \\
\texttt{hcl} & 62.87 $-$ & \cellcolor{p}38.89 $-$ & 64.17 $-$ & 65.83 $-$ & 67.31 $-$ & \cellcolor{g}71.02 $\sim$ & 67.78 $-$ & 66.48 $-$ & 70.00 $\sim$ & 70.93 \\
\texttt{mam} & 82.03 $\sim$ & 81.00 $\sim$ & 81.13 $\sim$ & \cellcolor{g}83.23 $\sim$ & 81.89 $\sim$ & \cellcolor{p}80.07 $-$ & 81.37 $-$ & 82.85 $\sim$ & 82.96 $\sim$ & 82.82 \\
\texttt{mop} & 87.41 $-$ & 85.08 $-$ & 84.39 $-$ & \cellcolor{g}92.42 $+$ & 87.27 $-$ & \cellcolor{p}83.01 $-$ & 83.91 $-$ & 87.08 $-$ & 91.82 $+$ & 91.24 \\
{\texttt{mul}} & 76.19 $-$ & \cellcolor{p}60.95 $-$ & 85.83 $\sim$ & \cellcolor{g}86.67 $\sim$ & 67.86 $-$ & 79.17 $-$ & 68.10 $-$ & 73.81 $-$ & 83.57 $\sim$ & 83.93 \\
\texttt{mux} & 55.53 $-$ & \cellcolor{p}49.84 $-$ & 58.17 $-$ & 67.14 $-$ & 50.11 $-$ & 64.30 $-$ & 57.51 $-$ & 84.16 $-$ & 94.99 $-$ & \cellcolor{g}96.49 \\
{\texttt{nph}} & 55.21 $\sim$ & \cellcolor{p}54.04 $\sim$ & 56.57 $\sim$ & 56.29 $\sim$ & 56.29 $\sim$ & \cellcolor{g}57.37 $+$ & 56.06 $\sim$ & 55.02 $\sim$ & 54.88 $\sim$ & 54.98 \\
\texttt{pdy} & 88.10 $-$ & 86.44 $-$ & 87.47 $-$ & 87.09 $-$ & 87.18 $-$ & \cellcolor{p}76.72 $-$ & 85.68 $-$ & 88.41 $\sim$ & 88.74 $\sim$ & \cellcolor{g}88.79 \\
\texttt{pis} & 86.31 $\sim$ & 87.13 $\sim$ & 86.65 $\sim$ & \cellcolor{p}86.12 $\sim$ & \cellcolor{g}87.21 $\sim$ & 86.23 $\sim$ & 86.17 $\sim$ & 86.62 $\sim$ & 86.87 $\sim$ & 86.68 \\
\texttt{pmp} & 87.23 $\sim$ & \cellcolor{g}87.88 $\sim$ & 87.83 $\sim$ & 86.55 $\sim$ & 86.77 $\sim$ & 86.65 $\sim$ & \cellcolor{p}86.45 $-$ & 87.31 $\sim$ & 87.44 $\sim$ & 87.21 \\
\texttt{rsn} & 83.44 $\sim$ & \cellcolor{g}85.89 $\sim$ & 85.37 $\sim$ & 85.56 $\sim$ & 85.22 $\sim$ & \cellcolor{p}83.37 $-$ & 84.70 $\sim$ & 85.15 $\sim$ & 85.19 $\sim$ & 85.00 \\
\texttt{soy} & 66.97 $-$ & \cellcolor{p}49.35 $-$ & 62.14 $-$ & 62.89 $-$ & 62.19 $-$ & \cellcolor{g}79.20 $+$ & 78.76 $+$ & 64.23 $-$ & 70.90 $\sim$ & 70.45 \\
\texttt{tae} & 47.78 $\sim$ & 52.22 $+$ & 52.00 $+$ & \cellcolor{g}54.44 $+$ & 49.78 $\sim$ & \cellcolor{p}42.67 $\sim$ & 52.00 $+$ & 54.22 $+$ & 46.22 $\sim$ & 46.67 \\
\texttt{wne} & 84.44 $-$ & \cellcolor{g}95.37 $\sim$ & \cellcolor{p}71.30 $-$ & 85.00 $-$ & 94.44 $\sim$ & 92.22 $\sim$ & 90.19 $-$ & 93.52 $\sim$ & 92.22 $\sim$ & 93.52 \\
\texttt{wpb} & 69.50 $\sim$ & 74.67 $\sim$ & 75.00 $\sim$ & 74.50 $\sim$ & \cellcolor{p}59.67 $-$ & \cellcolor{g}77.17 $+$ & 67.50 $-$ & 73.83 $\sim$ & 74.00 $\sim$ & 73.33 \\
\texttt{yst} & 54.99 $-$ & \cellcolor{g}59.21 $\sim$ & 56.67 $\sim$ & 57.28 $\sim$ & 57.64 $\sim$ & \cellcolor{p}48.23 $-$ & 56.51 $\sim$ & 56.78 $\sim$ & 57.05 $\sim$ & 57.41 \\
\bhline{1pt}
Rank & \textit{6.82} & \textit{5.64} & \textit{5.84} & \textit{5.32} & \textit{5.68} & \textit{6.36} & \cellcolor{p}\textit{7.46} & \textit{4.24} & \cellcolor{g}\textit{3.76} & \textit{3.88} \\
Position & \textit{9} & \textit{5} & \textit{7} & \textit{4} & \textit{6} & \textit{8} & \textit{10} & \textit{3} & \textit{1} & \textit{2} \\
$+/-/\sim$ & 0/15/10 & 1/10/14 & 1/11/13 & 2/10/13 & 0/12/13 & 4/13/8 & 2/17/6 & 2/10/13 & 2/1/22 & - \\
\bhline{1pt}
$p$-value & 1.51E-05 & 0.0236 & 0.0173 & 0.0710 & 0.00737 & 0.00613 & 0.000808 & 0.0755 & 0.578 & - \\
$p_\text{Holm}$-value & 0.000136 & 0.0946 & 0.0866 & 0.213 & 0.0442 & 0.0429 & 0.00646 & 0.213 & 0.578 & - \\

\bhline{1pt}
\end{tabular}}
\end{center}
\end{table*}

\clearpage
\section{Experimental Results With Extended Training}
\label{sec: sup experimental results with extended learning}
{To provide a more comprehensive evaluation of achievable performance, we conduct additional experiments using 10-fold cross-validation with extended training (200 epochs). The datasets and settings used in our experiments have adhered to those outlined in Section \ref{sec: experiment}.

Table \ref{tb: sup 10-fold cv 200 epoch} presents each representation's average test classification accuracy and average rank across all datasets. The table reveals that our proposed FBR and FBRC achieve the {first} (average rank {3.20}) and {second} (average rank {3.44}) positions, respectively, among all representations. Statistical analysis shows that FBRC demonstrates superior performance on more individual datasets than FBR. These results indicate the general effectiveness of \ours\ and the crispification operator, even with a different cross-validation protocol and a different number of training epochs.}

 \begin{table*}[ht]
\begin{center}
\caption{Summary of Results Using 10-Fold Cross Validation, Displaying Average Test Classification Accuracy. The Number of Training Iterations is 200 Epochs. ``$+$'', ``$-$'', and ``$\sim$''; ``Rank''; ``Position''; ``$p$-Value''; ``$p_\text{Holm}$-Value''; and Green- and Peach-Shaded Values Should be Interpreted as in Table \ref{tb: sup f1}.
}
\label{tb: sup 10-fold cv 200 epoch}
\normalsize
\resizebox{\textwidth}{!}{

\begin{tabular}{c|cccc|cccc|cc}
\bhline{1pt}
&\multicolumn{4}{c|}{UCS} & \multicolumn{4}{c|}{Fuzzy-UCS} & \multicolumn{2}{c}{\ours}
\\
\cline{2-11}
ID.  & \textit{Hyperrectangles} & \textit{Hyperellipsoids} & \textit{CurvedPolytopes} & \textit{Self-Ada-RP}  & \textit{Triangles}&\textit{Trapezoids}&\textit{SymmetricBells}&\textit{Self-Ada-RT}& \ourrep&\ourrep C
\\
\bhline{1pt}
{\texttt{cae}} & 58.75 $\sim$ & 58.75 $\sim$ & \cellcolor{p}56.25 $\sim$ & 60.00 $\sim$ & 61.25 $\sim$ & \cellcolor{g}65.00 $\sim$ & 63.75 $\sim$ & 58.75 $\sim$ & 58.75 $\sim$ & \cellcolor{p}56.25 \\
\texttt{can} & 92.14 $\sim$ & 90.89 $-$ & \cellcolor{p}75.18 $-$ & 86.96 $-$ & 76.79 $-$ & 94.82 $\sim$ & 85.89 $-$ & 92.68 $\sim$ & \cellcolor{g}95.89 $\sim$ & \cellcolor{g}95.89 \\
\texttt{car} & 94.18 $\sim$ & 85.17 $-$ & 87.83 $-$ & 94.18 $\sim$ & 90.55 $-$ & 91.62 $-$ & \cellcolor{p}84.67 $-$ & 92.43 $-$ & \cellcolor{g}95.25 $\sim$ & 94.68 \\
\texttt{chk} & 65.10 $-$ & 56.35 $-$ & \cellcolor{p}55.13 $-$ & 58.07 $-$ & 70.72 $-$ & 61.45 $-$ & 59.02 $-$ & 85.72 $\sim$ & \cellcolor{g}86.07 $\sim$ & 85.82 \\
\texttt{cmx} & 88.02 $-$ & 60.78 $-$ & 54.85 $-$ & 90.05 $-$ & \cellcolor{p}12.17 $-$ & \cellcolor{g}95.13 $+$ & 14.82 $-$ & 91.52 $-$ & 93.35 $\sim$ & 93.73 \\
\texttt{col} & 80.97 $\sim$ & 82.26 $\sim$ & \cellcolor{g}85.16 $\sim$ & 83.55 $\sim$ & 82.58 $\sim$ & 82.58 $\sim$ & \cellcolor{p}78.06 $\sim$ & 84.52 $\sim$ & 84.19 $\sim$ & 82.90 \\
\texttt{dbt} & 74.87 $\sim$ & 76.18 $\sim$ & 76.58 $\sim$ & \cellcolor{g}76.71 $\sim$ & 75.79 $\sim$ & \cellcolor{p}72.24 $\sim$ & 74.74 $\sim$ & 75.00 $\sim$ & 76.58 $\sim$ & 75.53 \\
\texttt{ecl} & 81.82 $\sim$ & 85.76 $\sim$ & 84.24 $\sim$ & 84.55 $\sim$ & 81.52 $\sim$ & 85.15 $\sim$ & \cellcolor{p}80.30 $\sim$ & 83.03 $\sim$ & \cellcolor{g}86.06 $\sim$ & 84.85 \\
\texttt{frt} & 81.91 $\sim$ & 84.72 $\sim$ & \cellcolor{p}27.64 $-$ & 69.44 $-$ & 71.80 $-$ & 84.83 $\sim$ & 72.25 $-$ & 83.37 $\sim$ & 83.48 $\sim$ & \cellcolor{g}85.39 \\
\texttt{gls} & \cellcolor{p}62.38 $\sim$ & 70.00 $\sim$ & 67.14 $\sim$ & 67.62 $\sim$ & 69.52 $\sim$ & 67.62 $\sim$ & 65.71 $\sim$ & 70.00 $\sim$ & \cellcolor{g}73.81 $\sim$ & 72.38 \\
\texttt{hcl} & 59.17 $-$ & \cellcolor{p}39.72 $-$ & 71.94 $\sim$ & 71.11 $\sim$ & 63.33 $-$ & 70.28 $\sim$ & 67.22 $\sim$ & 72.50 $\sim$ & \cellcolor{g}72.78 $\sim$ & 70.00 \\
\texttt{mam} & 81.35 $\sim$ & 80.94 $\sim$ & 81.46 $\sim$ & 81.56 $\sim$ & 82.29 $\sim$ & 81.88 $\sim$ & \cellcolor{p}80.83 $\sim$ & 82.29 $\sim$ & \cellcolor{g}83.44 $\sim$ & 82.29 \\
\texttt{mop} & 89.03 $-$ & 85.35 $-$ & 84.02 $-$ & \cellcolor{g}93.75 $+$ & 88.27 $-$ & 84.90 $-$ & \cellcolor{p}83.27 $-$ & 87.80 $-$ & 92.25 $\sim$ & 91.85 \\
{\texttt{mul}} & 79.63 $-$ & \cellcolor{p}63.33 $-$ & 98.15 $+$ & \cellcolor{g}98.52 $+$ & 67.78 $-$ & 84.07 $-$ & 67.41 $-$ & 78.52 $-$ & 93.70 $\sim$ & 92.96 \\
\texttt{mux} & 84.78 $-$ & \cellcolor{p}49.82 $-$ & 62.97 $-$ & 87.33 $-$ & 51.83 $-$ & 94.48 $-$ & 58.00 $-$ & 95.90 $-$ & 96.77 $-$ & \cellcolor{g}97.65 \\
{\texttt{nph}} & 53.24 $\sim$ & 56.06 $\sim$ & \cellcolor{g}57.18 $\sim$ & 54.65 $\sim$ & 56.06 $\sim$ & 55.07 $\sim$ & \cellcolor{g}57.18 $\sim$ & 55.07 $\sim$ & \cellcolor{p}52.82 $\sim$ & 52.96 \\
\texttt{pdy} & 88.48 $\sim$ & 88.55 $\sim$ & 87.90 $\sim$ & 87.83 $\sim$ & 88.10 $\sim$ & 87.52 $-$ & \cellcolor{p}86.43 $-$ & 88.83 $\sim$ & \cellcolor{g}89.42 $\sim$ & 89.28 \\
\texttt{pis} & \cellcolor{p}85.65 $\sim$ & 86.59 $\sim$ & 86.82 $\sim$ & 86.21 $\sim$ & 87.15 $\sim$ & 86.17 $\sim$ & 86.07 $\sim$ & 86.68 $\sim$ & \cellcolor{g}87.20 $\sim$ & 86.64 \\
\texttt{pmp} & \cellcolor{p}85.72 $\sim$ & 87.84 $\sim$ & 87.28 $\sim$ & 87.76 $\sim$ & \cellcolor{g}88.24 $\sim$ & 86.80 $-$ & 87.56 $\sim$ & 87.20 $\sim$ & 87.68 $\sim$ & 88.00 \\
\texttt{rsn} & \cellcolor{p}83.89 $\sim$ & 85.33 $\sim$ & 85.56 $\sim$ & 85.56 $\sim$ & 85.44 $\sim$ & \cellcolor{g}85.67 $\sim$ & 85.67 $\sim$ & 85.44 $\sim$ & 85.56 $\sim$ & \cellcolor{g}85.67 \\
\texttt{soy} & 65.88 $\sim$ & \cellcolor{p}46.62 $-$ & 67.94 $\sim$ & 66.62 $\sim$ & 61.76 $\sim$ & 78.24 $+$ & \cellcolor{g}78.82 $+$ & 64.85 $\sim$ & 69.56 $\sim$ & 69.56 \\
\texttt{tae} & 51.33 $\sim$ & 59.33 $\sim$ & 57.33 $\sim$ & 56.00 $\sim$ & 55.33 $\sim$ & \cellcolor{p}50.67 $\sim$ & 54.00 $\sim$ & 55.33 $\sim$ & \cellcolor{g}60.67 $\sim$ & 55.33 \\
\texttt{wne} & 92.94 $\sim$ & \cellcolor{g}96.47 $\sim$ & \cellcolor{p}90.00 $\sim$ & 95.88 $\sim$ & 95.29 $\sim$ & 95.88 $\sim$ & 91.18 $\sim$ & 93.53 $\sim$ & 95.29 $\sim$ & \cellcolor{g}96.47 \\
\texttt{wpb} & \cellcolor{p}66.32 $\sim$ & 76.32 $\sim$ & \cellcolor{g}76.84 $\sim$ & 73.68 $\sim$ & 67.89 $\sim$ & 74.74 $\sim$ & 74.21 $\sim$ & 75.79 $\sim$ & 72.63 $\sim$ & \cellcolor{g}76.84 \\
\texttt{yst} & \cellcolor{p}54.39 $-$ & 58.78 $\sim$ & 57.64 $\sim$ & 58.24 $\sim$ & 57.91 $\sim$ & 57.77 $\sim$ & 57.64 $\sim$ & 56.89 $\sim$ & 57.09 $\sim$ & \cellcolor{g}58.92 \\
\bhline{1pt}
Rank & \cellcolor{p}\textit{7.50} & \textit{5.92} & \textit{6.10} & \textit{4.94} & \textit{6.04} & \textit{5.24} & \textit{7.38} & \textit{5.24} & \cellcolor{g}\textit{3.20} & \textit{3.44} \\
Position & \textit{10} & \textit{6} & \textit{8} & \textit{3} & \textit{7} & \textit{4} & \textit{9} & \textit{5} & \textit{1} & \textit{2} \\
$+/-/\sim$ & 0/7/18 & 0/9/16 & 1/7/17 & 2/5/18 & 0/9/16 & 2/7/16 & 1/9/15 & 0/5/20 & 0/1/24 & - \\
\bhline{1pt}
$p$-value & 3.28E-06 & 0.00792 & 0.0327 & 0.141 & 0.000963 & 0.0229 & 0.000912 & 0.0123 & 0.241 & - \\
$p_\text{Holm}$-value & 2.95E-05 & 0.0475 & 0.0980 & 0.282 & 0.0073 & 0.0917 & 0.00730 & 0.0615 & 0.282 & - \\

\bhline{1pt}
\end{tabular}}
\end{center}
\end{table*}

\clearpage
\section{Experimental Results on Training Accuracy}
\label{sec: training accuracy}
{Table \ref{tb: training accuracy} presents each representation's average training classification accuracy and average rank across all datasets.}

\begin{table*}[ht]
\begin{center}
\caption{Summary of Results, Displaying Average Training Classification Accuracy Across 30 Runs. ``$+$'', ``$-$'', and ``$\sim$''; ``Rank''; ``Position''; ``$p$-Value''; ``$p_\text{Holm}$-Value''; and Green- and Peach-Shaded Values Should be Interpreted as in Table \ref{tb: sup f1}.}
\label{tb: training accuracy}
\normalsize
\resizebox{\textwidth}{!}
{
\begin{tabular}{c|cccc|cccc|cc}
\bhline{1pt}
&\multicolumn{4}{c|}{UCS} & \multicolumn{4}{c|}{Fuzzy-UCS} & \multicolumn{2}{c}{\ours}
\\
\cline{2-11}
ID.  & \textit{Hyperrectangles} & \textit{Hyperellipsoids} & \textit{CurvedPolytopes} & \textit{Self-Ada-RP}  & \textit{Triangles}&\textit{Trapezoids}&\textit{SymmetricBells}&\textit{Self-Ada-RT}& \ourrep&\ourrep C
\\
\bhline{1pt}
{\texttt{cae}} & \cellcolor{p}71.81 $-$ & \cellcolor{g}78.01 $+$ & 77.22 $+$ & 76.81 $+$ & 76.94 $+$ & 72.73 $\sim$ & 76.30 $+$ & 77.27 $+$ & 73.66 $\sim$ & 74.12 \\
\texttt{can} & 97.19 $\sim$ & 98.30 $+$ & 97.29 $\sim$ & 97.34 $\sim$ & \cellcolor{p}94.22 $-$ & 94.29 $-$ & \cellcolor{g}99.13 $+$ & 95.57 $-$ & 97.32 $\sim$ & 97.54 \\
\texttt{car} & 93.89 $-$ & 87.13 $-$ & 88.11 $-$ & 93.82 $-$ & 90.24 $-$ & 89.80 $-$ & \cellcolor{p}83.85 $-$ & 92.05 $-$ & 95.00 $-$ & \cellcolor{g}95.29 \\
\texttt{chk} & 60.66 $-$ & 57.50 $-$ & 57.44 $-$ & 58.30 $-$ & 63.27 $-$ & \cellcolor{p}53.82 $-$ & 58.61 $-$ & 83.98 $-$ & \cellcolor{g}85.59 $\sim$ & 85.02 \\
\texttt{cmx} & 76.87 $-$ & 64.46 $-$ & 53.27 $-$ & 83.07 $-$ & \cellcolor{p}7.807 $-$ & 91.14 $-$ & 12.62 $-$ & 90.55 $-$ & 94.64 $-$ & \cellcolor{g}95.17 \\
\texttt{col} & 76.01 $-$ & 77.92 $-$ & 83.81 $\sim$ & 83.51 $\sim$ & 81.65 $-$ & \cellcolor{p}74.93 $-$ & 77.81 $-$ & \cellcolor{g}84.21 $\sim$ & 83.41 $\sim$ & 83.56 \\
\texttt{dbt} & 78.12 $-$ & 79.43 $-$ & 79.35 $-$ & 80.36 $-$ & 78.52 $-$ & \cellcolor{p}76.83 $-$ & 78.14 $-$ & 80.28 $-$ & \cellcolor{g}81.09 $\sim$ & 81.05 \\
\texttt{ecl} & \cellcolor{p}78.55 $-$ & 89.17 $\sim$ & 88.51 $-$ & 89.33 $\sim$ & 87.74 $-$ & 82.47 $-$ & 88.38 $-$ & \cellcolor{g}90.17 $\sim$ & 89.86 $\sim$ & 89.91 \\
\texttt{frt} & 93.67 $\sim$ & \cellcolor{g}94.32 $\sim$ & 88.37 $-$ & 89.60 $-$ & \cellcolor{p}81.82 $-$ & 88.99 $-$ & 90.23 $-$ & 89.40 $-$ & 93.66 $\sim$ & 93.90 \\
\texttt{gls} & \cellcolor{p}65.19 $-$ & 77.99 $-$ & 79.03 $-$ & 78.33 $-$ & 80.61 $-$ & 69.13 $-$ & 81.91 $-$ & 85.10 $\sim$ & \cellcolor{g}85.75 $\sim$ & 85.38 \\
\texttt{hcl} & 89.86 $-$ & \cellcolor{g}97.75 $+$ & 87.80 $-$ & 88.93 $-$ & 86.64 $-$ & \cellcolor{p}86.00 $-$ & 89.61 $-$ & 86.18 $-$ & 91.74 $\sim$ & 91.58 \\
\texttt{mam} & \cellcolor{g}83.10 $\sim$ & 81.14 $-$ & 82.16 $-$ & 82.46 $\sim$ & 81.26 $-$ & 80.10 $-$ & \cellcolor{p}79.76 $-$ & 82.21 $\sim$ & 82.90 $\sim$ & 82.80 \\
\texttt{mop} & 90.19 $-$ & 86.18 $-$ & 84.97 $-$ & 93.53 $\sim$ & 88.15 $-$ & 84.36 $-$ & \cellcolor{p}84.24 $-$ & 88.92 $-$ & 93.69 $\sim$ & \cellcolor{g}93.73 \\
{\texttt{mul}} & 98.94 $-$ & 98.71 $-$ & 99.40 $\sim$ & \cellcolor{g}99.54 $\sim$ & \cellcolor{p}93.44 $-$ & 98.68 $-$ & 98.26 $-$ & 97.71 $-$ & 99.24 $\sim$ & 99.48 \\
\texttt{mux} & 61.96 $-$ & 67.04 $-$ & 60.07 $-$ & 69.69 $-$ & \cellcolor{p}50.41 $-$ & 68.07 $-$ & 58.96 $-$ & 88.86 $-$ & 96.25 $-$ & \cellcolor{g}97.59 \\
{\texttt{nph}} & 70.86 $-$ & 75.70 $-$ & \cellcolor{p}66.25 $-$ & 67.39 $-$ & 73.69 $-$ & 70.00 $-$ & 72.22 $-$ & 75.22 $-$ & \cellcolor{g}76.86 $\sim$ & 76.67 \\
\texttt{pdy} & 89.16 $-$ & 86.65 $-$ & 88.16 $-$ & 88.10 $-$ & 87.96 $-$ & \cellcolor{p}78.40 $-$ & 85.78 $-$ & 89.69 $-$ & 90.23 $\sim$ & \cellcolor{g}90.26 \\
\texttt{pis} & 87.54 $\sim$ & 87.15 $\sim$ & 86.59 $-$ & 86.99 $-$ & 87.72 $+$ & \cellcolor{p}86.50 $-$ & \cellcolor{g}87.78 $+$ & 87.13 $\sim$ & 87.61 $\sim$ & 87.38 \\
\texttt{pmp} & 87.24 $-$ & 88.02 $\sim$ & 87.53 $-$ & 87.60 $-$ & 87.76 $-$ & \cellcolor{p}86.91 $-$ & 87.09 $-$ & 87.85 $\sim$ & 87.81 $-$ & \cellcolor{g}88.11 \\
\texttt{rsn} & 85.02 $-$ & \cellcolor{g}86.32 $\sim$ & 85.95 $\sim$ & 86.16 $\sim$ & 85.86 $\sim$ & \cellcolor{p}84.49 $-$ & 85.69 $\sim$ & 86.13 $\sim$ & 85.80 $\sim$ & 86.00 \\
\texttt{soy} & 95.49 $+$ & \cellcolor{g}96.56 $+$ & 95.74 $+$ & 95.51 $+$ & \cellcolor{p}68.84 $-$ & 93.18 $+$ & 84.52 $\sim$ & 74.50 $-$ & 84.21 $\sim$ & 84.74 \\
\texttt{tae} & 59.23 $-$ & 63.04 $\sim$ & 64.69 $\sim$ & 66.12 $\sim$ & 64.49 $\sim$ & \cellcolor{p}54.17 $-$ & \cellcolor{g}71.93 $+$ & 70.99 $+$ & 63.75 $\sim$ & 64.00 \\
\texttt{wne} & \cellcolor{p}93.62 $-$ & 99.85 $+$ & 99.02 $+$ & 99.25 $+$ & 99.17 $+$ & 93.71 $-$ & \cellcolor{g}99.96 $+$ & 97.62 $\sim$ & 96.65 $-$ & 97.42 \\
\texttt{wpb} & 95.21 $-$ & 99.91 $+$ & 99.89 $+$ & 98.95 $+$ & 91.18 $-$ & \cellcolor{p}83.20 $-$ & \cellcolor{g}100.0 $+$ & 93.61 $-$ & 96.40 $\sim$ & 96.57 \\
\texttt{yst} & 60.73 $-$ & 62.76 $-$ & 61.33 $-$ & 62.66 $-$ & 63.08 $-$ & \cellcolor{p}51.46 $-$ & 60.96 $-$ & 66.20 $-$ & \cellcolor{g}68.44 $\sim$ & 68.24 \\
\bhline{1pt}
Rank & \textit{6.56} & \textit{4.76} & \textit{6.08} & \textit{4.68} & \textit{6.72} & \cellcolor{p}\textit{8.64} & \textit{6.28} & \textit{4.68} & \textit{3.64} & \cellcolor{g}\textit{2.96} \\
Position & \textit{8} & \textit{5} & \textit{6} & \textit{3} & \textit{9} & \textit{10} & \textit{7} & \textit{4} & \textit{2} & \textit{1} \\
$+/-/\sim$ & 1/20/4 & 6/13/6 & 4/16/5 & 4/13/8 & 3/20/2 & 1/23/1 & 6/17/2 & 2/15/8 & 0/5/20 & - \\
\bhline{1pt}
$p$-value & 4.54E-05 & 0.0626 & 0.0115 & 0.0342 & 3.81E-05 & 1.01E-05 & 0.00558 & 0.00278 & 0.0451 & - \\
$p_\text{Holm}$-value & 0.000318 & 0.103 & 0.0458 & 0.103 & 0.000305 & 9.07E-05 & 0.0279 & 0.0167 & 0.103 & - \\

\bhline{1pt}
\end{tabular}}
\end{center}
\end{table*}

\clearpage
\section{Experimental Results on Population Size}
\label{sec: population size}
{Table \ref{tb: population size} presents each representation's average population size and average rank across all datasets.}

{Note that rule compaction techniques \cite{tan2013rapid,liu2021comparison} can be applied post-training to remove inaccurate and redundant rules, significantly reducing the final rule count while maintaining classification performance. Moreover, the number of rules participating in decision-making (i.e., rules in $[M]$) is significantly smaller than the population size. This smaller set of active rules preserves interpretability by allowing users to focus only on rules that contribute to classifications.}

\begin{table*}[ht]
\begin{center}
\caption{Summary of Results, Displaying Average Population Size Across 30 Runs. ``$+$'', ``$-$'', and ``$\sim$''; ``Rank''; ``Position''; ``$p$-Value''; ``$p_\text{Holm}$-Value''; and Green- and Peach-Shaded Values Should be Interpreted as in Table \ref{tb: sup f1}.}
\label{tb: population size}
\normalsize
\resizebox{\textwidth}{!}
{
\begin{tabular}{c|cccc|cccc|cc}
\bhline{1pt}
&\multicolumn{4}{c|}{UCS} & \multicolumn{4}{c|}{Fuzzy-UCS} & \multicolumn{2}{c}{\ours}
\\
\cline{2-11}
ID.  & \textit{Hyperrectangles} & \textit{Hyperellipsoids} & \textit{CurvedPolytopes} & \textit{Self-Ada-RP}  & \textit{Triangles}&\textit{Trapezoids}&\textit{SymmetricBells}&\textit{Self-Ada-RT}& \ourrep&\ourrep C
\\
\bhline{1pt}
{\texttt{cae}} & 535.7 $-$ & \cellcolor{p}966.5 $-$ & 755.3 $-$ & 717.9 $-$ & 489.5 $\sim$ & 426.1 $\sim$ & \cellcolor{g}245.3 $+$ & 592.9 $-$ & 455.8 $\sim$ & 457.5 \\
\texttt{can} & 1754 $-$ & 1860 $-$ & \cellcolor{g}1704 $+$ & 1712 $\sim$ & 1873 $-$ & 1813 $-$ & 1786 $-$ & \cellcolor{p}1938 $-$ & 1746 $-$ & 1722 \\
\texttt{car} & 1578 $-$ & 1672 $-$ & 1652 $-$ & 1570 $-$ & 1726 $-$ & 1717 $-$ & 1683 $-$ & \cellcolor{p}1732 $-$ & \cellcolor{g}1367 $+$ & 1399 \\
\texttt{chk} & 1475 $-$ & 1527 $-$ & 1503 $-$ & 1504 $-$ & 1520 $-$ & \cellcolor{p}1536 $-$ & 1409 $-$ & 1418 $-$ & 1358 $\sim$ & \cellcolor{g}1354 \\
\texttt{cmx} & 1625 $-$ & 1665 $-$ & 1619 $-$ & 1581 $-$ & 1768 $-$ & 1688 $-$ & \cellcolor{p}1909 $-$ & 1712 $-$ & 1544 $-$ & \cellcolor{g}1525 \\
\texttt{col} & 1286 $\sim$ & 1450 $-$ & \cellcolor{p}1552 $-$ & 1542 $-$ & 1360 $-$ & 1299 $\sim$ & \cellcolor{g}1031 $+$ & 1508 $-$ & 1366 $-$ & 1309 \\
\texttt{dbt} & 1598 $-$ & 1652 $-$ & 1652 $-$ & 1637 $-$ & 1628 $-$ & 1644 $-$ & \cellcolor{g}1494 $+$ & \cellcolor{p}1681 $-$ & 1532 $\sim$ & 1519 \\
\texttt{ecl} & 1479 $-$ & 1541 $-$ & 1536 $-$ & 1537 $-$ & 1526 $-$ & 1494 $-$ & \cellcolor{g}1350 $+$ & \cellcolor{p}1584 $-$ & 1449 $-$ & 1408 \\
\texttt{frt} & 1828 $+$ & 1857 $-$ & \cellcolor{g}1771 $+$ & 1780 $+$ & 1891 $-$ & 1883 $-$ & 1893 $-$ & \cellcolor{p}1964 $-$ & 1851 $\sim$ & 1845 \\
\texttt{gls} & 1442 $\sim$ & 1578 $-$ & 1487 $-$ & 1490 $-$ & 1516 $-$ & 1378 $+$ & \cellcolor{g}1215 $+$ & \cellcolor{p}1651 $-$ & 1470 $-$ & 1440 \\
\texttt{hcl} & 1773 $+$ & \cellcolor{g}1726 $+$ & 1758 $+$ & 1760 $+$ & 1838 $-$ & 1844 $-$ & 1768 $+$ & \cellcolor{p}1942 $-$ & 1803 $\sim$ & 1798 \\
\texttt{mam} & 1500 $\sim$ & 1587 $-$ & 1552 $-$ & 1541 $-$ & 1556 $-$ & 1578 $-$ & \cellcolor{p}1722 $-$ & 1612 $-$ & 1512 $-$ & \cellcolor{g}1490 \\
\texttt{mop} & 1670 $-$ & 1718 $-$ & 1699 $-$ & 1684 $-$ & 1725 $-$ & 1720 $-$ & 1687 $-$ & \cellcolor{p}1761 $-$ & 1611 $-$ & \cellcolor{g}1597 \\
{\texttt{mul}} & 1675 $+$ & 1676 $+$ & 1695 $\sim$ & 1686 $+$ & 1766 $-$ & 1770 $-$ & \cellcolor{g}1080 $+$ & \cellcolor{p}1877 $-$ & 1711 $\sim$ & 1709 \\
\texttt{mux} & 1793 $-$ & \cellcolor{p}1945 $-$ & 1768 $-$ & 1775 $-$ & 1928 $-$ & 1809 $-$ & 1790 $-$ & 1870 $-$ & 1640 $-$ & \cellcolor{g}1559 \\
{\texttt{nph}} & 1691 $+$ & 1713 $\sim$ & 1729 $-$ & 1729 $-$ & 1734 $-$ & 1741 $-$ & \cellcolor{g}1548 $+$ & \cellcolor{p}1834 $-$ & 1713 $\sim$ & 1712 \\
\texttt{pdy} & 1411 $-$ & 1439 $-$ & 1446 $-$ & 1450 $-$ & \cellcolor{p}1533 $-$ & 1469 $-$ & 1406 $-$ & 1471 $-$ & 1354 $\sim$ & \cellcolor{g}1353 \\
\texttt{pis} & 1670 $-$ & 1767 $-$ & 1754 $-$ & 1707 $-$ & 1743 $-$ & 1765 $-$ & 1761 $-$ & \cellcolor{p}1792 $-$ & \cellcolor{g}1617 $+$ & 1636 \\
\texttt{pmp} & 1628 $-$ & 1719 $-$ & 1701 $-$ & 1671 $-$ & 1708 $-$ & 1722 $-$ & 1719 $-$ & \cellcolor{p}1743 $-$ & \cellcolor{g}1539 $+$ & 1561 \\
\texttt{rsn} & 1517 $-$ & 1617 $-$ & \cellcolor{p}1641 $-$ & 1601 $-$ & 1590 $-$ & 1581 $-$ & 1529 $-$ & 1637 $-$ & \cellcolor{g}1393 $\sim$ & 1394 \\
\texttt{soy} & 1789 $+$ & 1808 $+$ & 1770 $+$ & \cellcolor{g}1759 $+$ & 1887 $-$ & 1891 $-$ & 1799 $+$ & \cellcolor{p}1970 $-$ & 1872 $\sim$ & 1871 \\
\texttt{tae} & 1075 $\sim$ & \cellcolor{p}1349 $-$ & 1235 $-$ & 1260 $-$ & 1182 $-$ & 826.8 $+$ & \cellcolor{g}690.2 $+$ & 1296 $-$ & 1098 $-$ & 1077 \\
\texttt{wne} & 1399 $-$ & \cellcolor{p}1703 $-$ & 1559 $-$ & 1609 $-$ & 1569 $-$ & 857.8 $+$ & \cellcolor{g}745.4 $+$ & 1676 $-$ & 1421 $-$ & 1328 \\
\texttt{wpb} & 1828 $+$ & 1849 $-$ & 1720 $+$ & 1748 $+$ & 1881 $-$ & 1828 $+$ & \cellcolor{g}835.2 $+$ & \cellcolor{p}1969 $-$ & 1847 $\sim$ & 1842 \\
\texttt{yst} & 1637 $-$ & 1640 $-$ & 1584 $-$ & 1599 $-$ & 1602 $-$ & 1638 $-$ & \cellcolor{g}1556 $+$ & \cellcolor{p}1669 $-$ & 1571 $\sim$ & 1568 \\
\bhline{1pt}
Rank & \textit{3.88} & \textit{7.40} & \textit{5.32} & \textit{5.08} & \textit{7.32} & \textit{6.48} & \textit{3.80} & \cellcolor{p}\textit{9.20} & \textit{3.64} & \cellcolor{g}\textit{2.88} \\
Position & \textit{4} & \textit{9} & \textit{6} & \textit{5} & \textit{8} & \textit{7} & \textit{3} & \textit{10} & \textit{2} & \textit{1} \\
$+/-/\sim$ & 6/15/4 & 3/21/1 & 5/19/1 & 5/19/1 & 0/24/1 & 4/19/2 & 13/12/0 & 0/25/0 & 3/10/12 & - \\
\bhline{1pt}
$p$-value & 0.00613 & 8.17E-06 & 0.000715 & 0.000912 & 5.96E-08 & 0.00737 & 0.653 & 5.96E-08 & 0.00631 & - \\
$p_\text{Holm}$-value & 0.0245 & 5.72E-05 & 0.00429 & 0.00456 & 5.36E-07 & 0.0245 & 0.653 & 5.36E-07 & 0.0245 & - \\

\bhline{1pt}
\end{tabular}}
\end{center}
\end{table*}

\clearpage

\section{Experimental Results on \ours\ With $s_0\in\{1.0, 1.5, 2.0, 2.5, 3.0,3.5,4.0,4.5,5.0\}$}
\label{sec: sup experimental results on ours with s0}
{Table \ref{tb: s0_test} presents the test classification accuracy of \ours\ with \ourrep C for various $s_0$ settings ($s_0\in\{1.0,1.5,2.0,2.5,3.0,3.5,4.0,4.5,5.0\}$) across all datasets.}

\begin{table*}[ht]
\begin{center}
\caption{Summary of Results, Displaying Average Test Classification Accuracy of \ours\ With Various $s_0$ Settings Across 30 Runs. ``$+$'', ``$-$'', and ``$\sim$'' for Significantly Better, Worse, and Competitive Compared to the Default $s_0=1.0$ Setting in Test Classification Accuracy, Respectively. ``Rank''; ``Position''; ``$p$-Value''; ``$p_\text{Holm}$-Value''; and Green- and Peach-Shaded Values Should be Interpreted as in Table \ref{tb: sup sota}.}
\label{tb: s0_test}
\normalsize
\scalebox{0.825}{\begin{tabular}{c|c|cccccccc}
\bhline{1pt}
$s_0$ & 1.0 & 1.5 & 2.0 & 2.5 & 3.0 & 3.5 & 4.0 & 4.5 & 5.0
\\
\bhline{1pt}
{\texttt{cae}} & 62.08 & \cellcolor{p}59.58 $\sim$ & 62.92 $\sim$ & 64.58 $\sim$ & 63.33 $\sim$ & 63.75 $\sim$ & 62.50 $\sim$ & 62.08 $\sim$ & \cellcolor{g}65.00 $\sim$ \\
\texttt{can} & \cellcolor{g}95.32 & 94.44 $\sim$ & 93.10 $-$ & 92.75 $-$ & 92.22 $-$ & \cellcolor{p}91.05 $-$ & 91.52 $-$ & 91.70 $-$ & 91.75 $-$ \\
\texttt{car} & \cellcolor{g}94.58 & 92.19 $-$ & 91.47 $-$ & 90.87 $-$ & 89.67 $-$ & 89.35 $-$ & 89.18 $-$ & \cellcolor{p}88.73 $-$ & 88.91 $-$ \\
\texttt{chk} & 81.32 & \cellcolor{g}81.58 $\sim$ & 81.33 $\sim$ & 81.55 $\sim$ & 80.76 $\sim$ & 81.03 $\sim$ & 79.65 $\sim$ & 80.61 $\sim$ & \cellcolor{p}79.54 $-$ \\
\texttt{cmx} & \cellcolor{g}92.86 & 91.33 $-$ & 88.90 $-$ & 84.92 $-$ & 78.87 $-$ & 73.84 $-$ & 67.46 $-$ & 61.81 $-$ & \cellcolor{p}59.57 $-$ \\
\texttt{col} & 77.63 & 78.49 $\sim$ & \cellcolor{g}79.03 $\sim$ & 78.60 $\sim$ & \cellcolor{p}76.24 $\sim$ & 77.63 $\sim$ & 78.06 $\sim$ & 77.63 $\sim$ & 78.28 $\sim$ \\
\texttt{dbt} & \cellcolor{g}74.89 & 74.11 $\sim$ & 73.42 $\sim$ & 74.72 $\sim$ & 73.29 $\sim$ & 73.07 $-$ & \cellcolor{p}72.90 $-$ & 73.03 $\sim$ & 73.07 $-$ \\
\texttt{ecl} & 85.39 & 85.69 $\sim$ & 83.73 $\sim$ & 84.22 $\sim$ & 85.10 $\sim$ & \cellcolor{g}86.18 $\sim$ & 84.71 $\sim$ & \cellcolor{p}83.33 $\sim$ & 85.39 $\sim$ \\
\texttt{frt} & \cellcolor{g}85.19 & 83.41 $-$ & 81.70 $-$ & 82.00 $-$ & 81.59 $-$ & 82.33 $-$ & \cellcolor{p}80.41 $-$ & 80.56 $-$ & 81.15 $-$ \\
\texttt{gls} & 67.73 & 66.21 $\sim$ & 67.88 $\sim$ & 66.36 $\sim$ & 66.82 $\sim$ & \cellcolor{g}68.18 $\sim$ & \cellcolor{p}64.85 $\sim$ & 66.36 $\sim$ & 65.76 $\sim$ \\
\texttt{hcl} & \cellcolor{g}70.90 & 70.00 $\sim$ & 68.38 $\sim$ & 67.93 $\sim$ & 68.20 $\sim$ & \cellcolor{p}67.39 $-$ & 68.83 $\sim$ & 69.19 $\sim$ & 68.65 $\sim$ \\
\texttt{mam} & 81.41 & \cellcolor{g}81.51 $\sim$ & 81.17 $\sim$ & 81.17 $\sim$ & 80.96 $\sim$ & 81.24 $\sim$ & 81.37 $\sim$ & \cellcolor{p}80.86 $\sim$ & 81.17 $\sim$ \\
\texttt{mop} & \cellcolor{g}91.71 & 90.71 $-$ & 89.19 $-$ & 88.24 $-$ & 86.71 $-$ & 86.36 $-$ & 84.93 $-$ & 84.02 $-$ & \cellcolor{p}82.93 $-$ \\
{\texttt{mul}} & \cellcolor{g}86.55 & 76.07 $-$ & 74.40 $-$ & 72.98 $-$ & 74.52 $-$ & 73.69 $-$ & \cellcolor{p}71.07 $-$ & 71.67 $-$ & 71.19 $-$ \\
\texttt{mux} & \cellcolor{g}96.44 & 90.42 $-$ & 75.03 $-$ & 63.24 $-$ & 60.14 $-$ & 58.86 $-$ & 56.88 $-$ & 56.78 $-$ & \cellcolor{p}55.63 $-$ \\
{\texttt{nph}} & \cellcolor{p}53.98 & 54.44 $\sim$ & 56.06 $\sim$ & 54.68 $\sim$ & 55.51 $\sim$ & 55.19 $\sim$ & 54.95 $\sim$ & 55.46 $\sim$ & \cellcolor{g}56.67 $+$ \\
\texttt{pdy} & 89.02 & \cellcolor{g}89.30 $\sim$ & 88.98 $\sim$ & 88.75 $\sim$ & 88.65 $\sim$ & 88.78 $\sim$ & 88.72 $\sim$ & 88.68 $\sim$ & \cellcolor{p}88.64 $\sim$ \\
\texttt{pis} & \cellcolor{g}86.60 & 85.94 $-$ & 86.16 $\sim$ & 85.66 $-$ & 85.97 $\sim$ & 85.75 $-$ & 86.02 $\sim$ & 85.75 $-$ & \cellcolor{p}85.47 $-$ \\
\texttt{pmp} & \cellcolor{g}87.19 & 86.96 $\sim$ & 86.81 $\sim$ & 86.59 $-$ & 86.72 $\sim$ & 86.59 $-$ & 86.52 $-$ & \cellcolor{p}86.36 $\sim$ & 86.61 $-$ \\
\texttt{rsn} & \cellcolor{g}85.52 & 84.67 $-$ & 85.48 $\sim$ & 85.07 $\sim$ & \cellcolor{p}84.59 $-$ & 84.81 $\sim$ & 85.30 $\sim$ & 84.89 $\sim$ & 85.11 $\sim$ \\
\texttt{soy} & \cellcolor{g}69.52 & 60.92 $-$ & 60.72 $-$ & 57.73 $-$ & 58.55 $-$ & 59.37 $-$ & 59.18 $-$ & 56.91 $-$ & \cellcolor{p}56.71 $-$ \\
\texttt{tae} & \cellcolor{p}51.04 & 52.08 $\sim$ & \cellcolor{g}55.62 $+$ & 55.00 $\sim$ & 55.21 $\sim$ & 54.38 $\sim$ & 51.88 $\sim$ & 54.17 $\sim$ & 55.00 $\sim$ \\
\texttt{wne} & 92.04 & 92.04 $\sim$ & 92.04 $\sim$ & 92.41 $\sim$ & 91.30 $\sim$ & \cellcolor{g}93.15 $\sim$ & \cellcolor{p}90.37 $\sim$ & 90.93 $\sim$ & 90.56 $\sim$ \\
\texttt{wpb} & 73.83 & \cellcolor{g}74.17 $\sim$ & 72.83 $\sim$ & \cellcolor{g}74.17 $\sim$ & 73.83 $\sim$ & 72.00 $\sim$ & \cellcolor{p}71.50 $\sim$ & 72.33 $\sim$ & 71.67 $\sim$ \\
\texttt{yst} & \cellcolor{g}59.71 & 58.14 $\sim$ & 57.36 $-$ & 56.60 $-$ & 57.27 $-$ & \cellcolor{p}56.17 $-$ & 56.85 $-$ & 56.47 $-$ & 56.38 $-$ \\
\bhline{1pt}
Rank & \cellcolor{g}\textit{2.74} & \textit{3.32} & \textit{3.46} & \textit{4.56} & \textit{5.26} & \textit{5.32} & \textit{6.64} & \cellcolor{p}\textit{7.14} & \textit{6.56} \\
Position & \textit{1} & \textit{2} & \textit{3} & \textit{4} & \textit{5} & \textit{6} & \textit{8} & \textit{9} & \textit{7} \\
$+/-/\sim$ & - & 0/9/16 & 1/9/15 & 0/11/14 & 0/10/15 & 0/13/12 & 0/11/14 & 0/10/15 & 1/13/11 \\
\bhline{1pt}
$p$-value & - & 0.00390 & 0.00673 & 0.00964 & 0.000963 & 0.0059 & 8.80E-05 & 0.000108 & 0.00281 \\
$p_\text{Holm}$-value & - & 0.0156 & 0.0177 & 0.0177 & 0.00578 & 0.0177 & 0.000704 & 0.000754 & 0.0141 \\

\bhline{1pt}
\end{tabular}}
\end{center}
\end{table*}

\clearpage

\twocolumn
\section{Analysis of Rule Representation Adaptation Across Datasets on \ours}
\label{sec: sup analysis of rule representation adaptation across dataset}
{To understand how \ours\ with FBRC adapts its rule representation across different problem types, we analyze the proportion of hyperrectangular crisp rules (rules where $\alpha_i^k=\beta_i^k=1.0$ for all dimensions $i$) versus fuzzy rules generated by \ours\ for all datasets used in Section \ref{sec: experiment}. Table \ref{tb: sup crisp fuzzy propotion} shows these results, revealing that \ours\ generates different proportions of crisp and fuzzy rules for each dataset.

The table offers several insights:
\begin{itemize}
\item For problems with high uncertainty (i.e., \texttt{hcl}: 23.80\% missing, \texttt{soy}: 9.78\% missing), \ours\ generated predominantly fuzzy rules (73.07\%, 77.03\% respectively) and significantly outperformed UCS with \textit{Hyperrectangles}, demonstrating FBRC's effectiveness in handling dataset uncertainty.

    \item For problems where UCS with \textit{Hyperrectangles} recorded worst test accuracy among all 10 representations (i.e., \texttt{col}, \texttt{ecl}, \texttt{gls}, cf. Table \ref{tb: result}), \ours\ generated predominantly fuzzy rules (62.11\%, 54.82\%, 60.53\% respectively) and achieved significantly higher test classification accuracy. This emphasizes FBRC's effectiveness for problems where pure hyperrectangular crisp rules struggle.

\end{itemize}

In summary, the above analysis concludes that \ours\ adapts representations to dataset and subregion characteristics.}

{This analysis raises one important question regarding our system's adaptive capabilities: Does \ours\ discover the true nature of decision boundaries (crisp, fuzzy, or mixed) in a problem, or does it simply adapt representations to optimize accuracy? This distinction is particularly relevant for real-world problems where the true nature of decision boundaries is often unknown. Our analysis of the rotated checkerboard problem provides some insight into this question. As shown in Section \ref{ss: visualization of decision boundaries and rule discovery strategy of ours}, \ours\ naturally employs crisp rules in regions far from diagonal boundaries and fuzzy rules near these boundaries. This adaptation aligns with our intuitive understanding of the problem's structure. Specifically, regions far from decision boundaries can be cleanly separated with crisp rules, while boundary regions benefit from fuzzy rules.

However, for most real-world problems, determining the true nature of decision boundaries remains challenging. While \ours\ primarily aims to optimize classification accuracy through representation adaptation, the results on the rotated checkerboard problem suggest this adaptation may naturally discover inherent boundary characteristics. This relationship between accuracy-driven adaptation and the discovery of inherent problem structure represents an important direction for future research into LCS's local adaptation capabilities.}\label{r4-2}

 \begin{table}[ht]
\begin{center}
\caption{Proportion of Crisp and Fuzzy Rules Generated by \ours\ With FBRC at the End of Training (i.e., 50th Epoch). The Bar Chart Shows the Proportion of Crisp ({\blue $\blacksquare$}) and Fuzzy Rules ({\red $\blacksquare$}).
}
\label{tb: sup crisp fuzzy propotion}
\normalsize
\scalebox{0.8}
{
\begin{tabular}{c|ccc}
\bhline{1pt}
 
ID. & Proportional Bar Chart & Crisp ({\blue $\blacksquare$}) & Fuzzy ({\red $\blacksquare$})\\
\bhline{1pt}
{\texttt{cae}} & \barchart{52.26}{47.74} \\
\texttt{can} & \barchart{66.30}{33.70} \\
\texttt{car} & \barchart{83.18}{16.82} \\
\texttt{chk} & \barchart{46.83}{53.17} \\
\texttt{cmx} & \barchart{57.13}{42.87} \\
\texttt{col} & \barchart{37.89}{62.11} \\
\texttt{dbt} & \barchart{56.22}{43.78} \\
\texttt{ecl} & \barchart{45.18}{54.82} \\
\texttt{frt} & \barchart{32.60}{67.40} \\
\texttt{gls} & \barchart{39.47}{60.53} \\
\texttt{hcl} & \barchart{26.93}{73.07} \\
\texttt{mam} & \barchart{58.51}{41.49} \\
\texttt{mop} & \barchart{72.59}{27.41} \\
{\texttt{mul}} & \barchart{53.45}{46.55} \\
\texttt{mux} & \barchart{55.41}{44.59} \\
{\texttt{nph}} & \barchart{19.46}{80.54} \\
\texttt{pdy} & \barchart{77.78}{22.22} \\
\texttt{pis} & \barchart{68.32}{31.68} \\
\texttt{pmp} & \barchart{77.90}{22.10} \\
\texttt{rsn} & \barchart{62.67}{37.33} \\
\texttt{soy} & \barchart{22.97}{77.03} \\
\texttt{tae} & \barchart{34.19}{65.81} \\
\texttt{wne} & \barchart{70.10}{29.90} \\
\texttt{wpb} & \barchart{29.14}{70.86} \\
\texttt{yst} & \barchart{27.74}{72.26} \\
\bhline{1pt}
\end{tabular}
}
\end{center}
\vspace{-2mm}
\end{table}
\end{appendices}

\bibliographystyle{IEEEtran}
\bibliography{tevc}

\begin{thebibliography}{10}
\providecommand{\url}[1]{#1}
\csname url@samestyle\endcsname
\providecommand{\newblock}{\relax}
\providecommand{\bibinfo}[2]{#2}
\providecommand{\BIBentrySTDinterwordspacing}{\spaceskip=0pt\relax}
\providecommand{\BIBentryALTinterwordstretchfactor}{4}
\providecommand{\BIBentryALTinterwordspacing}{\spaceskip=\fontdimen2\font plus
\BIBentryALTinterwordstretchfactor\fontdimen3\font minus \fontdimen4\font\relax}
\providecommand{\BIBforeignlanguage}[2]{{%
\expandafter\ifx\csname l@#1\endcsname\relax
\typeout{** WARNING: IEEEtran.bst: No hyphenation pattern has been}%
\typeout{** loaded for the language `#1'. Using the pattern for}%
\typeout{** the default language instead.}%
\else
\language=\csname l@#1\endcsname
\fi
#2}}
\providecommand{\BIBdecl}{\relax}
\BIBdecl

\bibitem{shiraishi2025evolutionary}
H.~Shiraishi, Y.~Hayamizu, T.~Hashiyama, K.~Takadama, H.~Ishibuchi, and M.~Nakata, ``Evolutionary co-optimization of rule shape and fuzziness in rule-based machine learning,'' \emph{Authorea Preprints}, 2025.

\bibitem{urbanowicz2017introduction}
R.~J. Urbanowicz and W.~N. Browne, \emph{Introduction to Learning Classifier Systems}, 1st~ed.\hskip 1em plus 0.5em minus 0.4em\relax Springer Publishing Company, Incorporated, 2017.

\bibitem{butz2006rule}
M.~V. Butz, \emph{Rule-based evolutionary online learning systems}.\hskip 1em plus 0.5em minus 0.4em\relax Springer, 2006.

\bibitem{kovacs1998xcs}
T.~Kovacs, ``{XCS} classifier system reliably evolves accurate, complete, and minimal representations for {B}oolean functions,'' in \emph{Soft computing in engineering design and manufacturing}.\hskip 1em plus 0.5em minus 0.4em\relax Springer, 1998, pp. 59--68.

\bibitem{preen2021autoencoding}
R.~J. Preen, S.~W. Wilson, and L.~Bull, ``Autoencoding with a classifier system,'' \emph{IEEE Trans. Evol. Comput.}, vol.~25, no.~6, pp. 1079--1090, 2021.

\bibitem{bernado2003accuracy}
E.~Bernad\'{o}-Mansilla and J.~M. Garrell-Guiu, ``Accuracy-based learning classifier systems: Models, analysis and applications to classification tasks,'' \emph{Evol. Comput.}, vol.~11, no.~3, p. 209–238, sep 2003.

\bibitem{woodward2024survival}
A.~Woodward, H.~Bandhey, J.~H. Moore, and R.~J. Urbanowicz, ``Survival-{LCS}: A rule-based machine learning approach to survival analysis,'' in \emph{Proc. Genet. Evol. Comput. Conf.}, 2024, pp. 431--439.

\bibitem{shiraishi2022can}
H.~Shiraishi, Y.~Hayamizu, H.~Sato, and K.~Takadama, ``Can the same rule representation change its matching area? enhancing representation in {XCS} for continuous space by probability distribution in multiple dimension,'' in \emph{Proc. Genet. Evol. Comput. Conf.}, 2022, p. 431–439.

\bibitem{wilson2000mining}
S.~W. Wilson, ``Mining oblique data with {XCS},'' in \emph{Int. Worksh. Learn. Classif. Syst.}, 2000, pp. 158--174.

\bibitem{butz2005kernel}
M.~V. Butz, ``Kernel-based, ellipsoidal conditions in the real-valued {XCS} classifier system,'' in \emph{Proc. 7th Annu. Conf. Genet. Evol. Comput.}, 2005, pp. 1835--1842.

\bibitem{wilson2008classifier}
S.~W. Wilson, ``Classifier conditions using gene expression programming,'' in \emph{Learn. Classif. Syst.}, 2008, pp. 206--217.

\bibitem{bull2002accuracy}
L.~Bull and T.~O'Hara, ``Accuracy-based neuro and neuro-fuzzy classifier systems,'' in \emph{Proc. 4th Annu. Conf. Genet. Evol. Comput.}, 2002, p. 905–911.

\bibitem{casillas2007fuzzy}
J.~Casillas, B.~Carse, and L.~Bull, ``Fuzzy-{XCS}: A michigan genetic fuzzy system,'' \emph{IEEE Trans. Fuzzy Syst.}, vol.~15, no.~4, pp. 536--550, Aug 2007.

\bibitem{shoeleh2011towards}
F.~Shoeleh, A.~Hamzeh, and S.~Hashemi, ``Towards final rule set reduction in {XCS}: A fuzzy representation approach,'' in \emph{Proc. 13th Annu. Conf. Genet. Evol. Comput.}, 2011, pp. 1211--1218.

\bibitem{shiraishi2023fuzzy}
H.~Shiraishi, Y.~Hayamizu, and T.~Hashiyama, ``Fuzzy-{UCS} revisited: Self-adaptation of rule representations in michigan-style learning fuzzy-classifier systems,'' in \emph{Proc. Genet. Evol. Comput. Conf.}, 2023, p. 548–557.

\bibitem{orriols2008fuzzy}
A.~Orriols-Puig, J.~Casillas, and E.~Bernad{\'o}-Mansilla, ``Fuzzy-{UCS}: A michigan-style learning fuzzy-classifier system for supervised learning,'' \emph{IEEE Trans. Evol. Comput.}, vol.~13, no.~2, pp. 260--283, April 2009.

\bibitem{wilson1998generalization}
S.~W. Wilson, ``Generalization in the {XCS} classifier system,'' \emph{Proc. Genetic Programming 1998}, 1998.

\bibitem{takadama2015extracting}
K.~Takadama and M.~Nakata, ``Extracting both generalized and specialized knowledge by {XCS} using attribute tracking and feedback,'' in \emph{2015 IEEE Congr. Evol. Comput. (CEC)}.\hskip 1em plus 0.5em minus 0.4em\relax IEEE, 2015, pp. 3034--3041.

\bibitem{crackel2017bayesian}
R.~Crackel and J.~Flegal, ``Bayesian inference for a flexible class of bivariate beta distributions,'' \emph{J. Stat. Comput. Simul.}, vol.~87, no.~2, pp. 295--312, 2017.

\bibitem{wilson1999xcsr}
S.~W. Wilson, ``Get real! {XCS} with continuous-valued inputs,'' in \emph{Int. Worksh. Learn. Classif. Syst.}\hskip 1em plus 0.5em minus 0.4em\relax Springer, 1999, pp. 209--219.

\bibitem{stone2003real}
C.~Stone and L.~Bull, ``For real! {XCS} with continuous-valued inputs,'' \emph{Evol. Comput.}, vol.~11, no.~3, p. 299–336, sep 2003.

\bibitem{dam2005real}
H.~H. Dam, H.~A. Abbass, and C.~Lokan, ``Be real! {XCS} with continuous-valued inputs,'' in \emph{Proc. 7th Annu. Worksh. Genet. Evol. Comput.}, 2005, pp. 85--87.

\bibitem{lanzi2006using}
P.~L. Lanzi and S.~W. Wilson, ``Using convex hulls to represent classifier conditions,'' in \emph{Proc. 8th Annu. Conf. Genet. Evol. Comput.}, 2006, pp. 1481--1488.

\bibitem{arif2017solving}
M.~H. Arif, J.~Li, and M.~Iqbal, ``Solving social media text classification problems using code fragment-based {XCSR},'' in \emph{2017 IEEE 29th Int. Conf. Tools Artif. Intell. (ICTAI)}, Nov 2017, pp. 485--492.

\bibitem{shiraishi2022beta}
H.~Shiraishi, Y.~Hayamizu, H.~Sato, and K.~Takadama, ``Beta distribution based {XCS} classifier system,'' in \emph{2022 IEEE Congr. Evol. Comput. (CEC)}, July 2022, pp. 1--8.

\bibitem{llora2002coevolving}
X.~Llora and J.~M. Garrell, ``Coevolving different knowledge representations with fine-grained parallel learning classifier systems,'' in \emph{Proc. 4th Annu. Conf. Genet. Evol. Comput.}, 2002, pp. 934--941.

\bibitem{bacardit2003evolving}
J.~Bacardit and J.~M. Garrell, ``Evolving multiple discretizations with adaptive intervals for a pittsburgh rule-based learning classifier system,'' in \emph{Genet. Evol. Comput. Conf.}\hskip 1em plus 0.5em minus 0.4em\relax Springer, 2003, pp. 1818--1831.

\bibitem{orriols2008approximate}
A.~Orriols-Puig, J.~Casillas, and E.~Bernad{\'o}-Mansilla, ``Approximate versus linguistic representation in {F}uzzy-{UCS},'' in \emph{Hybrid Artif. Intell. Syst.}, 2008, pp. 722--729.

\bibitem{tadokoro2021xcs}
M.~Tadokoro, H.~Sato, and K.~Takadama, ``{XCS} with weight-based matching in {VAE} latent space and additional learning of high-dimensional data,'' in \emph{2021 IEEE Congr. Evol. Comput. (CEC)}, June 2021, pp. 304--310.

\bibitem{butz2008function}
M.~V. Butz, P.~L. Lanzi, and S.~W. Wilson, ``Function approximation with {XCS}: Hyperellipsoidal conditions, recursive least squares, and compaction,'' \emph{IEEE Trans. Evol. Comput.}, vol.~12, no.~3, pp. 355--376, June 2008.

\bibitem{wilson1995xcs}
S.~W. Wilson, ``Classifier fitness based on accuracy,'' \emph{Evol. Comput.}, vol.~3, no.~2, p. 149–175, jun 1995.

\bibitem{ferreira2006gene}
C.~Ferreira, \emph{Gene expression programming: mathematical modeling by an artificial intelligence}.\hskip 1em plus 0.5em minus 0.4em\relax Springer, 2006, vol.~21.

\bibitem{iqbal2013reusing}
M.~Iqbal, W.~N. Browne, and M.~Zhang, ``Reusing building blocks of extracted knowledge to solve complex, large-scale boolean problems,'' \emph{IEEE Trans. Evol. Comput.}, vol.~18, no.~4, pp. 465--480, 2014.

\bibitem{ishibuchi2005comparison}
H.~Ishibuchi and Y.~Nojima, ``Comparison between fuzzy and interval partitions in evolutionary multiobjective design of rule-based classification systems,'' in \emph{14th IEEE Int. Conf. Fuzzy Syst.}, 2005, pp. 430--435.

\bibitem{orriols2011fuzzy}
A.~Orriols-Puig and J.~Casillas, ``Fuzzy knowledge representation study for incremental learning in data streams and classification problems,'' \emph{Soft Comput.}, vol.~15, no.~12, pp. 2389--2414, 2011.

\bibitem{alimi1997beta}
A.~M. Alimi, ``The beta fuzzy system: approximation of standard membership functions,'' \emph{Proc. 17eme Journees Tunisiennes d'Electrotechnique et d'Automatique: JTEA}, vol.~97, pp. 108--112, 1997.

\bibitem{alimi2003beta}
A.~M. Alimi, R.~Hassine, and M.~Selmi, ``Beta fuzzy logic systems: approximation properties in the mimo case,'' \emph{Int. J. Appl. Math. Comput. Sci.}, vol.~13, no.~2, pp. 225--238, 2003.

\bibitem{bouaziz2013hybrid}
S.~Bouaziz, H.~Dhahri, A.~M. Alimi, and A.~Abraham, ``A hybrid learning algorithm for evolving flexible beta basis function neural tree model,'' \emph{Neurocomputing}, vol. 117, pp. 107--117, 2013.

\bibitem{baklouti2018beta}
N.~Baklouti, A.~Abraham, and A.~M. Alimi, ``A beta basis function interval type-2 fuzzy neural network for time series applications,'' \emph{Eng. Appl. Artif. Intell.}, vol.~71, pp. 259--274, 2018.

\bibitem{ishibuchi2005rule}
H.~Ishibuchi and T.~Yamamoto, ``Rule weight specification in fuzzy rule-based classification systems,'' \emph{IEEE Trans. Fuzzy Syst.}, vol.~13, no.~4, pp. 428--435, Aug 2005.

\bibitem{butz2004toward}
M.~Butz, T.~Kovacs, P.~Lanzi, and S.~Wilson, ``Toward a theory of generalization and learning in {XCS},'' \emph{IEEE Trans. Evol. Comput.}, vol.~8, no.~1, pp. 28--46, Feb 2004.

\bibitem{shiraishi2022absumption}
H.~Shiraishi, Y.~Hayamizu, H.~Sato, and K.~Takadama, ``Absumption based on overgenerality and condition-clustering based specialization for {XCS} with continuous-valued inputs,'' in \emph{Proc. Genet. Evol. Comput. Conf.}, 2022, p. 422–430.

\bibitem{nakata2020learning}
M.~Nakata and W.~N. Browne, ``Learning optimality theory for accuracy-based learning classifier systems,'' \emph{IEEE Trans. Evol. Comput.}, vol.~25, no.~1, pp. 61--74, Feb 2021.

\bibitem{shoeleh2010handle}
F.~Shoeleh, A.~Hamzeh, and S.~Hashemi, ``To handle real valued input in {XCS}: using fuzzy hyper-trapezoidal membership in classifier condition,'' in \emph{Simul. Evol. Learn. 8th Int. Conf.}, 2010, pp. 55--64.

\bibitem{liu2019absumption}
Y.~Liu, W.~N. Browne, and B.~Xue, ``Absumption to complement subsumption in learning classifier systems,'' in \emph{Proc. Genet. Evol. Comput. Conf.}, 2019, pp. 410--418.

\bibitem{iacca2012ockham}
G.~Iacca, F.~Neri, E.~Mininno, Y.-S. Ong, and M.-H. Lim, ``Ockham’s razor in memetic computing: three stage optimal memetic exploration,'' \emph{Inform. Sci.}, vol. 188, pp. 17--43, 2012.

\bibitem{dua2019uci}
\BIBentryALTinterwordspacing
D.~Dua and C.~Graff, ``{UCI} machine learning repository,'' 2017. [Online]. Available: \url{http://archive.ics.uci.edu/ml}
\BIBentrySTDinterwordspacing

\bibitem{hamasaki2021minimum}
K.~Hamasaki and M.~Nakata, ``Minimum rule-repair algorithm for supervised learning classifier systems on real-valued classification tasks,'' in \emph{Int. Conf. Metaheur. Nat. Inspir. Comput.}, 2021, pp. 137--151.

\bibitem{tzima2013strength}
F.~A. Tzima and P.~A. Mitkas, ``Strength-based learning classifier systems revisited: Effective rule evolution in supervised classification tasks,'' \emph{Eng. Appl. Artif. Intell.}, vol.~26, no.~2, pp. 818--832, 2013.

\bibitem{urbanowicz2015exstracs}
R.~J. Urbanowicz and J.~H. Moore, ``{ExSTraCS} 2.0: description and evaluation of a scalable learning classifier system,'' \emph{Evol. Intell.}, vol.~8, no.~2, pp. 89--116, 2015.

\bibitem{wilson2002classifiers}
S.~W. Wilson, ``Classifiers that approximate functions,'' \emph{Natural Comput.}, vol.~1, no.~2, pp. 211--234, 04 2002.

\bibitem{butz2003analysis}
M.~V. Butz, D.~E. Goldberg, and K.~Tharakunnel, ``Analysis and improvement of fitness exploitation in xcs: Bounding models, tournament selection, and bilateral accuracy,'' \emph{Evol. Comput.}, vol.~11, no.~3, pp. 239--277, 2003.

\bibitem{koklu2021classification}
M.~Koklu, R.~Kursun, Y.~S. Taspinar, and I.~Cinar, ``Classification of date fruits into genetic varieties using image analysis,'' \emph{Math. Probl. Eng.}, vol. 2021, 2021.

\bibitem{chavarria2023conversion}
V.~Chavarria, G.~Espinosa-Ram{\'\i}rez, J.~Sotelo, J.~Flores-Rivera, O.~Anguiano, A.~C. Hern{\'a}ndez, E.~D. Guzm{\'a}n-R{\'\i}os, A.~Salazar, G.~Ordo{\~n}ez, and B.~Pineda, ``Conversion predictors of clinically isolated syndrome to multiple sclerosis in mexican patients: a prospective study,'' \emph{Archives of Medical Research}, vol.~54, no.~5, p. 102843, 2023.

\bibitem{malani2019npha}
P.~N. Malani, J.~Kullgren, and E.~Solway, ``National poll on healthy aging ({NPHA}), [united states], april 2017.'' 2019.

\bibitem{singh2022classification}
D.~Singh, Y.~S. Taspinar, R.~Kursun, I.~Cinar, M.~Koklu, I.~A. Ozkan, and H.-N. Lee, ``Classification and analysis of pistachio species with pre-trained deep learning models,'' \emph{Electronics}, vol.~11, no.~7, 2022.

\bibitem{koklu2021use}
M.~Koklu, S.~Sarigil, and O.~Ozbek, ``The use of machine learning methods in classification of pumpkin seeds (cucurbita pepo l.),'' \emph{Genet. Resour. Crop Evol.}, vol.~68, no.~7, pp. 2713--2726, 2021.

\bibitem{ccinar2020classification}
{\.I}.~{\c{C}}inar, M.~Koklu, and {\c{S}}.~Ta{\c{s}}dem{\.i}r, ``Classification of raisin grains using machine vision and artificial intelligence methods,'' \emph{Gazi M{\"u}hendislik Bilimleri Dergisi}, vol.~6, no.~3, pp. 200--209, 12 2020.

\bibitem{bezanson2017julia}
J.~Bezanson, A.~Edelman, S.~Karpinski, and V.~B. Shah, ``Julia: A fresh approach to numerical computing,'' \emph{SIAM Review}, vol.~59, no.~1, pp. 65--98, 2017.

\bibitem{rosenbauer2020generic}
L.~Rosenbauer, A.~Stein, and J.~H{\"a}hner, ``Generic approaches for parallel rule matching in learning classifier systems,'' in \emph{Proc. 2020 Genet. Evol. Comput. Conf. Companion}, 2020, pp. 1789--1797.

\bibitem{hinder2023model}
F.~Hinder, V.~Vaquet, J.~Brinkrolf, and B.~Hammer, ``Model-based explanations of concept drift,'' \emph{Neurocomput.}, vol. 555, p. 126640, 2023.

\bibitem{omrane2016fuzzy}
H.~Omrane, M.~S. Masmoudi, and M.~Masmoudi, ``Fuzzy logic based control for autonomous mobile robot navigation,'' \emph{Comput. Intell. Neurosci.}, vol. 2016, no.~1, p. 9548482, 2016.

\bibitem{ortigossa2024explainable}
E.~S. Ortigossa, T.~Gonçalves, and L.~G. Nonato, ``Explainable artificial intelligence (xai)—from theory to methods and applications,'' \emph{IEEE Access}, vol.~12, pp. 80\,799--80\,846, 2024.

\bibitem{wagner2022mechanisms}
A.~R.~M. Wagner and A.~Stein, ``Mechanisms to alleviate over-generalization in {XCS} for continuous-valued input spaces,'' \emph{SN Comput. Sci.}, vol.~3, no.~2, pp. 1--23, 2022.

\bibitem{heider2023suprb}
M.~Heider, H.~Stegherr, R.~Sraj, D.~Pätzel, J.~Wurth, and J.~Hähner, ``Sup{RB} in the context of rule-based machine learning methods: A comparative study,'' \emph{Appl. Soft Comput.}, p. 110706, 2023.

\bibitem{shiraishi2024variable}
H.~Shiraishi, R.~Ye, H.~Ishibuchi, and M.~Nakata, ``A variable-length fuzzy set representation for learning fuzzy-classifier systems,'' in \emph{Int. Conf. Parallel Problem Solving from Nature}, 2024, pp. 386--402.

\bibitem{guan2022prediction}
L.~Guan and R.~Tibshirani, ``Prediction and outlier detection in classification problems,'' \emph{Journal of the Royal Statistical Society Series B: Statistical Methodology}, vol.~84, no.~2, pp. 524--546, 2022.

\bibitem{tan2013rapid}
J.~Tan, J.~Moore, and R.~Urbanowicz, ``Rapid rule compaction strategies for global knowledge discovery in a supervised learning classifier system,'' in \emph{Proc. Artificial Life Conf.}, 2013, pp. 110--117.

\bibitem{liu2021comparison}
Y.~Liu, W.~N. Browne, and B.~Xue, ``A comparison of learning classifier systems’ rule compaction algorithms for knowledge visualization,'' \emph{ACM Trans. Evol. Learn. Optim.}, vol.~1, no.~3, pp. 1--38, 2021.

\end{thebibliography}
%

 \begin{IEEEbiography}
 [{\includegraphics[width=1in,height=1.25in,clip,keepaspectratio]{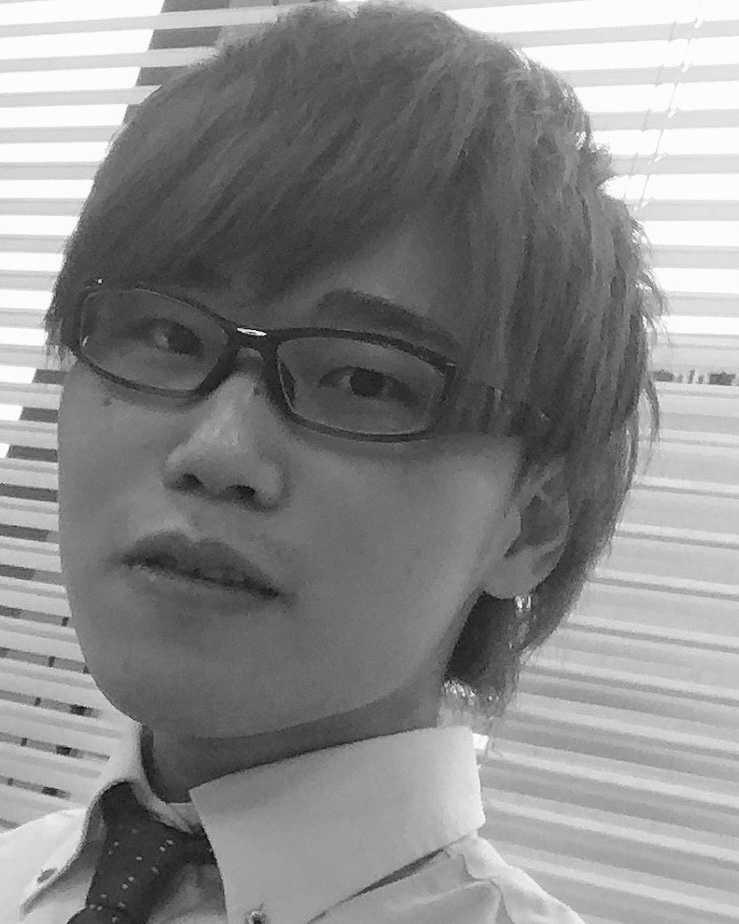}}]
 {Hiroki Shiraishi}
was born in Chiba, Japan, in 1999.
He received his B.E. and M.E. degrees in informatics from the University of Electro-Communications, Tokyo, Japan, in 2021 and 2023, respectively.

Since 2023, he has been a Ph.D. candidate in the Faculty of Engineering at Yokohama National University, Yokohama, Japan. 
His research interests include evolutionary machine learning, 
fuzzy systems, and 
learning classifier systems. 
His contributions have been published in leading journals and conferences on evolutionary computation, fuzzy systems, and artificial intelligence,
including 
\textsc{IEEE Transactions on Evolutionary Computation}, 
\textit{ACM Transactions on Evolutionary Learning and Optimization}, 
GECCO, IEEE CEC, EvoStar, 
PPSN, FUZZ-IEEE, and IJCAI.
He received a Best Paper Award at GECCO 2022 and a nomination for the Best Paper Award at GECCO 2023. He chaired the International Workshop on Evolutionary Rule-Based Machine Learning 
at GECCO in 2024 and 2025.
\end{IEEEbiography}
\begin{IEEEbiography}
[{\includegraphics[width=1in,height=1.25in,clip,keepaspectratio]{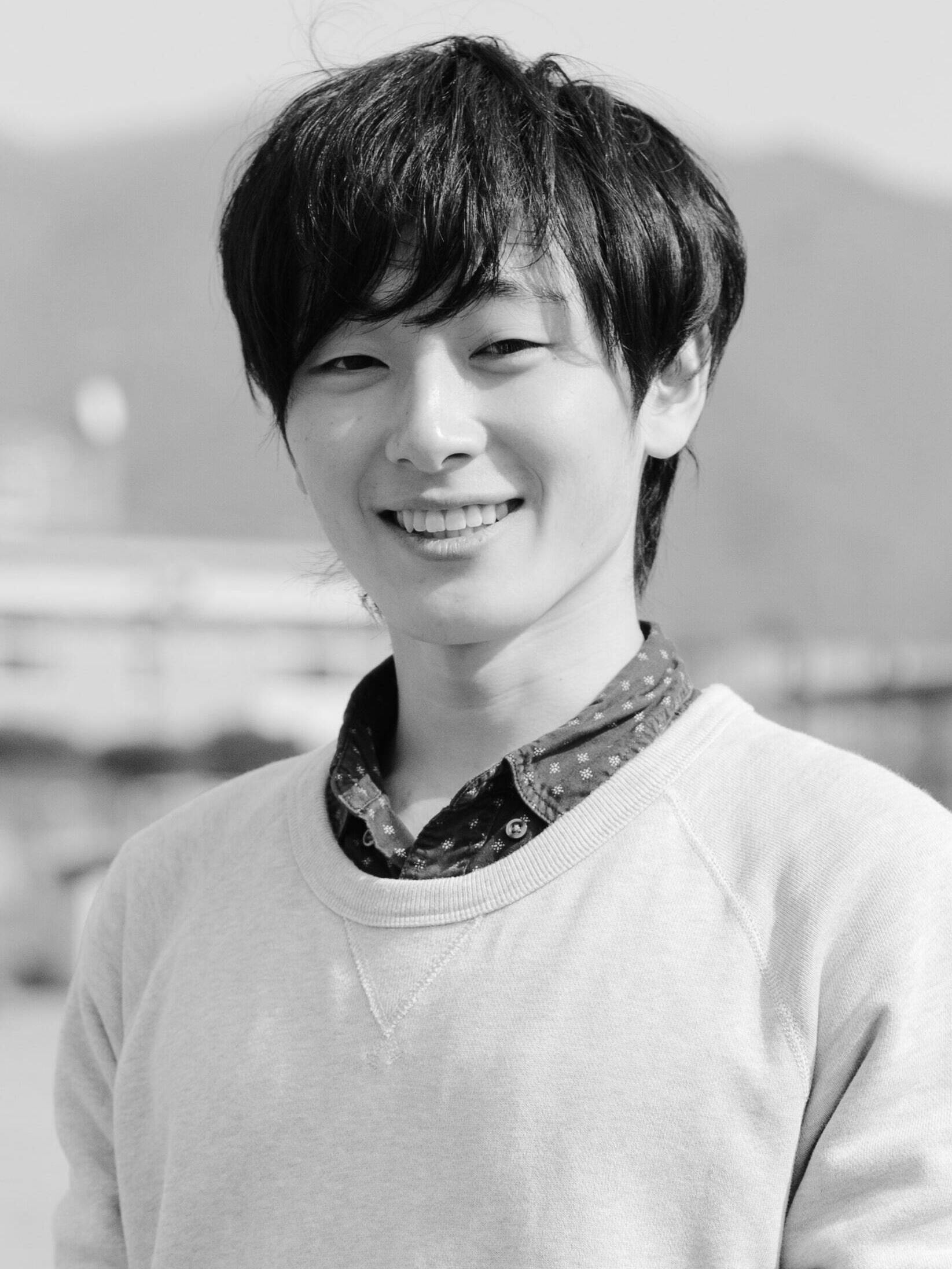}}]
{Yohei Hayamizu}
(Graduate Student Member, IEEE)
was born in Akita, Japan, in 1996. He received his B.E. in electrical engineering and computer science from Iwate University, Morioka, Japan, in 2018, and his M.E. in informatics from the University of Electro-Communications, Tokyo, Japan, in 2021. 

He joined the Ph.D. program in the Department of Computer Science at the State University of New York at Binghamton, NY, USA, in 2021. His research interests are artificial intelligence and robotics, specifically reinforcement learning, knowledge representation and reasoning, and learning classifier systems. 

\end{IEEEbiography}
\begin{IEEEbiography}
[{\includegraphics[width=1in,height=1.25in,clip,keepaspectratio]{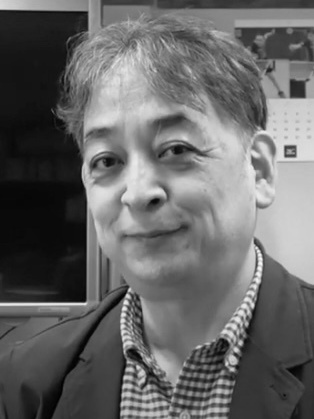}}]
{Tomonori Hashiyama} (Member, IEEE) was born in Hakodate, Japan, in 1967. He received his B.Eng., M.Eng. degrees in mechatronics, and Dr.Eng. degree in information electronics from Nagoya University, Nagoya, Japan, in 1991, 1993, and 1996, respectively.

He joined Nagoya University in 1996 and Nagoya City University, Nagoya, in 2000. Since 2003, he has been with
the University of Electro-Communications, Tokyo, Japan. He is a Professor and joined the West Tokyo Joint Doctoral Program for Sustainability Research with the Tokyo University of Foreign Studies, Fuchu, Japan, and the Tokyo University of Agriculture and Technology, Tokyo, in 2020.
His research interests include computational intelligence for human-computer interactions and its applications to real-world problems.

\end{IEEEbiography}
\begin{IEEEbiography}
[{\includegraphics[width=1in,height=1.25in,clip,keepaspectratio]{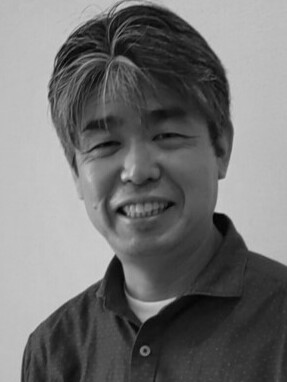}}]
{Keiki Takadama} 
(Member, IEEE) received his Dr.Eng degree from the University of
Tokyo, Tokyo, Japan, in 1998. He joined Advanced Telecommunications Research
Institute International, Kyoto, Japan, from 1998 to 2002 as a Visiting
Researcher and worked at the Tokyo Institute of Technology, Meguro, Japan, from 2002 to
2006 as a Lecturer. He moved to the University of
Electro-Communications, Tokyo, as an Associate Professor in 2006 and became a
Professor since 2011. 

He has been at the University of Tokyo
as a Professor since 2024. He served as the general chair of GECCO
2018.  His research
interests include evolutionary computation, reinforcement learning,
multiagent systems, autonomous systems, space intelligent systems, and
health care systems. He is a member of ACM and a member of
major AI- and informatics-related academic societies in Japan. He received a Best Paper Award from GECCO 2022 and IEEE CEC 2024.
\end{IEEEbiography}
\begin{IEEEbiography}
[{\includegraphics[width=1in,height=1.25in,clip,keepaspectratio]{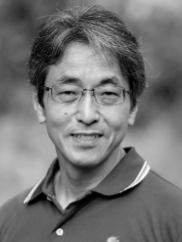}}]
{Hisao Ishibuchi} (Fellow, IEEE) received the B.S.
 and M.S. degrees from Kyoto University, Kyoto,
 Japan, in 1985 and 1987, respectively, and the Ph.D.
 degree from Osaka Prefecture University, Sakai,
 Japan, in 1992. 
 
 He is a Chair Professor at Southern University of Science and Technology, Shenzhen, China. He was the IEEE Computational Intelligence Society (CIS) Vice-President for Technical Activities in 2010-2013 and the Editor-in-Chief of IEEE Computational Intelligence Magazine in 2014-2019. 
Currently, he is an IEEE CIS Administrative Committee Member, and an Associate Editor of several journals such as \textsc{IEEE Transactions on Evolutionary Computation}, \textit{Evolutionary Computation Journal}, and \textit{ACM Computing Surveys}. He received a Fuzzy Systems Pioneer Award from IEEE CIS in 2019, an Outstanding Paper Award from \textsc{IEEE Transactions on Evolutionary Computation} in 2020, an Enrique Ruspini Award for Meritorious Service from IEEE CIS in 2023, and Best Paper Awards from FUZZ-IEEE 2009, 2011, EMO 2019, 2025, and GECCO 2004, 2017, 2018, 2020, 2021, 2024. He also received a JSPS prize in 2007.

\end{IEEEbiography}
\begin{IEEEbiography}
[{\includegraphics[width=1in,height=1.25in,clip,keepaspectratio]{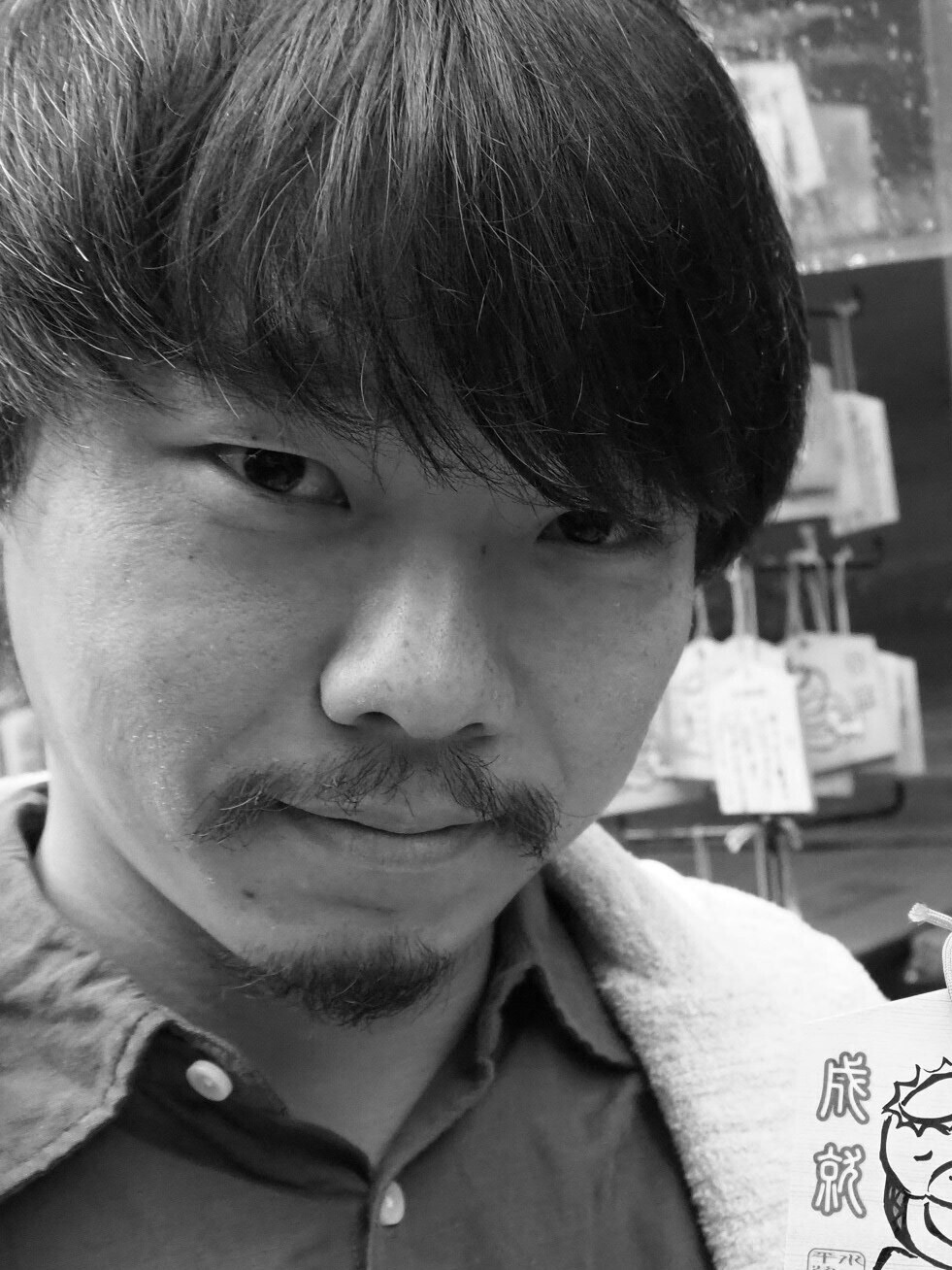}}]
{Masaya Nakata} (Member, IEEE) received a Ph.D. degree in informatics from the University of Electro-Communications, Tokyo, Japan, in 2016. 

He is an Associate Professor with the Faculty of Engineering, Yokohama National University, Japan. He was mainly working on evolutionary machine learning, data mining, and theoretical analysis of learning classifier systems.  His contributions have been published through more than 25 journal papers and 50 conference papers, such as \textsc{IEEE Transactions on Evolutionary Computation}, IEEE CEC, GECCO, and PPSN. 
He chaired the International Workshop on Learning Classifier Systems in 2015, 2016, 2018-2020 in GECCO. 
Since 2019, he has been focusing his research on surrogate-assisted evolutionary algorithms.

\end{IEEEbiography}
\vfill
\end{document}